%% file: main.tex
\documentclass[10pt,twocolumn,letterpaper]{article}

\usepackage[pagenumbers]{iccv} % To force page numbers, e.g. for an arXiv version

\input{preamble}

\definecolor{iccvblue}{rgb}{0.21,0.49,0.74}
\usepackage[pagebackref,breaklinks,colorlinks,allcolors=iccvblue]{hyperref}
\usepackage{xargs}
\usepackage{arydshln}
\usepackage{graphicx}   % For including images
\usepackage{subcaption} % For subfigures
\usepackage{adjustbox}  % For row labels

\newcommandx{\Farnood}[1]{\customComment{Farnood}{magenta}{#1}}
\newcommandx{\Tunc}[1]{\customComment{Tunc}{seagreen}{#1}}
\newcommandx{\Amir}[1]{\customComment{Amir}{violet}{#1}}

\def\paperID{13645} % *** Enter the Paper ID here
\def\confName{ICCV}
\def\confYear{2025}

\title{Stylized Structural Patterns for Improved Neural Network Pre-training}

\author{
Farnood Salehi$^{1}$\thanks{Equal contribution.} \quad Vandit Sharma$^2$\footnotemark[1] \quad Amirhossein Askari Farsangi$^1$ \quad Tun\c{c} Ozan Ayd{\i}n$^{1,2}$\\
$^1$Disney Research \textbar \ Studios \quad $^2$ETH Zürich \\
}

\begin{document}
\maketitle
\input{sec/0_abstract}    
\input{sec/1_intro}

\input{sec/2_related_work}

\input{sec/3_methodology}
\input{sec/4_experiments}

\input{sec/5_discussion_limitations}
\input{sec/6_conclusion}
% \input{sec/_2_formatting}
% \input{sec/_3_finalcopy}

{
    \small
    \bibliographystyle{ieeenat_fullname}
    \bibliography{main}
}
\input{sec/X_suppl}

\end{document}

%% file: preamble.tex
\usepackage{tikz}
\usepackage{pgfplots}
\usetikzlibrary{patterns}

\usepackage{xcolor}
\definecolor{lightgray}{gray}{0.8} % Define light gray color if allowed
\definecolor{forestgreen}{rgb}{0.133,0.545,0.133}

\usepackage{colortbl}

\usepackage[ruled,vlined]{algorithm2e}
\usepackage{algpseudocode}

\SetKwBlock{Initialization}{Initialization:}{end}

\newcommand{\RSNF}{\textit{reverse-stylized Neural Fractals}}
\newcommand{\nnst}{\textit{NNST}}
\newcommand{\gatys}{\textit{Gatys}}

%% file: sec/0_abstract.tex
\begin{abstract}
Modern deep learning models in computer vision require large datasets of real images, which are difficult to curate and pose privacy and legal concerns, limiting their commercial use. Recent works suggest synthetic data as an alternative, yet models trained with it often underperform. This paper proposes a two-step approach to bridge this gap. First, we propose an improved neural fractal formulation through which we introduce a new class of synthetic data. Second, we propose \emph{reverse stylization}, a technique that transfers visual features from a small, license-free set of real images onto synthetic datasets, enhancing their effectiveness. 
We analyze the domain gap between our synthetic datasets and real images using Kernel Inception Distance (KID) and show that our method achieves a significantly lower distributional gap compared to existing synthetic datasets. Furthermore, our experiments across different tasks demonstrate the practical impact of this reduced gap. We show that pretraining the EDM2 diffusion model on our synthetic dataset leads to an 11\% reduction in FID during image generation, compared to models trained on existing synthetic datasets, and a 20\% decrease in autoencoder reconstruction error, indicating improved performance in data representation. Furthermore, a ViT-S model trained for classification on this synthetic data achieves over a 10\% improvement in ImageNet-100 accuracy. Our work opens up exciting possibilities for training practical models when sufficiently large real training sets are not available.

\end{abstract}

%% file: sec/1_intro.tex
\section{Introduction}
\label{sec:intro}

%The era of deep learning for computer vision was kick-started by 
The introduction of the ImageNet dataset~\cite{deng2009imagenet} was one of the main catalysts for the use of deep learning models in computer vision. 
Today, most computer vision models rely on extensive collections of real-world image data for training. As model architectures grow in complexity and achieve superior performance, the demand for larger datasets has surged, pushing the boundaries of data collection efforts, which remain challenging and costly.
%Likewise, most computer vision models today are trained using large collections of real image data. Although these models achieve superior performance, collecting large-scale real image data for training is quite challenging and expensive. 
These datasets usually come with concerns related to privacy (e.g., including sensitive personal information), licensing (e.g., including copyright-protected data), and ethics (e.g., low wages for data labeling workers) in the curation process, making them unsuitable for the production environment.

Recent works have shown that computer generated images, which we call synthetic datasets, can be used to pre-train deep learning models. Several classes of synthetic datasets, such as dynamical systems-based (e.g., FractalDB~\cite{kataoka2020pre}), noise-based (e.g. statistical noise~\cite{baradad2021learning}), and CGI-based (e.g., Shaders~\cite{baradad2022procedural}), have been proposed for training computer vision models by previous work. 
With some fine-tuning on a small amount of real data, these datasets lead to significant improvements over training from scratch. However, there still exists a significant difference in the standalone performance of models trained with synthetic datasets vs. real datasets, primarily due to the domain gap between real and synthetic images. Moreover, previous methods have predominantly focused on discriminative tasks. In contrast, we focus on generative tasks, which are crucial for applications such as image synthesis, data augmentation, and representation learning, where high-quality synthetic data can significantly impact downstream performance. % \Farnood{I added focus on generative tasks, so the reviewers do not complain about classification results.}

% Moreover, previous methods have predominantly focused on discriminative tasks, which may not fully address the complexities of this domain gap.
% With some fine-tuning on a small amount of real data these datasets lead to significant improvements over training from scratch. However, there still exists a significant difference in the standalone performance of models trained with synthetic datasets, vs real datasets, primarily due to the domain gap between real and synthetic images.

In this work, we tackle this problem by making the following contributions:

% In this work, we aim to reduce this gap by proposing neural stylization of existing synthetic datasets with a small sample of real data to address the domain gap. We show that the pseudo-synthetic datasets resulting from this procedure achieve significant performance gains over the models trained with just the raw synthetic data. 

\textit{1. Neural Fractals:} We propose an improved formulation of neural fractals, which are derived from dynamical systems based on complex-valued neural networks. Our improved neural fractals outperform previous synthetic datasets on three out of our four evaluation pipelines, contributing to the existing literature on synthetic data for training computer vision models. 

\textit{2. Synthetic Dataset Stylization:} In typical image stylization, the contents of real images are combined with the style of an artwork, resulting in images with visual features of the \emph{style image} and the contents of the \emph{content image}. We, however, are interested in creating synthetic images with visual features that resemble real images. To this end we introduce \emph{reverse-stylization}, in which we transfer features from a small sample of license-free real images to synthetic images. We show that a modest dataset of only 7000 real images is sufficient to effectively reverse-stylize a significantly larger synthetic dataset, leading to improvements in evaluation metrics of up to 20\% compared to models trained solely on existing synthetic datasets or limited real data.

\textit{3. Extensive Experimental Evaluation:}
We conduct experiments on three tasks: Image Autoencoding, Image Generation, and Representation Learning. Our models trained on reverse-stylized neural fractals outperform those trained on baseline datasets. This improvement can be attributed to the reduced domain gap between our synthetic data and real images. \autoref{tab:cosine_similarity} highlights this with a lower KID score (third column). KID measures the distribution difference between synthetic and real data (ImageNet). A lower KID score suggests that models trained on our dataset produce feature representations that better align with those from real data (first two columns). Further details of this experiment are provided in the experiments section.

\begin{table}[t]
    \renewcommand{\arraystretch}{0.8} % Reduce space between rows
    \setlength{\tabcolsep}{6pt} % Reduce horizontal padding
    \small % Set a smaller font size
    \centering
    \begin{tabular}{lc| c | c}
        \toprule
        & \multicolumn{1}{c}{\footnotesize{\textbf{AutoEncoding}$\uparrow$}}  & \multicolumn{1}{c}{\footnotesize{\textbf{ImageGen}$\uparrow$}} & \multicolumn{1}{c}{\scriptsize{\textbf{Domain Gap}$\downarrow$}}\\ 
        \cmidrule(lr){2-2} \cmidrule(lr){3-3} \cmidrule(lr){4-4}
        \footnotesize{\textbf{Train}}/\footnotesize{\textbf{Eval Data}} & \footnotesize{\textbf{ImgNet}} & \footnotesize{\textbf{ImgNet}} & \footnotesize{\textbf{KID}} \\ 
        \midrule
        VisualAtom & 0.28 & 0.42 & 0.203 \\
        Mandelbulb & 0.17 & 0.35 & 0.165\\
        \footnotesize{FracDB-comp} & 0.50 &  0.51 & 0.214\\
        \footnotesize{StGAN-oriented} & 0.47 &  0.54 & 0.170\\
        \midrule
        \footnotesize{NeuFractal} (ours) & \textbf{0.55} & \textbf{0.57} & \textbf{0.162} 
        \\
        % \rowcolor{lightgray}
        \midrule
        +\footnotesize{RevStylize} (ours) & \textbf{0.68} & \textbf{0.61} & \textbf{0.120} \\
        \bottomrule
    \end{tabular}
    \caption{
    The first two columns show cosine similarity between attention maps of a network trained on ImageNet-100 and those trained on synthetic datasets. The left column corresponds to an autoencoder \cite{rombach2022high} trained for image reconstruction, and the middle column to EDM2 \cite{karras2024analyzing} trained with a diffusion loss. The rightmost column reports the Kernel Inception Distance (KiD) between each synthetic dataset and ImageNet, measuring distribution-level similarity. Neural fractal achieves the highest cosine similarity and the lowest KiD score, indicating the closest alignment with ImageNet-trained network and the smallest domain gap among all evaluated datasets.}
\label{tab:cosine_similarity}
\end{table}

%% file: sec/2_related_work.tex
\section{Related Work}
\label{sec:related_work}

\subsection{Fractals}
\label{fractals}

Fractals are geometric patterns that exhibit great complexity and self-similarity at different scales. They are found everywhere in nature, from snowflakes to tree leaves~\cite{gunther2024fractals}. They can also be generated using computer programs using simple rules. Due to these properties, fractals have been investigated by previous works such as Kataoka et al.~\cite{kataoka2020pre} and Anderson et al.~\cite{anderson2022improving} as a potential replacement for real datasets to pre-train vision models.

There is a wide variety of fractals generated by several techniques~\cite{wiki:fractalgeneration}. The Mandelbrot set~\cite{mandelbrot1983fractal}, defined as complex numbers $\displaystyle c$ where $\displaystyle f_{c}(z)=z^{2}+c$ does not diverge from $\displaystyle z=0$, is the most well-known. Its intricate boundaries are promising for synthetic data, though zooming in is computationally expensive and often yields featureless areas. {\citet{asadi2021neuralfractal} proposed a neural dynamical system for visual fractal exploration, but it faces several challenges that limit its practicality beyond visual curiosity: it uses monochromic coloring, is slow to render, and requires manual hyperparameter tuning. In contrast, we address these challenges in the context of synthetic data generation, introducing an adaptive rendering technique to improve rendering speed, an escape-time coloring algorithm that produces diverse RGB images, and an algorithm for automatic hyperparameter tuning.}

% Mathematically, there exists a wide variety of fractals, and they can be generated using several techniques~\cite{wiki:fractalgeneration}. In this work, we focus on a particular class of fractals known as escape-time fractals, which use a recurrence relation to determine the fractal form. Arguably, the most popular example of an escape-time fractal is the Mandelbrot set~\cite{mandelbrot1983fractal}. The Mandelbrot set is defined in the complex plane as the set of complex numbers $\displaystyle c$ for which the function $\displaystyle f_{c}(z)=z^{2}+c$ does not diverge to infinity when iterated starting at $\displaystyle z=0$. Zooming into the boundary of the Mandelbrot set can reveal incredibly complicated and diverse patterns making it a potential candidate to generate high-quality synthetic training data. However, programmatically zooming into the boundary of the Mandelbrot set is extremely tricky, as one can easily land into uninteresting or featureless areas. It is also computationally expensive, especially at higher zoom levels.

\subsection{Synthetic Data Pre-training}

The idea of using synthetically generated data to train vision models was first explored by Kataoka et al.~\cite{kataoka2020pre}. They introduced FractalDB, a synthetic dataset consisting of mathematically-generated fractal images, which could outperform ImageNet-1k~\cite{deng2009imagenet} and Places365~\cite{zhou2017places} pre-trained models with some fine-tuning. Anderson et al.~\cite{anderson2022improving} improved fractal pre-training by proposing a better sampling process for Iterated Function System (IFS) codes leading to higher quality fractal images. \citet{nakashima2022can} demonstrated the effectiveness of FractalDB to pre-train Vision Transformers~\cite{dosovitskiy2020image}, or ViTs, and observed that ViTs pay attention to contour lines in the pre-training phase. Subsequently, works such as Kataoka et al.~\cite{kataoka2022replacing} and Takashima et al.~\cite{takashima2023visual} explored contour-based datasets built using overlapping polygons (RCDB) and sinusoidal waves (VisualAtom) respectively, the latter achieving state-of-the-art fine-tuning performance. Unlike our method, FractalDB and VisualAtom is designed for supervised training.
\citet{shinoda2023segrcdb} extended the concept of synthetic pre-training from classification to semantic segmentation, and introduced the SegRCDB dataset with automatically-generated segmentation labels. Baradad et al.~\cite{baradad2021learning} investigated if images generated from procedural noise processes could suffice to train vision models. They experimented with several models (procedural graphics, statistical, StyleGAN initializations, etc.) and concluded that models trained on structured noise could indeed learn meaningful features. Baradad et al.~\cite{baradad2022procedural} curated a large collection of OpenGL shaders to generate a diverse and effective procedural graphics-based dataset of synthetic images.

Using synthetic data comes with several advantages: 1. It is free from privacy, ethical, and legal issues, 2. Data labelling can be performed cheaply and accurately, 3. It can be generated in large quantities without signficant human effort, and 4. It is quite effective for pre-training vision models. 
However, existing synthetic datasets still perform significantly worse than real-world datasets, particularly for image generation tasks such as image autoencoding and image diffusion. Our work aims to narrow this gap.

\subsection{Neural Style Transfer}

Neural Style Transfer (NST) is a Deep Learning-based technique that allows an ordinary image (\textit{content} image) to be generated in the style of an artistic image (\textit{style} image). 
% As an example, one could use it to reproduce a real portrait in the style of \textit{Starry Night} by Vincent Van Gogh. 
While image stylization has been a long-standing field of research, the use of deep neural networks to transfer artistic style across images was first proposed by Gatys et al.~\cite{gatys2015neural}. Their approach uses a bipartite loss function to optimize the resulting image. The first part (\textit{content} loss) aimed to preserve content features with respect to the \textit{content} image, and is derived using higher layers from a CNN network. The second part (\textit{style} loss), is responsible for preserving style features with respect to the \textit{style} image, which are modelled using Gram matrices derived from feature representations at intermediate CNN layers. The authors showed that such \textit{content} and \textit{style} representations are separable, thus allowing the generation of images that combine traits of both \textit{content} and \textit{style} images using optimization.

Since then, NST has gained tremendous popularity. It has been expanded to various domains Cai et al.~\cite{cai2023image} and several new approaches have been proposed. Jing et al.~\cite{jing2019neural} conducted a literature review of the advances in this area and performed a qualitative and quantitative comparison of various approaches. Kolkin et al.~\cite{kolkin2022neural} proposed the Neural Neighbor Style Transfer (NNST) approach using nearest-neighbour matching to replace neural features in the \textit{content} image with ones from the \textit{style} image, resulting in high-quality stylized images. More recently with the Generative AI wave, several diffusion-based ~\cite{rombach2022high} techniques~\cite{hamazaspyan2023diffusion, zhang2023inversion, ruta2023diff} have also been proposed. In this work, we restrict ourselves to two texture-based style transfer methods: Gatys et al.~\cite{gatys2015neural} and Kolkin et al.~\cite{kolkin2022neural}.

%% file: sec/3_methodology.tex
\section{Methodology}
\label{sec:methodology}

We propose a novel two-step approach to generate high-quality synthetic data for training neural networks for a wide variety of computer vision tasks. The first step involves generating a base synthetic dataset. Here, we propose an improved formulation of neural fractals as a new type of synthetic data. These are fractal patterns generated using dynamical systems defined by complex-valued neural networks.
The second step involves stylizing the base synthetic dataset using a small sample of real images. We show that this step helps boost the dataset quality tremendously as it incorporates real image features while retaining properties of the base synthetic dataset. 

\subsection{Neural Fractal Generation}

We explore neural fractals as a potential alternative to the Mandelbrot set limitations described in Section~\ref{fractals}. Neural fractals are derived from the following generalization of the original Mandelbrot recurrence relation:
\begin{equation}
  z_{n+1} = g(z_{n}) + c, \hspace{1em}  z_0=0,
  \label{eq:recurrence}
\end{equation}
where $c=x+y\cdot j\in \mathbb{C} $ represents the continuous 2D coordinate $(x,y)$ in the image plane as a complex number. In the case of the Mandelbrot set, $\displaystyle g(z)=z^{2}$. However, in principle, $\displaystyle g$ can be extended to any real or complex-valued function. Neural fractals use a randomly initialized complex-valued neural network as $\displaystyle g$. To determine whether a coordinate $c=x+y\cdot j$ belongs to the fractal set, the recurrence relation in \autoref{eq:recurrence} is iterated for a fixed number of iterations $T$ to obtain $z_T(c)$. Points for which $z_T(c)$ remains bounded (i.e., $|z_T(c)|\leq\tau$ where $\tau$ is the threshold ) belong to the fractal set. To rasterize the final image, we  integrate over each pixel $P$ by computing
\begin{equation}
  \int_{(x,y)\in P} 1\{|z_T(x+y\cdot j)|\leq\tau\} dxdy.
  \label{eq:rasterize}
\end{equation}
Similar to \cite{asadi2021neuralfractal}, our implementation works as follows: To compute the integral in \autoref{eq:rasterize}, we use Monte Carlo estimation.
At the start of each epoch, a subset of points $\{c_0,\ldots,c_n\}$ is uniformly sampled from the available image plane (a pre-determined rectangular window in the complex plane based on the center's coordinates). Next, we compute $z_T(c_i)  \; \forall \; i$ by iteratively applying \autoref{eq:recurrence}.
% these points are iteratively passed through the MLP-based recurrence relation \eqref{eq:recurrence} for a fixed number of iterations $T$. 
At the end of this iterative process, the pixel value is determined by counting the number of $c_i$s within the pixel for which $|z_T(c_i)|$ remains within the threshold $\tau$.

\begin{figure}[t]
    \centering
    \begin{minipage}{0.45\linewidth}
        \centering
        \includegraphics[width=\textwidth]{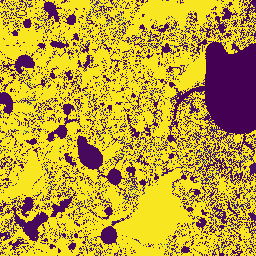}
        \subcaption{Image after 1 pass}
        \label{fig:image50}
    \end{minipage}
    \hspace{0.2cm}
    \begin{minipage}{0.45\linewidth}
        \centering
        \includegraphics[width=\textwidth]{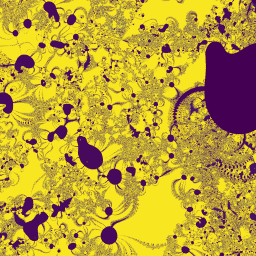}
        \subcaption{Image after 50 passes}
        \label{fig:image1}
    \end{minipage}
    \caption{Rendering the neural fractal is a noisy process.}
    \label{fig:noisyness}
\end{figure}

% all points whose iterated value $\displaystyle z_T(c_i)$ exceeds a pre-determined threshold are discarded, and the number of remaining points within a pixel determines the pixel value. 
Note that this process is inherently noisy, as different points landing in a pixel can have different locations and subsequently different values (see \autoref{fig:noisyness}).
To mitigate this noise, the rendering process is repeated over several passes, allowing each pixel to converge to its final value.
After completing all passes, the intensity of each pixel in the final image is determined by averaging the results across all passes. 
This procedure results in a grayscale image by default, which is then colorized, whereby pixel intensities are mapped to colors using a color map that we discuss next.

Each random initialization of the network weights results in a unique and complex recurrence relation, rendering a completely unique fractal image. Since neural fractals derive variations in the fractal pattern from the random initialization of network weights and not the zoom location or level, they offer a  practical way to generate practically infinite, diverse fractal patterns. We use a fully connected neural network with 6 hidden layers of 3 neurons each, with complex weights drawn from a normal distribution. The network uses tanh activation, and the output of the network is raised to the third power.
In later sections, we show that this process generates meaningful fractal images that perform well for training vision models.

\paragraph{Selecting $\tau$ in \autoref{eq:rasterize}} 
We choose $\tau$ in \autoref{eq:rasterize} to ensure the rendered fractal exhibits intricate structure and visual richness. If $\tau$ is too small or too large, the image becomes entirely black or white, respectively. Complicating the selection, the optimal $\tau$ depends on the neural network $g$, which varies for each fractal. To address this, we employ an automatic adjustment process: after an initial rendering pass and obtaining values $|z|$, starting from a small initial value $\tau$ is iteratively increased by a factor of 1.1  until at least 40\% of the rendered values $|z|$ fall below the threshold. The remaining rendering process then uses this adjusted $\tau$, ensuring a well-balanced distribution of values, preserving structural detail. For completeness, we provide the algorithm in the Appendix.

\begin{figure}[t]
    \centering
    \begin{minipage}{0.4\linewidth}
        \centering
        \includegraphics[width=\textwidth]{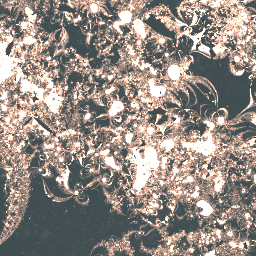}
        \subcaption{Pseudo-coloring}
        \label{fig:sampling1}
    \end{minipage}
    \hspace{0.2cm}
    \begin{minipage}{0.4\linewidth}
        \centering
        \includegraphics[width=\textwidth]{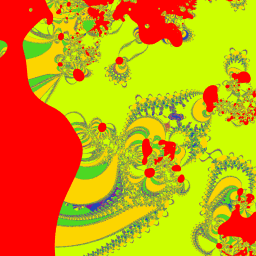}
        \subcaption{Escape time coloring}
        \label{fig:sampling2}
    \end{minipage}
    \caption{Pseudo-coloring vs Escape time coloring.}
    \label{fig:coloring}
\end{figure}

\paragraph{Escape Time Coloring}
%The original implementation utilizes pseudo-coloring to colorize the fractals, we implement coloring using the escape-time algorithm, which is an algorithm used to colorize fractals~\cite{mandelbrot1983fractal}. 
We adopt escape-time algorithm~\cite{mandelbrot1983fractal}, in which we keep track of \emph{escape count}, i.e. the number of iterations required for each sampled point to escape the threshold $\tau$.
%It works by additionally counting the number of iterations it takes for each sampled point to escape the set threshold, also referred to as the escape count. 
All points that escape the dynamical system with the same escape count (i.e. in a particular iteration) are then assigned the same color sampled randomly from a color map. Consequently, the pixels corresponding to these escaped points also hold the same color given the one-to-one mapping. Once a point escapes, its pixel color is frozen, and subsequent iterations are used to colorize only the remaining pixels. The final image is formed by averaging the intermediate images across all passes. Escape-time coloring reveals underlying patterns that are not captured by  pseudo-coloring scheme proposed in \cite{asadi2021neuralfractal}. \autoref{fig:coloring} compares the two coloring algorithms. 
% Finally, we use a function to dynamically set the escape threshold during the first pass, targeting a 40\% escape rate. This prevents rejecting too few or too many points during image generation. 
A colored dataset is crucial for applications such as image autoencoding and image generation pretraining.
We provide the full coloring algorithm in the Appendix.

\paragraph{Adaptive Sampling} Second, we tackle the issue of slow rendering, which arises from the naive sampling of the image plane. This naive approach requires a large number of iterations to remove noise and achieve a final high-quality fractal.
Inspired by adaptive sampling techniques in Monte Carlo rendering \cite{Hyperion}, we introduce an adaptive sampling of query points, which significantly accelerates the neural fractal generation process. 
% This method uses a non-uniform sampling distribution and it should sample points in a pixel based on its rendering complexity. 
Rather than uniformly sampling points across the entire image plane, adaptive sampling allocates samples nonuniformly according to the rendering difficulty of each pixel.
By selectively avoiding sampling pixels that are easy to render, this approach reduces computational costs and accelerates rendering. 

\begin{figure}[t]
    \centering
    \begin{minipage}{0.4\linewidth}
        \centering
        \includegraphics[width=\textwidth]{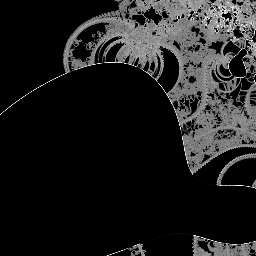}
        \subcaption{Estimated Variance at final iteration}
        \label{fig:sampling1b}
    \end{minipage}
    \hspace{0.2cm}
    \begin{minipage}{0.4\linewidth}
        \centering
        \includegraphics[width=\textwidth]{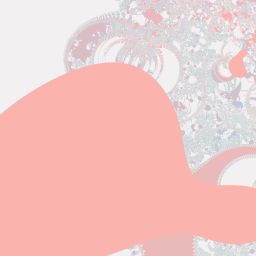}
        \subcaption{Rendered Image}
        \label{fig:sampling2b}
    \end{minipage}
    \caption{Adaptive sampling. As shown in the estimated variance, certain pixels converge more quickly and require fewer samples.}
    \label{fig:grid}
\end{figure}

\begin{figure}[t]
   \centering
   \includegraphics[width=\linewidth]{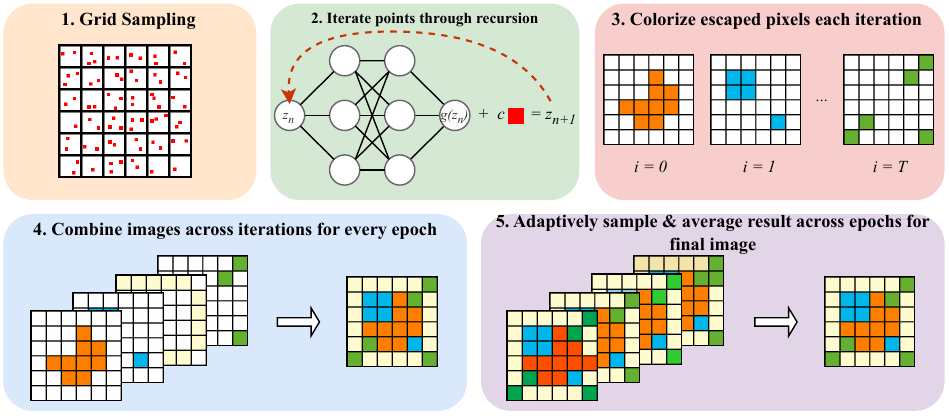}
   \caption{Overview of the neural fractal generation process.}
   \vspace{-1em}
   \label{fig:neural_fractal_generation}
\end{figure}

In a first pass, we distribute a few samples uniformly per pixel and render their corresponding color. We calculate the variance of these rendered colors for each pixel and smoothen the resulting variance map with a box filter using a kernel of size 5. This smoothed map is normalized and serves as a sampling map to allocate the samples in the next step. After each pass, we update the variance map and adjust the sampling distribution accordingly. Our experiments show that adaptive sampling can increase rendering speed on average by a factor of 8 and produce visually noiseless images. Using a single RTX 3090 GPU, generating an image takes on average about 6 seconds.
\autoref{fig:grid} shows an example of the estimated variance and the rendered image. As is seen, the variance is very low across large portions of the image, allowing them to be rendered with a small number of samples. Our adaptive sampling algorithm reduces the sample density in these areas. We also add a stopping criterion to prevent distributing samples to a pixel if its estimated variance is below a certain threshold. Adaptive sampling does not alter image content; it only accelerates the rendering process.
\autoref{fig:neural_fractal_generation} shows the complete neural fractal generation process.
For details on the adaptive sampling algorithm, please refer to the Appendix.

% As a result, the process converges more quickly, requiring fewer epochs, as every pixel is guaranteed to have a reference point in each iteration. 
% We use estimated online variance as a measure of the rendering difficulty for each pixel, ensuring that sample distribution is proportional to complexity. 
% First, we introduce grid sampling, which drastically speeds up the neural fractal generation process. Instead of randomly sampling points across the entire image space, grid sampling randomly samples one point from the image space corresponding to each pixel bin per epoch. This leads to faster convergence (fewer epochs) as all pixels are guaranteed to have a reference point in each epoch. Figure~\ref{fig:grid} highlights the difference between grid and random sampling.


Similar to \citet{baradad2021learning}, we generate a total of 100K, $256\times256$, random neural fractal images to use as our base synthetic dataset. For each image, we re-initialize the MLP network $g$ with random weights. To prevent featureless images from being included in the dataset, we exclude all generated image samples having a maximum standard deviation less than 0.025 across any channel. 

As baselines, we use several synthetic datasets such as StyleGAN-oriented~\cite{baradad2021learning}, FractalDB-composite~\cite{anderson2022improving}, VisualAtom~\cite{takashima2023visual}, and Mandelbulb~\cite{chiche2024pre}, which have been shown to perform well for synthetic data training in prior work. We generate 100K, $256\times256$, samples for each dataset using the original code provided by the authors.
% \footnote{Except Shaders21k-MixUp, where we randomly sample \& crop images from the provided online dataset.} 
Samples from all base datasets have been provided in \autoref{fig:base_grid} in Appendix.

Notably, networks trained on neural fractal exhibit a smaller domain gap than those trained on other synthetic datasets. As shown in the first two columns of \autoref{tab:cosine_similarity}, the attention map of a network trained on neural Fractal aligns more closely with that of a network trained on ImageNet, as measured by cosine similarity. This alignment is further supported by the lower KID score in the third column, indicating reduced distributional discrepancy.
The complete set of neural fractal experiments is in \autoref{sec:exp_nfractal}.

\subsection{Neural Fractal Reverse-stylization}

Previous works~\cite{baradad2021learning, kataoka2020pre} in the direction of synthetic pre-training have revealed that there still exists a large gap between the performance of models trained on synthetic data compared to real data such as ImageNet~\cite{deng2009imagenet}. We also experience this effect first hand when training purely on neural fractals mentioned above. We believe that one shortcoming of synthetic data is the lack of realistic textures and an inaccurate color distribution, which may contribute to this performance gap. In order to bridge this gap, we look into neural stylization of synthetic datasets as a way to boost their performance. 

Neural style transfer is typically used in the setting of artistic image generation, where the contents of a photograph are combined with the style of an artwork. This results in an image with visual features of the style image, but overall contents of the content image. We, however, are interested in creating synthetic datasets with visual features that resemble real datasets, to bridge the domain gap. We do this by running neural style transfer in the opposite direction, allowing the visual features of a real image to be transferred to a synthetic image. 
%With this motivation, we now describe our procedure to generate semi-synthetic datasets. 

\begin{figure}[t]
   \centering
   \includegraphics[width= 0.8\linewidth]{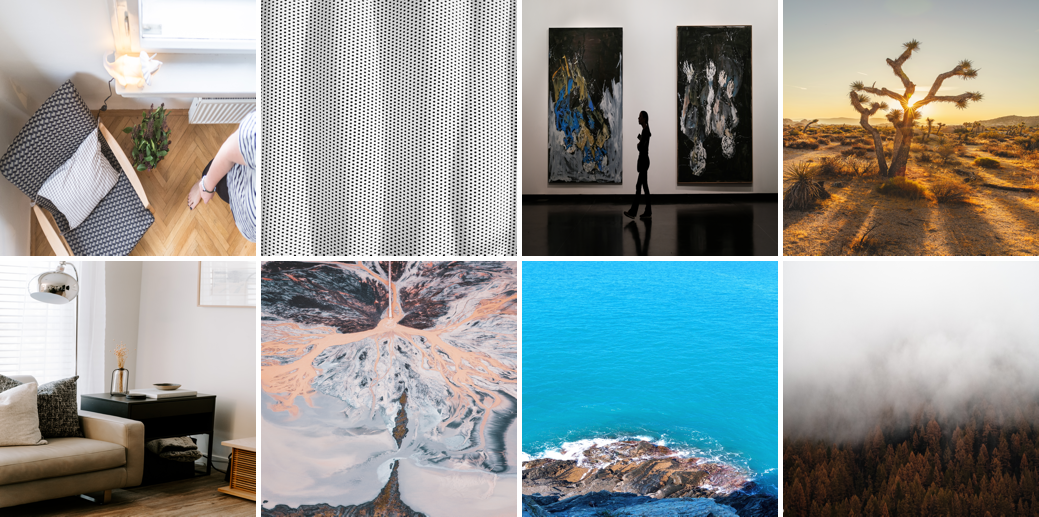}
   \caption{Samples from our \textit{Unsplash} dataset.}
   \label{fig:unsplash}
\end{figure}

%We begin by
To demonstrate our approach, we collect a small dataset of around 7k, real-world, license-free images, by scraping the website Unsplash\footnote{\href{https://unsplash.com/}{https://unsplash.com/}}. We downsize the images to $256\times256$ for our use case. We refer to the resulting dataset as \textit{Unsplash} (\autoref{fig:unsplash}), and use it for stylization as well as the real data baseline in our experiments. Unlike large-scale datasets such as ImageNet, collecting a small dataset requires significantly less manpower, and can be manually verified to be free of legal or privacy issues. 
%Samples from this dataset have been provided in Figure~\ref{fig:unsplash}. 
Next, we select a base synthetic dataset to be stylized. Here, we experiment and report results using our own neural fractal dataset, as well as other synthetic datasets. To generate style-transferred synthetic datasets, we iterate through each sample in a base synthetic dataset, and stylize it using a randomly sampled real image from the \textit{Unsplash} dataset. We experiment with two texture-based neural style transfer algorithms: The original algorithm by Gatys et al.~\cite{gatys2015neural}, and the NNST method proposed later by Kolkin et al.~\cite{kolkin2022neural}.
%
%\textbf{Gatys et al.~\cite{gatys2015neural}}: The original neural style transfer algorithm. Hereafter referred to as \textit{Gatys}.

%\textbf{Kolkin et al.~\cite{kolkin2022neural}}: An improved style transfer that gives visually superior results. Hereafter referred to as \textit{NNST}.

Using this procedure, we generate generate 100K, $256\times256$, samples for each combination of base synthetic dataset and stylization algorithm.  \autoref{fig:nnst_grid} show samples from all stylized synthetic datasets using  \textit{NNST} stylization algorithms, samples from \textit{Gatys} stylized method are shown in Appendix (\autoref{fig:gatys_grid}). Note that both stylization algorithms originally use features derived from a VGG network (trained on ImageNet). Therefore, we replace it with an encoder network pre-trained on \textit{Unsplash} to prevent any biases in results arising from using a network trained on large-scale real data. Based on our testing, the results from either yield visually very similar results. 

\autoref{tab:cosine_similarity} shows that reverse stylization reduces the domain gap significantly, as evidenced by both the improved alignment of attention maps with the ImageNet-trained network and the lower KID score, indicating closer distributional similarity.
Details on reverse stylization experiments are provided in \autoref{sec:exp_stylized}. 

% Note that we do not report results on fine-tuned models as commonly done in previous work~\cite{takashima2023visual, kataoka2020pre, kataoka2022replacing}. This is because our main goal is to make synthetic pre-training as strong as possible in itself by closing the domain gap with real images.

% Gatys image was here before. We moved it to Appendix

\input{sec/figs/nnst_grid}

%% file: sec/figs/nnst_grid.tex
\begin{figure*}[htbp]
    \centering
    \begin{minipage}{0.46\textwidth}
        \centering
        \includegraphics[width=\textwidth]{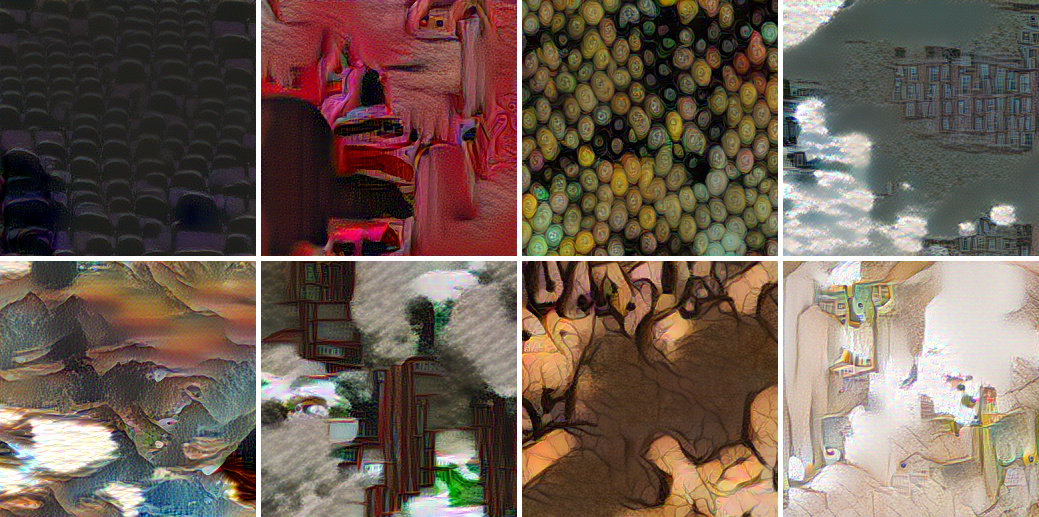}
        \subcaption{Neural Fractal-\textit{NNST}}
        \label{fig:image6}
    \end{minipage}
    \hspace{0.2cm}
    \begin{minipage}{0.46\textwidth}
        \centering
        \includegraphics[width=\textwidth]{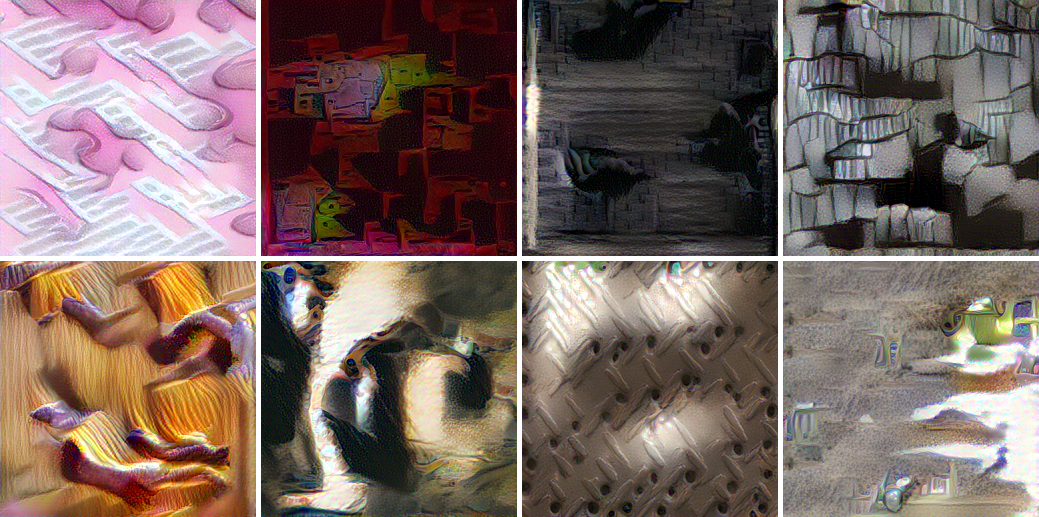}
        \subcaption{StyleGAN-oriented-\textit{NNST}}
        \label{fig:image5}
    \end{minipage}
    \vspace{0.2cm}
    \begin{minipage}{0.3\textwidth}
        \centering
        \includegraphics[width=\textwidth]{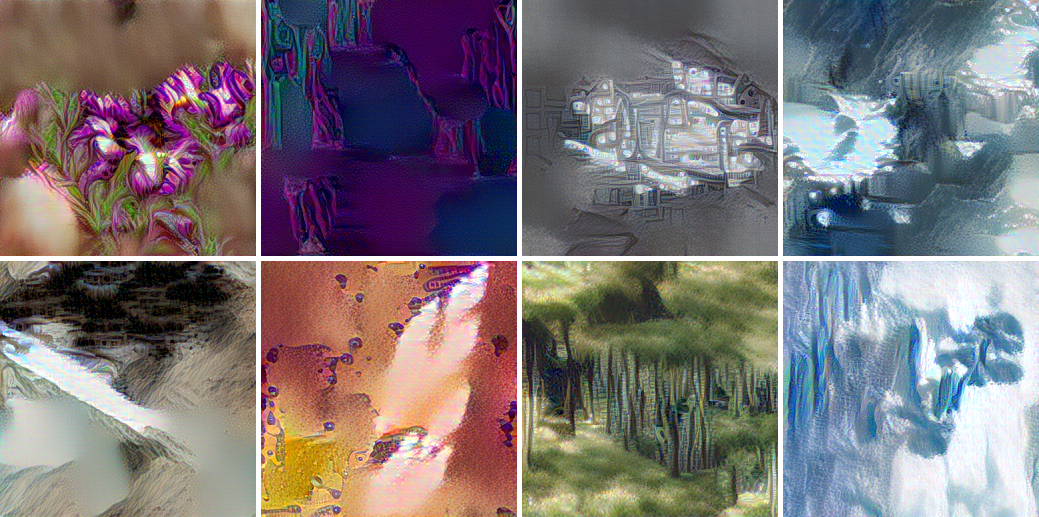}
        \subcaption{FractalDB-composite-\textit{NNST}}
        \label{fig:image2}
    \end{minipage}
    \hspace{0.2cm}
    \begin{minipage}{0.3\textwidth}
        \centering
        \includegraphics[width=\textwidth]{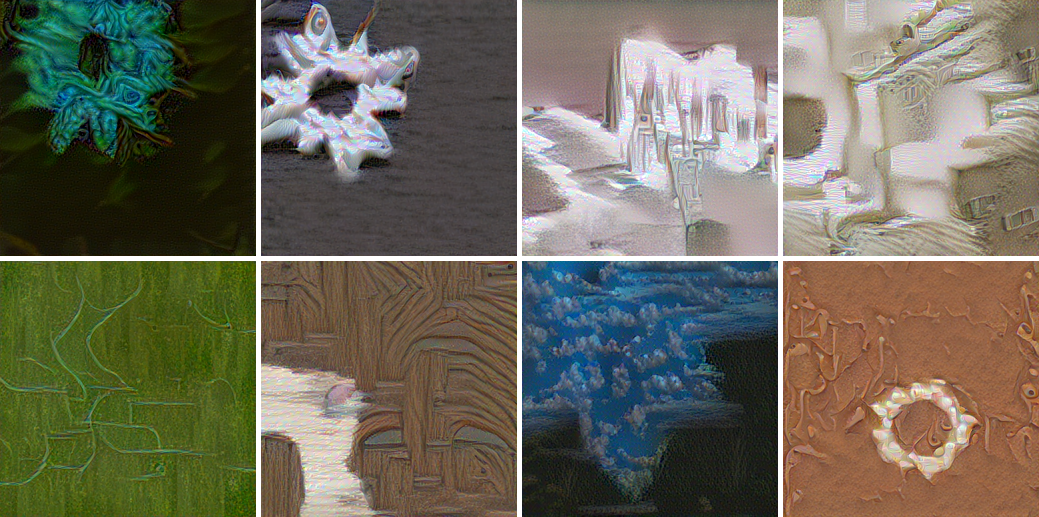}
        \subcaption{VisualAtom-\textit{NNST}}
        \label{fig:image1}
    \end{minipage}
    \hspace{0.2cm}
    \begin{minipage}{0.3\textwidth}
        \centering
        \includegraphics[width=\textwidth]{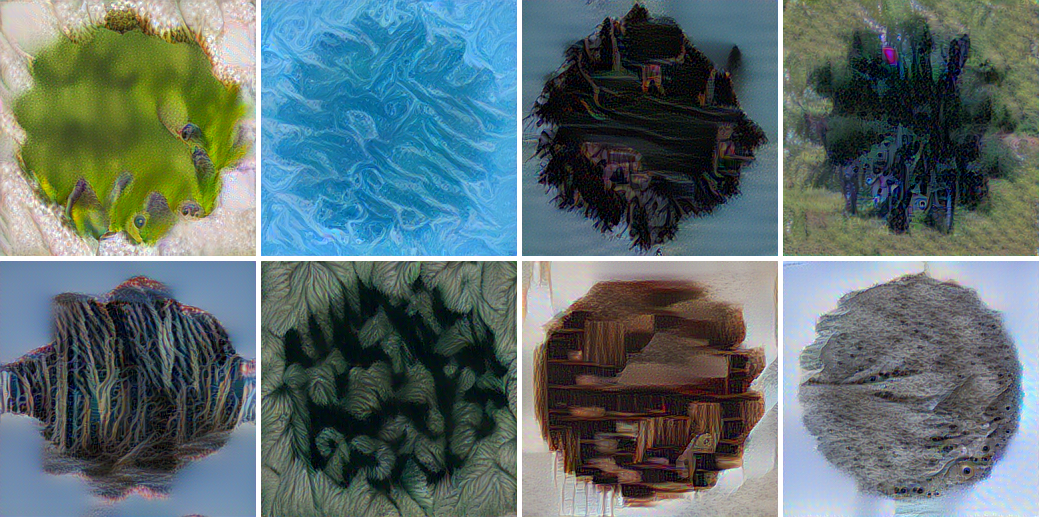}
        \subcaption{Mandelbulb-\textit{NNST}}
        \label{fig:image3}
    \end{minipage}
    \caption{Samples from different \textit{NNST}-stylized synthetic datasets.}
    \label{fig:nnst_grid}
\end{figure*}

%% file: sec/4_experiments.tex
\section{Experiments}
\label{sec:experiments}

We conduct experiments to test the quality of our synthetic datasets using three evaluation pipelines. 
% Since the generated neural fractal dataset does not have any labels, we investigate performance in an unsupervised setting. 
In \autoref{sec:exp_nfractal}, we compare the performance of neural fractals against other synthetic datasets proposed in previous works. Then, in \autoref{sec:exp_stylized}, we conduct experiments for our stylized synthetic dataset. Finally, we present ablation studies to gain a better understanding of the effect of different parameters.
 
% Finally, we experiment with pre-training Stable Diffusion on our synthetic datasets in Section~\ref{exp_diffusion}.

\subsection{Training with Neural Fractals}
\label{sec:exp_nfractal}

\textbf{Autoencoder}: We begin by training an autoencoder network designed for perceptual compression using an implementation and hyperparameters similar to Rombach et al.~\cite{rombach2022high}. We train solely using the perceptual loss~\cite{zhang2018unreasonable} and disable the adversarial objective (implemented for local realism) to avoid domain gap issues. We train on synthetic datasets and test on ImageNet's and COCO's~\cite{lin2014microsoft} validation dataset. We choose this network for its high reconstruction quality. Training hyperparameters are provided in the Appendix.
%and because we also pre-train Stable Diffusion in Section~\ref{exp_diffusion}. 
Importantly, we also use its encoder as a VGG replacement to generate features for style transfer. We train the network for 1M iterations.

\autoref{tab:base_autoencoder} shows the final reconstruction loss for different testing datasets. As can be seen, neural fractals and StyleGAN-oriented performs the best among all synthetic datasets.

\begin{table}[t]
\renewcommand{\arraystretch}{0.8} % Reduce space between rows
\setlength{\tabcolsep}{6pt} % Reduce horizontal padding
\small % Set a smaller font size
% \resizebox{\linewidth}{!}{
\centering
\begin{tabular}{lcc}
\toprule
\textbf{Train}/\textbf{Eval dataset} & \textbf{ImgNet} & \textbf{COCO} \\
\midrule
ImageNet-100 & 0.090 &  0.074 \\
\textit{Unsplash} & 0.153 &  0.121\\
\midrule
VisualAtom & 0.320 & 0.303 \\
Mandelbulb & 0.616 & 0.506 \\
FractalDB-composite & 0.224 & 0.195 \\
StyleGAN-oriented & \underline{0.138} &\textbf{0.114} \\
\midrule
\rowcolor{lightgray}
Neural Fractals & \textbf{0.137}  & \underline{0.116} \\
\bottomrule
\end{tabular}
% }
\caption{Reconstruction loss of autoencoder networks trained on different Train datasets and evaluated on ImgNet and COCO validation datasets.} %* indicates our synthetic baseline for the Autoencoder experiment.}
\vspace{-1em}
\label{tab:base_autoencoder}
\end{table}

\textbf{Diffusion}: 
To evaluate our approach in image generation tasks, we experiment with diffusion models, using the EDM2 framework~\cite{karras2024analyzing}. 
We adopt a network architecture based on the EDM2 repository, following their design guidelines. Similar to \cite{karras2024analyzing}, we downscale the images to a resolution of 64 and train the diffusion model in image space.
Full network and training hyperparameters are provided in the Appendix.
After pretraining the model on synthetic datasets, the network is fine-tuned on two small datasets, Flowers (8K images) and FFHQ (70K images).
~
We follow the standard FID computation procedure in \cite{karras2024analyzing} with 50K generated images and apply the independent condition guidance method from \cite{sadat2024no} to generate the images.
\autoref{tab:base_edm2} shows the results. Neural fractals perform the best among all synthetic datasets. 
\begin{table}[t]
\renewcommand{\arraystretch}{0.8} % Reduce space between rows
\setlength{\tabcolsep}{6pt} % Reduce horizontal padding
\small % Set a smaller font size
% \resizebox{\linewidth}{!}{
\centering
    \begin{tabular}{lcc}
        \toprule
        \textbf{Pretrain}/\textbf{Target dataset} & \textbf{Flowers}  & \textbf{FFHQ}\\
        \midrule
        ImageNet-100 & 10.2 & 7.2\\
        \textit{Unsplash} & 29.0 & 9.1\\
        \midrule
        VisualAtom & 43.1 & 8.3\\
        Mandelbulb & 28.8 & 8.5\\
        FractalDB-composite & 18.9 & 8.4\\
        StyleGAN-oriented & 20.1 & 8.1\\
        \midrule
        \rowcolor{lightgray}
        Neural Fractals & \textbf{18.3} & \textbf{7.4} \\
        \bottomrule
    \end{tabular}
% }
\caption{FID measured after pretraining with pretrain data and fine-tuning on the target dataset.} %* indicates our synthetic baseline for the experiment.}
\vspace{-1em}
\label{tab:base_edm2}
\end{table}

\textbf{Contrastive Learning}: 

Inspired by the contrastive learning-based evaluation pipeline from~\cite{baradad2022procedural}, we use DINO~\cite{caron2021emerging} with a ViT-S backbone.
DINO is pre-trained self-supervised with a contrastive loss on synthetic datasets, and we train only a linear classifier on the frozen features to evaluate representation quality---unlike~\cite{chiche2024pre,takashima2023visual}, which use fully supervised training of the entire network. We train the encoder for 1000 epochs before adding the linear classifier on top.
\input{sec/tables/accuracy_tab}

% 
\autoref{tab:base_contrastive2} (a) shows the top-1 accuracy achieved by different datasets for our contrastive learning pipeline. Neural fractals achieve the highest accuracy across all validation datasets. For completeness, we also tried a larger classifier head consisting of a linear layer, batch normalization, ReLU, and another linear layer. This setup also yielded the same result: neural fractals achieved the best performance.

% Similar results were reported in the previous work by Nakashima et al.~\cite{nakashima2022can}, who found that FractalDB performed better for ViT compared to CNN-based models.

\subsection{Training with Reverse Stylized Neural Fractals}
\label{sec:exp_stylized}

% We now repeat the same experiments performed in Section~\ref{sec:exp_nfractal}, but with stylized synthetic datasets. 
In this section, we evaluate the effectiveness of reverse stylization in enhancing network performance. We use algorithms similar to  \nnst{} and \gatys{}. For brevity, we refer to these simply as \nnst{} and \gatys{}.
Both stylization algorithms significantly improve the quality of neural fractals. 
% We use both \nnst{} and \gatys{}. is in the original paper

% \textcolor{red}{[In response to reviewer queries, we clarify that standard augmentation techniques were used following prior works. Additionally, we tested MixUp augmentation but found it provided minimal improvements compared to our reverse stylization approach. See Table Y for a comparison.]}

\textbf{Autoencoder}: The first set of results in \autoref{tab:stylized_all} presents the final reconstruction loss for an autoencoder trained on \RSNF{} dataset and tested on ImageNet-100 and COCO. Stylization reduces the loss by approximately $20\%$ compared to the non-stylized dataset, while \gatys{} outperform \nnst{}.
\input{sec/tables/rev_stylized_tab}

\textbf{Diffusion:}
We pretrain the network with \RSNF{} and finetune it on the Flowers and FFHQ datasets. We only use \nnst{} stylization in this section.
The second set of results in \autoref{tab:stylized_all} shows the FID reduction achieved through stylization. On Flowers, stylization reduces FID by 11\%, while on FFHQ, it achieves a 1\% reduction.
An uncurated selection of generated images is provided in the Appendix.

\textbf{Contrastive Learning}: The third set of results in \autoref{tab:stylized_all} shows the top-1 accuracy on ImageNet-100 and Flowers achieved by \RSNF{} using the DINO-based contrastive learning pipeline. Reverse stylization of the neural fractals significantly improves accuracy. \nnst{} outperforms \gatys{} in representation learning tasks.

In Appendix, we show that reverse stylization is effective not only for neural fractals but also consistently improves all other synthetic datasets. This suggests that reverse stylization can be a general technique for enhancing synthetic data across various applications. \autoref{fig:performance_summary} summarizes the performance of our approach for the three evaluation pipelines. Figures~\ref{fig:compare_autoencoder} and~\ref{fig:compare_dino} (Appendix) contain a more detailed performance comparison.
\input{sec/figs/performance_summary}

\subsection{Ablation Studies}
\label{sec:ablaiton}
\textbf{Domain Gap}: We analyze the domain gap between our \RSNF{} and real data. To quantify this gap, we compute the Kernel Inception Distance (KID) \cite{binkowski2018demystifying} between each synthetic dataset and ImageNet. We choose KID over FID due to its unbiased estimation and ability to report confidence intervals, which makes it more robust—particularly in high-divergence regimes. As shown in the third column of \autoref{tab:cosine_similarity}, the neural fractal dataset achieves the lowest KID with respect to ImageNet, indicating the closest distributional match. This gap is further reduced through our reverse stylization step, highlighting its effectiveness in aligning the synthetic data distribution with that of real images. For completeness, we report FID values in Appendix, which support the same conclusion.
Beyond distributional similarity, we further analyze the behavior of networks trained on different datasets. Specifically, we compare attention maps produced by autoencoders and diffusion models using ImageNet-100k trained-network as a reference. We calculate the cosine similarity between the attention maps of networks trained on synthetic datasets and those trained on ImageNet. Higher similarity indicates closer alignment in attended regions. Our method consistently achieves the highest similarity scores, as shown in \autoref{tab:cosine_similarity}, and is further supported by visualizations in \autoref{fig:feature_visualization_attention} (autoencoder attention maps) and \autoref{fig:feature_visualization_latents} (latent visualizations) in the Appendix. These results suggest a reduced domain gap compared to other synthetic datasets.

% We further analyze the behavior of the networks trained on different datasets. Specifically, the attention maps from autoencoder and diffusion models, using ImageNet-100k as reference. 
% We calculate cosine similarity between the attention maps of the network trained on synthetic data and ImageNet100.
% For the autoencoder, we input a sample and compute attention maps from its attention layers, while for diffusion, we add a small  noise and perform a single denoising step to extract maps. We calculate cosine similarity between these maps and those from the ImageNet-100k-trained model, with results in \autoref{tab:cosine_similarity}. 
% Higher similarity indicates closer alignment in attended regions. Ours consistently achieves the highest similarity scores (\autoref{tab:cosine_similarity}), supported visually by \autoref{fig:feature_visualization_attention} (autoencoder attention maps) and \autoref{fig:feature_visualization_latents} (latent visualizations) in Appendix, suggesting a reduced domain gap compared to other synthetic datasets.

\textbf{Impact of Dataset Scaling}:
In the diffusion experiment, we studied the effect of dataset scaling by generating 1 million images from an expanded neural fractals dataset and its stylized counterpart. We trained separate networks on each dataset and fine-tuned them on Flowers and FFHQ. Increasing dataset size reduced FID by approximately 15\% for the non-stylized version and 30\% for the reverse-stylized version (\autoref{tab:data_scale}).
\input{sec/tables/data_scale}

\textbf{Effect of style algorithm}: Comparing results between \textit{Gatys} and \textit{NNST} stylized datasets, we notice that \textit{NNST} tends to peform better on contrastive learning tasks, whereas \textit{Gatys} has a slight edge in the autoencoder setting. See \autoref{tab:stylized_autoencoder_full} and \autoref{tab:stylized_dino} in Appendix. 
% Further investigation is needed to explain this behavior.
%
% \textbf{Effect of base synthetic dataset}: Even though it is sometimes difficult to distinguish between the different base synthetic datasets simply through visual inspection of stylized samples, the base dataset does affect performance. From the results, we can observe that all stylized datasets with a neural fractal, StyleGAN-oriented, or FractalDB-composite base tend to perform better than the other base datasets.  Therefore, choosing a base that works well is important for our stylization-based approach.

\textbf{Impact of Network Architecture on Neural Fractals}:
%\textcolor{red}{For Tunc?}
We explore network complexity in neural fractals generation. 
%Higher complexity increases image frequency (see \autoref{sec:additional_result} in Appendix), making rendering harder and network calls slower. 
We observe that networks with more layers (see \autoref{sec:additional_result} in Appendix) are more likely to produce images with rich high-frequency details, while on the other hand making rendering harder and network calls slower.
To balance speed and detail, we use a 3-layer, 6-neuron network.

\textbf{Comparison to MixUp}: MixUp~\cite{zhang2017mixup} is a data augmentation technique that blends two input samples to improve model generalization. We compare Mixup with reverse stylization in \autoref{tab:mixup_vs_rstyle} and find that reverse stylization performs significantly better.
\input{sec/tables/mixup_vs_rstyle}

%% file: sec/tables/accuracy_tab.tex
\begin{table}[t]
 \renewcommand{\arraystretch}{0.8} % Reduce space between rows
 \setlength{\tabcolsep}{6pt}
 \small % Set a smaller font size
\centering
\begin{subtable}{0.45\textwidth}
\centering
\begin{tabular}{lccc}
\toprule
Dataset & ImgNet100 & Flowers & Food101 \\
\midrule
ImageNet-100 & 78.2 & 75.6 & 64.8 \\
Unsplash & 44.8 & 47.8 & 38.5 \\
\midrule
VisualAtom & 38.9 & 36.4 & 24.7 \\
Mandelbulb & 44.0 & 50.15 & 41.0 \\
FractalDB-composite & 45.3 & 55.6 & 37.4 \\
StyleGAN-oriented & 48.1 & 59.9 & 45.4 \\
\midrule
\rowcolor{lightgray}
\textbf{Neural Fractals} & \textbf{48.5} & \textbf{60.6} & \textbf{49.8} \\
\bottomrule
\end{tabular}
\caption{Results with a single linear classifier on frozen features.}
\end{subtable}
\hfill
\begin{subtable}{0.45\textwidth}
\centering
\begin{tabular}{lccc}
\toprule
Dataset & ImgNet100 & Flowers & Food101 \\
\midrule
ImageNet-100 & 70.7 & 80.3 & 67.1 \\
Unsplash & - & - & - \\
\midrule
VisualAtom & 44.2 & 42.6 & 30.7 \\
Mandelbulb & 48.5 & 55.1 & 45.2 \\
FractalDB-composite & 48.3 & 59.5 & 40.9 \\
StyleGAN-oriented & 51.5 & 63.5 & 48.9 \\
\midrule
\rowcolor{lightgray}
\textbf{Neural Fractals} & \textbf{56.3} & \textbf{65.6} & \textbf{55.1} \\
\bottomrule
\end{tabular}
\caption{Results using an MLP head with linear, batch normalization, ReLU, and another linear layer on frozen features.}
\end{subtable}
\caption{Top-1 accuracy on ImageNet-100, Flowers, and Food101 after training an encoder on synthetic data using the DINO pipeline, followed by training with different classifier head configurations on the respective datasets.}
\vspace{-1em}
\label{tab:base_contrastive2}
\end{table}

% \begin{table}[t]
%     \renewcommand{\arraystretch}{0.8} % Reduce space between rows
%     \setlength{\tabcolsep}{6pt} % Reduce horizontal padding
%     \small % Set a smaller font size
%     % \resizebox{\linewidth}{!}{
%     \centering
%     \begin{tabular}{lccc}
%         \toprule
%         \textbf{Dataset} & \textbf{{ImgNet100}} & \textbf{{Flowers}} & \textbf{{Food101}}\\
%         \midrule
%         \footnotesize{ImageNet-100} & 78.2 & 75.6 & 64.8\\
%         \textit{Unsplash}  & 44.8 & 47.8 & 38.5 \\
%         \midrule
%         VisualAtom  & 38.9 & 36.4 & 24.7\\
%         Mandelbulb  & 44.0 & 50.15 & 41.0 \\
%         {FractalDB-composite} & 45.3 & 55.6 & 37.4\\
%         {StyleGAN-oriented} & 48.1 & 59.9 & 45.4 \\
%         \midrule
%         \rowcolor{lightgray}
%         Neural Fractals & {\textbf{48.5}} & {\textbf{60.6}} & {\textbf{49.8}} \\
%         \bottomrule
%     \end{tabular}
%     % }
%     \caption{Top-1 accuracy on ImageNet-100, Flowers and Food101 after training an encoder on synthetic data using the DINO pipeline, followed by linear classifier training on the respective datasets.}
%     \vspace{-1em}
% \label{tab:base_contrastive2}
% \end{table}

%% file: sec/tables/rev_stylized_tab.tex
\begin{table*}[t]
\renewcommand{\arraystretch}{1} % Reduce space between rows
\setlength{\tabcolsep}{6pt} % Reduce horizontal padding
\footnotesize % Set a smaller font size
\centering
\begin{tabular}{l|cc|cc|ccc}
    \toprule
     & \multicolumn{2}{c|}{{\textbf{AutoEncoding}}\normalsize{$\downarrow$}} & \multicolumn{2}{c|}{{\textbf{ImageGen}}\normalsize{$\downarrow$}} & \multicolumn{3}{c}{\footnotesize{\textbf{Representation Learning}}\normalsize{$\uparrow$}} \\
    \cmidrule(lr){2-3} \cmidrule(lr){4-5} \cmidrule(lr){6-8} 
    \textbf{Dataset} & \textbf{ImgNet} & \textbf{COCO} & \textbf{Flowers} & \textbf{FFHQ} & \textbf{ImgNet} & \textbf{Flowers} & \textbf{Food101}\\
    \midrule
    Neural Fractals & 0.137 & 0.116 & 18.3 & 7.4 & 48.5 & 60.6 & 49.8 \\
    \midrule
    +RevStyl NNST & {0.111} & {0.088} & \textbf{16.3} & \textbf{7.3} & \textbf{59.2} & \textbf{71.3} & \textbf{55.0}
    \\
   & \textcolor{forestgreen}{+19\%} & \textcolor{forestgreen}{+24\%} & \textcolor{forestgreen}{+11\%} & \textcolor{forestgreen}{+1\%} & \textcolor{forestgreen}{+10.7\%} & \textcolor{forestgreen}{+10.7\%} & \textcolor{forestgreen}{+5.2\%}\\
   \midrule
   +RevStyl Gatys & \textbf{0.105} & \textbf{0.084} & - & - & {57.4} & {68.7} & 53.9\\
   & \textcolor{forestgreen}{+23\%} & \textcolor{forestgreen}{+28\%} & - & - & \textcolor{forestgreen}{+8.9\%} & \textcolor{forestgreen}{+8.1\%} & \textcolor{forestgreen}{+4.1\%}\\
    \bottomrule
\end{tabular}
\caption{Performance comparison of \RSNF{} in autoencoding, diffusion-based  generation and representation learning, using \nnst{} and \gatys{} for stylization. Improvement is computed as $100\times\frac{\text{NeuralFractal}-\text{RevStylNeuralFractal}}{\text{NeuralFractal}}$ for generation tasks and as $\text{RevStylNeuralFractal}-\text{NeuralFractal}$ for representation learning.
}
\label{tab:stylized_all}
\end{table*}
%Improvement=$100\times\frac{\text{RevStylNeuralFractal}}{\text{NeuralFractal}}$ and Improvement=$\text{RevStylNeuralFractal}-\text{NeuralFractal}$ for repres learning.
% Improvements are computed as $100\times\frac{\text{NeuralFractal+RevStlization}}{\text{NeuralFractal}}$.

%% file: sec/figs/performance_summary.tex
\begin{figure*}[htbp]
    \centering
    \begin{minipage}{0.33\textwidth}
        \centering
        \begin{tikzpicture}
            \begin{axis}[
                ybar,
                width=\textwidth, % Scale plot with the image width
                height=6.5cm,
                ymin=0,
                ymax=0.63,
                ylabel={Reconstruction Loss},
                ylabel style={yshift=-0.3cm}, % Adjust y-axis label position
                xtick=\empty, % Remove x ticks
                symbolic x coords={Dataset1, Dataset2, Dataset3, Dataset4, Dataset5, Dataset6, Dataset7, Dataset8}, % Dynamically handle x coords
                xticklabels={}, % Remove x labels
                bar width=0.4cm, % Dynamic width scaling with linewidth
                enlarge x limits={abs=0.25cm}, % Dynamic spacing
                nodes near coords, % Enable nodes near coordinates for dataset names
                every node near coord/.append style={rotate=90, anchor=west, font=\footnotesize}, % Rotate dataset names inside bars
                grid=major,
                major grid style={dotted},
                axis line style={-},
            ]
            
            % Bar plot with different colors and labels inside the bars
            \addplot[
                ybar,
                fill=yellow,
                nodes near coords={\text{Mandelbulb}}, % Dataset name inside the bar
                every node near coord/.append style={anchor=east},
                bar shift=0pt
            ] coordinates {(Dataset1,0.616)};
            
            \addplot[
                ybar,
                fill=yellow,
                nodes near coords={\text{VisualAtom}}, % Dataset name inside the bar
                every node near coord/.append style={anchor=east},
                bar shift=0pt,
            ] coordinates {(Dataset2,0.32)};
        
            \addplot[
                ybar,
                fill=yellow,
                nodes near coords={\text{FractalDB-composite}}, % Dataset name inside the bar
                % every node near coord/.append style={font=\small},
                bar shift=0pt
            ] coordinates {(Dataset3,0.224)};
        
            \addplot[
                ybar,
                fill=black,
                nodes near coords={\text{\textit{Unsplash}}}, % Dataset name inside the bar
                every node near coord/.append style={text=black}, % Change text color to white
                bar shift=0pt
            ] coordinates {(Dataset4,0.153)};
        
            \addplot[
                ybar,
                fill=yellow,
                nodes near coords={\text{StyleGAN-oriented}}, % Dataset name inside the bar
                bar shift=0pt
            ] coordinates {(Dataset5,0.138)};
            
            \addplot[
                ybar,
                fill=pink, 
                nodes near coords={\text{Neural Fractals}}, % Dataset name inside the bar
                bar shift=0pt
            ] coordinates {(Dataset6,0.137)};
            
            \addplot[
                ybar,
                fill=magenta,
                nodes near coords={\text{Neural Fractals-\textit{Gatys}}}, % Dataset name inside the bar
                bar shift=0pt
            ] coordinates {(Dataset7,0.105)};
        
            \addplot[
                ybar,
                fill=black,
                nodes near coords={\text{ImageNet-100}}, % Dataset name inside the bar
                every node near coord/.append style={text=black}, % Change text color to white
                bar shift=0pt
            ] coordinates {(Dataset8,0.09)};
        
            \draw[dashed, red] (rel axis cs:0,0.242857) -- (rel axis cs:1,0.242857);
            \draw[dashed, red] (rel axis cs:0,0.142857) -- (rel axis cs:1,0.142857);
            
            \end{axis}
            
            % Adding images below bars
            \node[anchor=north] at (rel axis cs:0.125,0) {\includegraphics[width=0.35cm]{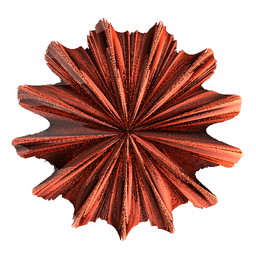}};
            \node[anchor=north] at (rel axis cs:0.251,0) {\includegraphics[width=0.35cm]{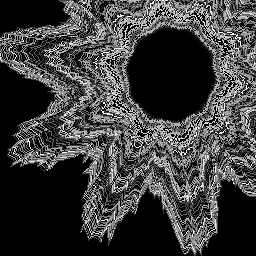}};
            \node[anchor=north] at (rel axis cs:0.377,0) {\includegraphics[width=0.35cm]{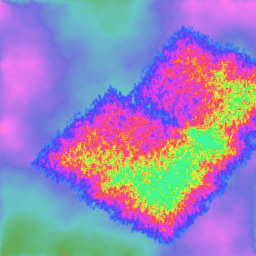}};
            \node[anchor=north] at (rel axis cs:0.5025,0) {\includegraphics[width=0.35cm]{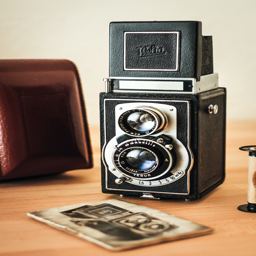}};
            \node[anchor=north] at (rel axis cs:0.628,0) {\includegraphics[width=0.35cm]{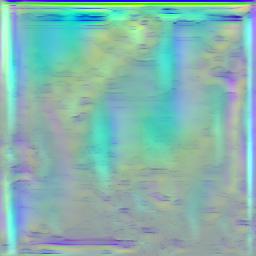}};
            \node[anchor=north] at (rel axis cs:0.754,0) {\includegraphics[width=0.35cm]{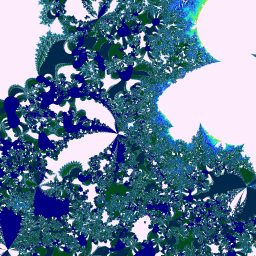}};
            \node[anchor=north] at (rel axis cs:0.880,0) {\includegraphics[width=0.35cm]{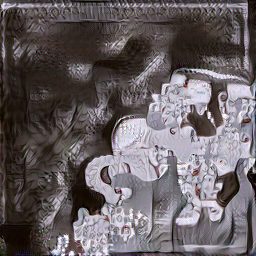}};
            \node[anchor=north] at (rel axis cs:1,0) {\includegraphics[width=0.35cm]{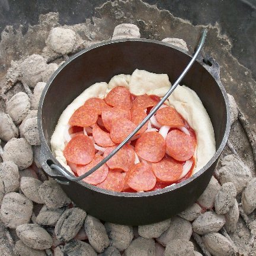}};
        \end{tikzpicture}
        \subcaption{Autoencoder}
        \label{fig:summary_autoencoder}
    \end{minipage}
    \begin{minipage}{0.33\textwidth}
        \centering
        \begin{tikzpicture}
            \begin{axis}[
                ybar,
                width=\textwidth, % Scale plot with the image width
                height=6.5cm,
                ymin=0,
                ymax=81,
                ylabel={Accuracy (\%)},
                ylabel style={yshift=-0.4cm},
                xtick=\empty, % Remove x ticks
                symbolic x coords={Dataset1, Dataset2, Dataset3, Dataset4, Dataset5, Dataset6, Dataset7, Dataset8}, % Dynamically handle x coords
                xticklabels={}, % Remove x labels
                bar width=0.4cm, % Dynamic width scaling with linewidth
                enlarge x limits={abs=0.25cm}, % Dynamic spacing
                nodes near coords, % Enable nodes near coordinates for dataset names
                every node near coord/.append style={rotate=90, anchor=east, font=\footnotesize}, % Rotate dataset names inside bars
                grid=major,
                major grid style={dotted},
                axis line style={-},
            ]
            
            % Bar plot with different colors and labels inside the bars
            \addplot[
                ybar,
                fill=yellow,
                nodes near coords={\text{VisualAtom}}, % Dataset name inside the bar
                bar shift=0pt,
            ] coordinates {(Dataset1,38.92)};
        
            \addplot[
                ybar,
                fill=yellow,
                nodes near coords={\text{Mandelbulb}}, % Dataset name inside the bar
                bar shift=0pt
            ] coordinates {(Dataset2,44.02)};
        
            \addplot[
                ybar,
                fill=black,
                nodes near coords={\text{\textit{Unsplash}}}, % Dataset name inside the bar
                every node near coord/.append style={text=white}, % Change text color to white
                bar shift=0pt
            ] coordinates {(Dataset3,44.8)};
        
            \addplot[
                ybar,
                fill=yellow,
                nodes near coords={\text{FractalDB-composite}}, % Dataset name inside the bar
                every node near coord/.append style={font=\footnotesize},
                bar shift=0pt
            ] coordinates {(Dataset4,45.32)};
        
            \addplot[
                ybar,
                fill=yellow,
                nodes near coords={\text{StyleGAN-oriented}}, % Dataset name inside the bar
                every node near coord/.append style={font=\small},
                bar shift=0pt
            ] coordinates {(Dataset5,48.06)};
            
            \addplot[
                ybar,
                fill=pink, 
                nodes near coords={\text{Neural Fractals}}, % Dataset name inside the bar
                bar shift=0pt
            ] coordinates {(Dataset6,48.48)};
            
            \addplot[
                ybar,
                fill=magenta, 
                nodes near coords={\text{Neural Fractals-\textit{NNST}}}, % Dataset name inside the bar
                bar shift=0pt
            ] coordinates {(Dataset7,59.18)};
        
            \addplot[
                ybar,
                fill=black,
                nodes near coords={\text{ImageNet-100}}, % Dataset name inside the bar
                every node near coord/.append style={text=white}, % Change text color to white
                bar shift=0pt
            ] coordinates {(Dataset8,78.2)};
        
            \draw[dashed, red] (rel axis cs:0,0.553086) -- (rel axis cs:1,0.553086);
            \draw[dashed, red] (rel axis cs:0,0.965432) -- (rel axis cs:1,0.965432);
            
            \end{axis}
            
            % Adding images below bars
            \node[anchor=north] at (rel axis cs:0.125,0) {\includegraphics[width=0.35cm]{figures/sample_VisualAtom.png}};
            \node[anchor=north] at (rel axis cs:0.251,0) {\includegraphics[width=0.35cm]{figures/sample_Mandelbulb.png}};
            \node[anchor=north] at (rel axis cs:0.377,0) {\includegraphics[width=0.35cm]{figures/sample_unsplash.png}};
            \node[anchor=north] at (rel axis cs:0.5025,0) {\includegraphics[width=0.35cm]{figures/sample_FractalDB++.png}};
            \node[anchor=north] at (rel axis cs:0.628,0) {\includegraphics[width=0.35cm]{figures/sample_StyleGAN_oriented.png}};
            \node[anchor=north] at (rel axis cs:0.754,0) {\includegraphics[width=0.35cm]{figures/sample_neural_fractal_colored.png}};
            \node[anchor=north] at (rel axis cs:0.880,0) {\includegraphics[width=0.35cm]{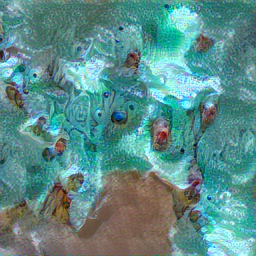}};
            \node[anchor=north] at (rel axis cs:1,0) {\includegraphics[width=0.35cm]{figures/sample_imagenet.png}};
        \end{tikzpicture}
        \subcaption{DINO}
        \label{fig:summary_dino}
    \end{minipage}    
    \begin{minipage}{0.33\textwidth}
        \centering
        \begin{tikzpicture}
            \begin{axis}[
                ybar,
                width=\textwidth, % Scale plot with the image width
                height=6.5cm,
                ymin=0,
                ymax=45,
                ylabel={FID},
                ylabel style={yshift=-0.3cm}, % Adjust y-axis label position
                xtick=\empty, % Remove x ticks
                symbolic x coords={Dataset1, Dataset2, Dataset3, Dataset4, Dataset5, Dataset6, Dataset7, Dataset8, Dataset9, Dataset10}, % Dynamically handle x coords
                xticklabels={}, % Remove x labels
                bar width=0.3cm, % Dynamic width scaling with linewidth
                enlarge x limits={abs=0.25cm}, % Dynamic spacing
                nodes near coords, % Enable nodes near coordinates for dataset names
                every node near coord/.append style={rotate=90, anchor=west, font=\footnotesize}, % Rotate dataset names inside bars
                grid=major,
                major grid style={dotted},
                axis line style={-},
            ]
            
            % Bar plot with different colors and labels inside the bars
            \addplot[
                ybar,
                fill=yellow,
                nodes near coords={\text{VisualAtom}}, % Dataset name inside the bar
                every node near coord/.append style={text=black, anchor=east},
                bar shift=0pt,
            ] coordinates {(Dataset1,43.1)};

            \addplot[
                ybar,
                fill=black,
                nodes near coords={\text{\textit{Unsplash}}}, % Dataset name inside the bar
                every node near coord/.append style={text=white, anchor=east}, % Change text color to white
                bar shift=0pt
            ] coordinates {(Dataset2,29)};
            
            \addplot[
                ybar,
                fill=yellow,
                nodes near coords={\text{Mandelbulb}}, % Dataset name inside the bar
                every node near coord/.append style={anchor=east},
                bar shift=0pt
            ] coordinates {(Dataset3,28.83)};

            \addplot[
                ybar,
                fill=yellow,
                nodes near coords={\text{StyleGAN-oriented}}, % Dataset name inside the bar
                every node near coord/.append style={text=black},
                bar shift=0pt
            ] coordinates {(Dataset4,20.1)};
        
            \addplot[
                ybar,
                fill=yellow,
                nodes near coords={\text{FractalDB-composite}}, % Dataset name inside the bar
                every node near coord/.append style={},
                bar shift=0pt
            ] coordinates {(Dataset5,18.85)};

            \addplot[
                ybar,
                fill=pink, 
                nodes near coords={\text{Neural Fractals}}, % Dataset name inside the bar
                every node near coord/.append style={text=black},
                bar shift=0pt
            ] coordinates {(Dataset6,18.37)};
            
            \addplot[
                ybar,
                fill=magenta,
                nodes near coords={\text{Neural Fractals-\textit{NNST}}}, % Dataset name inside the bar
                bar shift=0pt
            ] coordinates {(Dataset7,16.3)};
        
            \addplot[
                ybar,
                fill=black,
                nodes near coords={\text{ImageNet-100}}, % Dataset name inside the bar
                every node near coord/.append style={text=black}, % Change text color to white
                bar shift=0pt
            ] coordinates {(Dataset10,8.85)};

            % New datasets
            \addplot[
                ybar,
                fill=pink,
                nodes near coords={\text{Neural Fractals 1M}}, % Dataset name inside the bar
                bar shift=0pt
            ] coordinates {(Dataset8,15.6)};
            
            \addplot[
                ybar,
                fill=magenta,
                nodes near coords={\text{Neural Fractals-\textit{NNST} 1M}}, % Dataset name inside the bar
                bar shift=0pt
            ] coordinates {(Dataset9,10.2)};

            \draw[dashed, red] (rel axis cs:0,0.197) -- (rel axis cs:1,0.197);
            \draw[dashed, red] (rel axis cs:0,0.644) -- (rel axis cs:1,0.644);
            
            \end{axis}
            
            % Adding images below bars
            \node[anchor=north] at (rel axis cs:0.12,0) {\includegraphics[width=0.25cm]{figures/sample_VisualAtom.png}};
            % 2
            \node[anchor=north] at (rel axis cs:0.22,0) {\includegraphics[width=0.275cm]{figures/sample_unsplash.png}};
            %  3 
            \node[anchor=north] at (rel axis cs:0.32,0) {\includegraphics[width=0.275cm]{figures/sample_Mandelbulb.png}};
            % 4 
            \node[anchor=north] at (rel axis cs:0.415,0) {\includegraphics[width=0.275cm]{figures/sample_StyleGAN_oriented.png}};
            % 5
            \node[anchor=north] at (rel axis cs:0.515,0) {\includegraphics[width=0.275cm]{figures/sample_FractalDB++.png}};
            \node[anchor=north] at (rel axis cs:0.615,0) {\includegraphics[width=0.275cm]{figures/sample_neural_fractal_colored.png}};
            \node[anchor=north] at (rel axis cs:0.712,0) {\includegraphics[width=0.275cm]{figures/sample_unsplash_to_neural_fractal_colored_nnst.png}};
            % 877857143
            \node[anchor=north] at (rel axis cs:0.81,0) {\includegraphics[width=0.275cm]{figures/sample_neural_fractal_colored.png}};
            \node[anchor=north] at (rel axis cs:0.905,0) {\includegraphics[width=0.275cm]{figures/sample_unsplash_to_neural_fractal_colored_nnst.png}};
            \node[anchor=north] at (rel axis cs:1,0) {\includegraphics[width=0.275cm]{figures/sample_imagenet.png}};
        \end{tikzpicture}
        \subcaption{Image-Diffusion}
        \label{fig:summary_edm2}
    \end{minipage}   
    \caption{Dataset performance across evaluation pipelines. Most datasets have 100K images, except two with 1M in image-diffusion. Stylized synthetic datasets perform best. ImageNet is used for (a) and (b), Flowers for (c).}
    \label{fig:performance_summary}
\end{figure*}

%% file: sec/tables/data_scale.tex
\begin{table}[h]
    \renewcommand{\arraystretch}{1} % Reduce space between rows
    \setlength{\tabcolsep}{6pt} % Reduce horizontal padding
    \small % Set a smaller font size
    \centering
    \begin{tabular}{lcc}
        \toprule
        \textbf{Dataset} & \textbf{Flowers} & \textbf{FFHQ} \\
        \midrule
        nFractal 100K  & 18.4 & 7.4 \\
        nFractal 1M   & \textbf{15.6}  & \textbf{6.5} \\
        \midrule
        + RevStyle 100K & 16.3  & 7.3  \\
        + RevStyle 1M & \textbf{10.2}  & \textbf{5.5}  \\
        \bottomrule
    \end{tabular}
    \caption{Effect of dataset size on diffusion: Larger data improves FID for (non-)stylized neural fractal trained networks.}
    \label{tab:data_scale}
\end{table}

%% file: sec/tables/mixup_vs_rstyle.tex
\begin{table}[h]
    \centering
    \small % Set a smaller font size
    \begin{tabular}{lcc}
        \toprule
        \small{\textbf{Dataset / Method}} & \textbf{DINO} $\uparrow$ & \textbf{AEncoder} $\downarrow$ \\
        \midrule
        nFractal  & \underline{48.5} & 0.137 \\
        \midrule
        + MixUp   & 45.0  & \underline{0.128} \\
        + RevStylize  & \textbf{59.2}  & \textbf{0.105}  \\
        \bottomrule
    \end{tabular}
    \caption{Comparison of MixUp and reverse stylization  on DINO feature quality and Autoencoder reconstruction loss both on ImageNet. Higher DINO scores indicate better representation learning, while lower Autoencoder loss (AEncoder) indicates better reconstruction.}
    \vspace{-0.5em}
    \label{tab:mixup_vs_rstyle}
\end{table}

%% file: sec/5_discussion_limitations.tex
\section{Discussion \& Limitations}
\label{sec:discussion_limitations}

We showed that stylizing existing synthetic datasets with features from a small sample of real images greatly improves performance. Our reverse-stylization approach transfers visual features from real data to synthetic data, reducing the domain gap. 
Our approach has some limitations. First, for generating neural fractals, we restricted ourselves to a simple MLP architecture to model $\displaystyle g$ in recurrence relation~\ref{eq:recurrence}. However, it is likely that other model architectures can generate different or more sophisticated patterns. Perhaps, this is something that could be used to generate labeled neural fractal datasets for classification tasks. 
% Moreover, the recurrence relation itself could be modeled using sequential networks such as RNNs instead. 
In terms of stylization, we relied on a small sample of real images to transfer features to synthetic datasets. The exact effect of this real dataset on final results is not clear at the moment. 
% In our work, we treated stylization as a secondary step and performed it after the base synthetic datasets were completely rendered. However, previous works in synthetic training~\cite{anderson2022improving} have managed to generate synthetic data on-the-fly, saving both time and disk usage. We believe such an approach is also possible with stylization being a part of the generation pipeline. 
Jing et al.~\cite{jing2019neural} highlighted that there exist several stylization algorithms that can run in real time. However, this was not the focus of our work and can be explored in future studies.

% The first limitation is that our version of neural fractals still relies on a pre-determined color map. We experimented with directly sampling colors from the outputs of the complex-valued neural network used in the recurrence relation, and this has shown promising results. 

% We can observe that except for
% [53], [56], all the other model-optimisation-based NST algorithms are capable of stylising even high-resolution content
% images in real-time.

%% file: sec/6_conclusion.tex
\section{Conclusion}
\label{sec:conclusion}

In this work, we explored using synthetic data to train deep learning-based computer vision models. We proposed an improved version of neural fractals—a new class of synthetic training data derived from dynamical systems using complex-valued neural networks. Training models with neural fractals achieved the best performance across our synthetic dataset evaluation pipelines. To further boost performance, we introduced a technique involving neural stylization of synthetic datasets with a small sample of real images as a way of reducing the domain gap and adding meaningful features to synthetic data. The resulting stylized synthetic datasets achieved superior performance: up to 24\% for Autoencoder, 11\% for the DINO-based and 11\% for the diffusion evaluation pipelines. Stylization especially benefited weaker base synthetic datasets, with all stylized datasets managing to outperform the best synthetic data baseline. We believe our work will find applications in data-scarce settings, and hope that our strong results will pave the way for future research.

%Overall, we managed to achieve significant performance gains over prior synthetic data and narrow the gap with real data. 
%We firmly believe that this area has tremendous potential for improvements in future work.

%% file: sec/X_suppl.tex
\clearpage
\maketitlesupplementary

\section{Neural Fractal Generation}
\label{sec:supplementary_neural_fractal_gen}
In this section, we provide the pseudocode for the coloring and adaptive sampling algorithms.

\begin{algorithm}
\label{alg:select_tau}
\caption{Dynamic Threshold Adjustment}
% \begin{algorithmic}[1]
\KwIn{$z$ (first pass data from the rendering), ${ratio}$ (desired proportion), $\tau_{\text{init}}$ (initial threshold) }
\Initialization{
$\tau \gets \tau_{\text{init}}$ \Comment{Initialize threshold}

$\text{mask} \gets |z| \geq \tau$ \Comment{Create a binary mask for values above threshold}
}

\While{$\text{average(mask)} > \text{ratio}$}{
    $\tau \gets \tau \times 1.1$ \Comment{Increase threshold}
    
    $\text{mask} \gets |z| \geq \tau$ \Comment{Recompute mask}
}

\Return{ $\tau$ } \Comment{Final threshold}
% \State \textbf{return} $\tau$ \Comment{Final threshold}
% \end{algorithmic}
\end{algorithm}

\begin{algorithm}
\label{alg:escape_time_coloring}
\caption{Escape Time Coloring Algorithm}
\KwIn{$max\_iters$, $threshold$, $color\_map$ and a complex network $g$}
\KwIn{Sampled coordinate $c$ from the imageplane }
\KwOut{Color value of the sampled point}
\Initialization{
    $z \gets 0$ 
    
    $i \gets -1$
}
\While{$|z| \leq threshold$ \textbf{and} $i < max\_iters$}{
    $z \gets g(z) + c$ \Comment{$g(z)$ is the neural network}
    
    $i \gets i + 1$
}
$T_{\text{esc}} \gets i$ \Comment{$T_{\text{esc}}$ is the escape time}

\If{$T_{\text{esc}}==0$}{
$T_{\text{esc}} \gets max\_iters$
}

$T_{\text{esc}} \gets T_{\text{esc}}/max\_iters$

$color \gets color\_map[\text{len}(color\_map)*T_{\text{esc}}]$

\Return{$color$} 
\end{algorithm}

% \begin{algorithm}[H]
% \footnotesize
% \caption{Escape Time Coloring Algorithm}
% \label{alg:escape_time_coloring}
% \begin{algorithmic}[1]
% \Require $max\_iters$, $threshold$, $color\_map$, 
% \For{\textbf{each} pixel $(x, y)$ \textbf{in} image}
%     \State $c \gets$ \texttt{complex\_coordinate}$(x, y)$
%     \State $z \gets 0$
%     \State $iteration \gets 0$
%     \While{$|z| \leq threshold$ \textbf{and} $iteration < max\_iters$}
%         \State $z \gets g(z) + c$ \Comment{$g(z)$ is the neural network function}
%         \State $iteration \gets iteration + 1$
%     \EndWhile
%     \If{$iteration == max\_iters$}
%         \State $color \gets color\_map[0]$ \Comment{Point did not escape}
%     \Else
%         \State $color\_index \gets iteration \mod$ \texttt{length}$(color\_map)$ \\
%         \State $color \gets color\_map[color\_index]$
%     \EndIf
%     \State \texttt{Set pixel $(x, y)$ to color}
% \EndFor
% \end{algorithmic}
% \normalsize
% \end{algorithm}

\begin{algorithm}
\label{alg:adaptive_sampling}
\caption{Adaptive Sampling Algorithm}
\KwIn{$max\_epochs$, $max\_iters$, $initial\_samples\_per\_pixel$, Number of pixels $n$}
\KwOut{Color value of the the image plane}
\Initialization{
    $\mu = 0$ $\forall$pixels \Comment{$\mu$: sample mean}
    
    $N = 0$ $\forall$pixels \Comment{$N$: samples per pixel}
    
    $M_2 = 0$  $\forall$pixels \Comment{$M_2$: second moment}
}
\For{$i = 1$ \textbf{to} ${initial\_samples\_per\_pixel}$}{
    Sample $C=[c_1,\ldots,c_n]$ for each pixel
    
    image = get\_color($C$) \Comment{Use the escape time algorithm}
    
    $\mu = \mu + \text{image}$
    
    $M_2 = M_2 + \text{image}^2$
    
    $N \gets N +1$
}
\For{$i = 1$ \textbf{to} ${max\_epochs}$}{
$V = M_2 / N - (\mu/N)^2$ \Comment{Sample variance per pixel}

$V_m = V / N$ \Comment{Variance of mean per pixel}

$CV_2=\frac{V}{(\mu/N)^2}$ \Comment{Squared coefficient of variance per pixel}

$CV_2 \gets \text{average } CV_2 \text{ over RGB channels}$

$CV_2 \gets \text{box\_blur}(CV_2)$

$S \gets CV_2/\text{sum}(CV_2)$ \Comment{Sampling map}

\For{$j=1$ \textbf{to} $n$}{
    Sample a pixel $p_j$ from $S$
    
    Sample a coordinate $c_j$ inside $p_j$ uniformly
    
    cr = get\_color($c_j$)
    
    $\mu(p_j) = \mu(p_j) + \text{cr}$
    
    $M_2(p_j) = M_2(p_j) + \text{cr}^2$ 
    
    $N(p_j) \gets N(p_j) +1$ 
}}
\Return{$\mu/N$} 
\end{algorithm}

% \begin{algorithm}[H]
% \footnotesize
% \caption{Adaptive Sampling Algorithm}
% \label{alg:adaptive_sampling}
% \begin{algorithmic}[1]
% \Require $max\_epochs$, $max\_iters$, $initial\_samples\_per\_pixel$
% \State Initialize $variance\_map$ with large values
% \For{$epoch = 1$ \textbf{to} $max\_epochs$}
%     \State $sample\_map \gets$ \texttt{ComputeSampleMap}($variance\_map$)
%     \For{\textbf{each} pixel $(x, y)$ \textbf{according to} $sample\_map$}
%         \State $c \gets$ \texttt{complex\_coordinate}$(x, y)$
%         \State $z \gets 0$
%         \State $iteration \gets 0$
%         \While{$|z| \leq threshold$ \textbf{and} $iteration < max\_iters$}
%             \State $z \gets g(z) + c$
%             \State $iteration \gets iteration + 1$
%         \EndWhile
%         \State \texttt{Update pixel $(x, y)$ with new sample}
%         \State \texttt{Update $variance\_map$ at $(x, y)$}
%     \EndFor
%     \State Smooth $variance\_map$ using a box filter
%     \If{\texttt{convergence criteria met}}
%         \State \textbf{break}
%     \EndIf
% \EndFor
% \State \Return final image

% \Function{ComputeSampleMap}{$variance\_map$}
%     \State Normalize $variance\_map$
%     \State $sample\_map \gets$ Allocate samples based on $variance\_map$
%     \State \Return $sample\_map$
% \EndFunction
% \end{algorithmic}
% \normalsize
% \end{algorithm}

\section{Hyperparameters}

\textbf{AutoEncoder:}
We train the stable diffusion autoencoder (AE) following the hyperparameters in \cite{stability_autoencoder_config}. The latent space of AE has 4 channels as in  \cite{stability_autoencoder_config}.
We train the AE with images of resolution $128 \times 128$ and a batch size of 8. 
We use the Adam optimizer with the learning rate of $4.5 \times 10^{-6}$. We train the networks for $1.5$M iterations.

\textbf{DINO:}
We train a ViT-S encoder with DINO pipline with a batch size of $512$ for $200$ epochs. The ADAM optimizer is used with a learning rate of $0.0005$ with a linear warmup phase lasting $10$ epochs. The target learning rate at the end of optimization is $10^{-6}$, following a cosine learning rate schedule.
For weight decay (WD), we start with a value of $0.04$ and increase it to a final value of $0.4$ using a cosine schedule. As suggested in DINO, employing a larger weight decay towards the end of training enhances the performance of ViT-S. After training the ViT-S encoder, we train a single fully connected layer to map its outputs to the classes.
We train this layer for 100 epochs, and with Adam optimizer with learning rate $0.0001$ and momentum $0.9$. We also use cosine annealing learning rate.

\textbf{Diffusion:}
We use the EDM2-xs architecture from \cite{karras2024analyzing}. 
Karras et al. improved the denoising U-net to converge faster and achieve better results. We pretrain the network using a batch size of $256$ for $840$K iterations. 
We use the Adam optimizer with the learning rate $0.001$. We use the same learning rate decay in \cite{karras2024analyzing} with $T_{ref}=35000$. Dropout is not applied, and the original noise scheduler remains unchanged.
The Flowers dataset is relatively small, having approximately $8$K images. As a result, prolonged training leads to overfitting. 
Therefore, we fine-tune the model for $16$K iterations to avoid this problem.
In contrast, FFHQ is a larger dataset, containing $70$K images. For this setup, we fine-tune the model for $100$K iterations.

\section{Additional Results}
\label{sec:additional_result}
\label{sec:supplementary_figures}

\begin{figure*}[h]
    \centering
    \begin{tabular}{c c}  % Two columns: one for labels, one for images
        \begin{tabular}{c}  % Column for row labels (aligned center)
            \rotatebox{90}{\textbf{1 hidden}} \\[20mm]
            \rotatebox{90}{\textbf{3 hidden}} \\[20mm]
            \rotatebox{90}{\textbf{6 hidden}}
        \end{tabular} &
        \begin{tabular}{cccc} % 4 columns for images
            % Row 1: 1 Hidden Layer
            \includegraphics[width=0.2\textwidth]{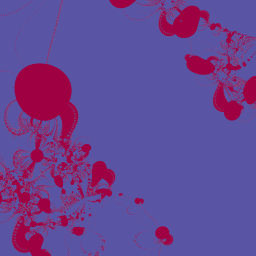} &
            \includegraphics[width=0.2\textwidth]{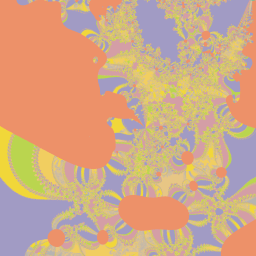} &
            \includegraphics[width=0.2\textwidth]{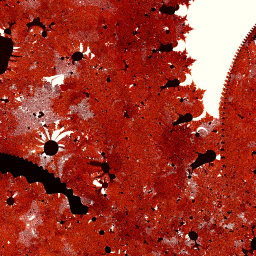} &
            \includegraphics[width=0.2\textwidth]{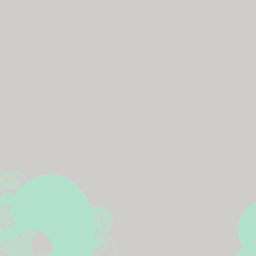} \\

            % Row 2: 3 Hidden Layers
            \includegraphics[width=0.20\textwidth]{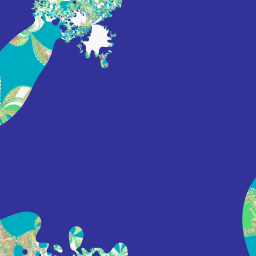} &
            \includegraphics[width=0.20\textwidth]{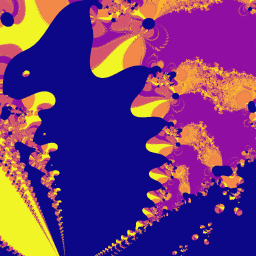} &
            \includegraphics[width=0.20\textwidth]{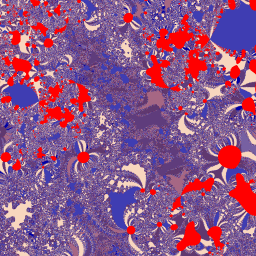} &
            \includegraphics[width=0.20\textwidth]{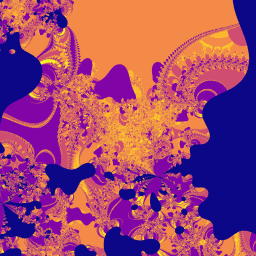} \\

            % % Row 3: 6 Hidden Layers
            \includegraphics[width=0.20\textwidth]{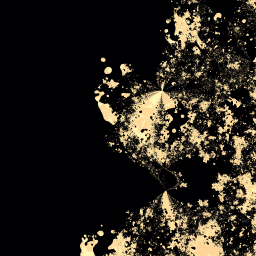} &
            \includegraphics[width=0.20\textwidth]{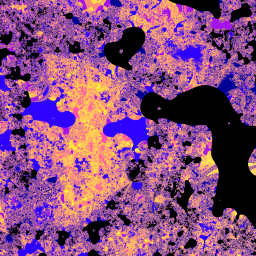} &
            \includegraphics[width=0.20\textwidth]{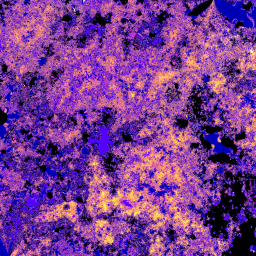} &
            \includegraphics[width=0.20\textwidth]{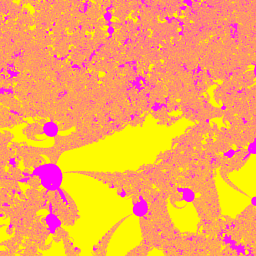} \\
        \end{tabular}
    \end{tabular}
    \caption{Uncurated samples generated from networks with 1, 3, and 6 hidden layers. The network with 6 hidden layers are more likely to generate images with excessively high frequencies.}
    \vspace{-0.25em}
    \label{fig:hidden_layers}
\end{figure*}

\textbf{Impact of Network Architecture on Neural Fractal:} We experiment with networks of varying complexity for neural fractal generation. Increasing the complexity of the network tends to increase the frequency of the output image (see \autoref{fig:hidden_layers}), making rendering more challenging as low-level details require more samples. Additionally, calling a more complex network slows down generation. To balance rendering speed and image richness, we use a network with 3 layers and 6 neurons. In \autoref{fig:hidden_layers}, we present uncurated samples from three networks with 1, 3, and 6 hidden layers.

\input{sec/tables/stylized_autoencoder_full}
% \input{sec/tables/mixup_vs_rstyle}

% AlexNet
% \begin{table}[t]
% \renewcommand{\arraystretch}{0.8} % Reduce space between rows
% \setlength{\tabcolsep}{6pt} % Reduce horizontal padding
% \small % Set a smaller font size
% % \resizebox{\linewidth}{!}{
% \centering
% \begin{tabular}{lll}
% \toprule
% \textbf{Base dataset} & \multicolumn{2}{c}{\textbf{AlexNet IN100 Accuracy}} \\
% & \textit{Gatys} & \textit{NNST}\\
% \midrule
% VisualAtom & 43.82 \textcolor{forestgreen}{(+0.96\%)} & 43.34 \textcolor{forestgreen}{(+0.48\%)}\\
% Mandelbulb & \textbf{45.60} \textcolor{forestgreen}{(+2.74\%)} & 44.82 \textcolor{forestgreen}{(+1.96\%)}\\
% FractalDB-composite & 44.20 \textcolor{forestgreen}{(+1.34\%)} & \textbf{46.28} \textcolor{forestgreen}{(+3.42\%)}\\
% StyleGAN-oriented & 45.02 \textcolor{forestgreen}{(+2.16\%)} & 45.36 \textcolor{forestgreen}{(+2.50\%)}\\
% \midrule
% \rowcolor{lightgray}
% Neural Fractal & 43.62 \textcolor{forestgreen}{(+0.76\%)} & 45.40 \textcolor{forestgreen}{(+2.54\%)}\\
% \bottomrule
% \end{tabular}
% % }
% \caption{Performance comparison of our stylized synthetic datasets for AlexNet trained using contrastive loss and linear evaluation on ImageNet-100. Improvements are calculated over the corresponding non-stylized synthetic baseline * from Table~\ref{tab:base_contrastive}.}
% \label{tab:stylized_alexnet}
% \end{table}

% DINO

\input{sec/tables/stylized_dino}
\input{sec/figs/base_grid}

\begin{figure*}[htbp]
    \centering
    \begin{minipage}{0.46\textwidth}
        \centering
        \includegraphics[width=\textwidth]{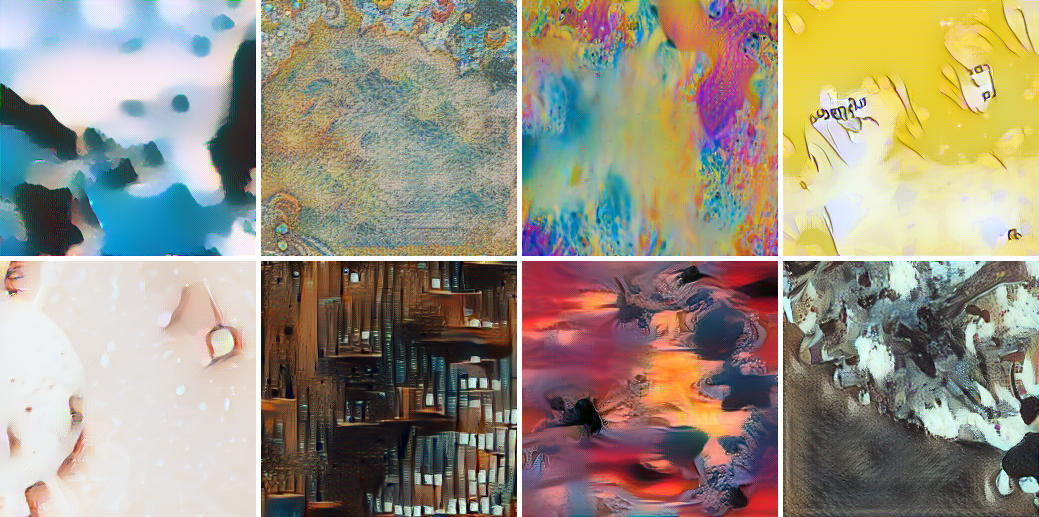}
        \subcaption{Neural Fractal-\textit{Gatys}}
        \label{fig:image6}
    \end{minipage}
    \hspace{0.2cm}
     \begin{minipage}{0.46\textwidth}
        \centering
        \includegraphics[width=\textwidth]{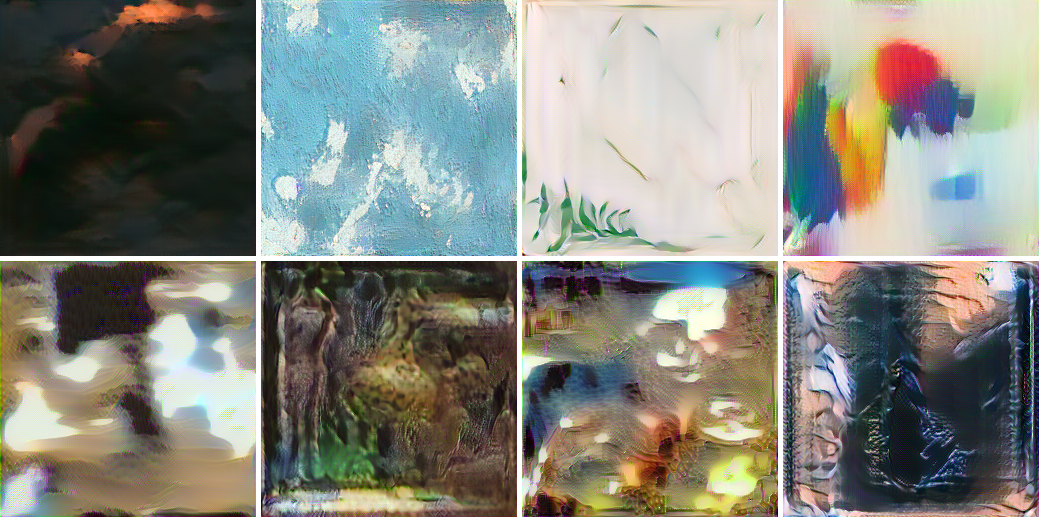}
        \subcaption{StyleGAN-oriented-\textit{Gatys}}
        \label{fig:image5}
    \end{minipage}
    \vspace{0.2cm}
    \begin{minipage}{0.3\textwidth}
        \centering
        \includegraphics[width=\textwidth]{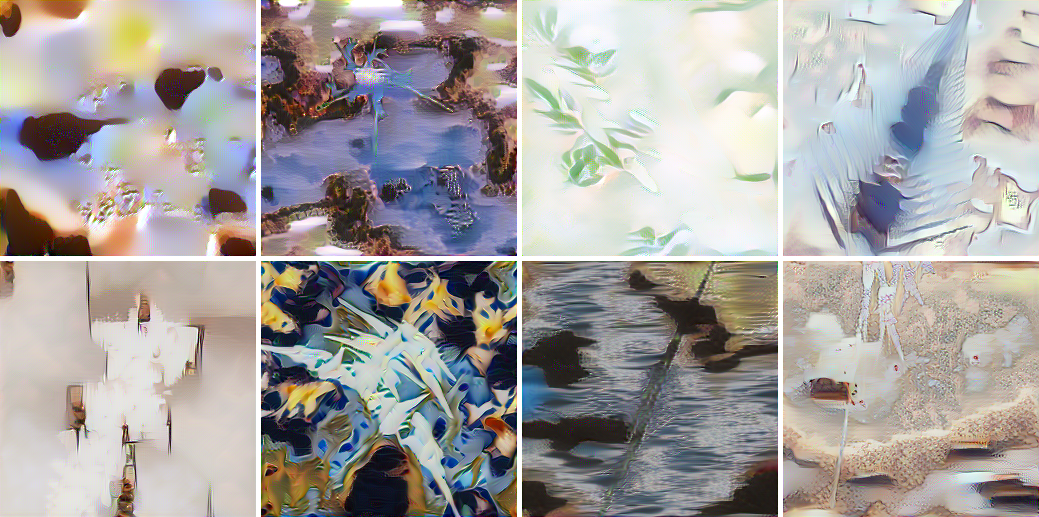}
        \subcaption{FractalDB-composite-\textit{Gatys}}
        \label{fig:image2}
    \end{minipage}
    \hspace{0.2cm}
    \begin{minipage}{0.3\textwidth}
        \centering
        \includegraphics[width=\textwidth]{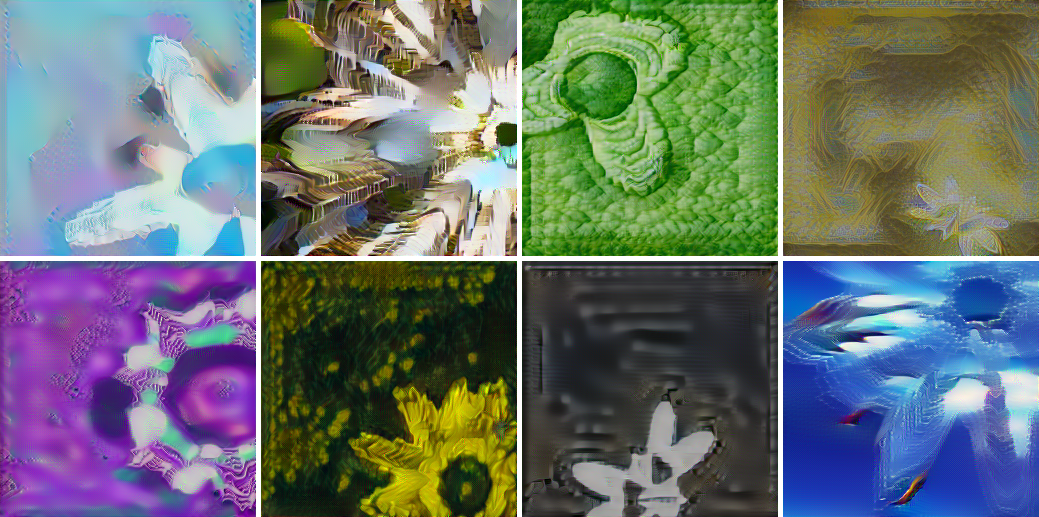}
        \subcaption{VisualAtom-\textit{Gatys}}
        \label{fig:image1}
    \end{minipage}
    \hspace{0.2cm}
    \begin{minipage}{0.3\textwidth}
        \centering
        \includegraphics[width=\textwidth]{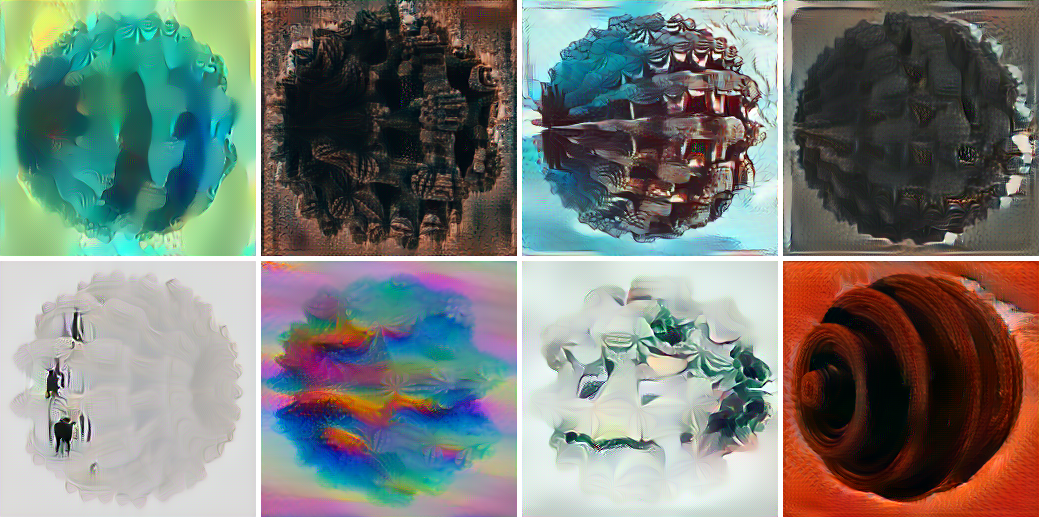}
        \subcaption{Mandelbulb-\textit{Gatys}}
        \label{fig:image3}
    \end{minipage}    
    \caption{Samples from different \textit{Gatys}-stylized synthetic datasets.}
    \label{fig:gatys_grid}
\end{figure*}

\textbf{Reverse Stylization with the Gatys Algorithm:}
The images shown in \autoref{fig:gatys_grid} are reverse-stylized synthetic images generated using the Gatys algorithm.

\begin{figure*}[htb]
\centering
\begin{tikzpicture}
    \begin{axis}[
        ybar,
        width=\textwidth, % Scale plot with the image width
        height=8.2cm,
        ymin=0,
        ymax=0.7,
        ylabel={Reconstruction Loss},
        xtick=\empty, % Remove x ticks
        symbolic x coords={Dataset1, Dataset2, Dataset3, Dataset4, Dataset5, Dataset6, Dataset7, Dataset8, Dataset9, Dataset10, Dataset11, Dataset12, Dataset13, Dataset14, Dataset15, Dataset16, Dataset17}, % Dynamically handle x coords
        xticklabels={}, % Remove x labels
        bar width=0.7cm, % Dynamic width scaling with linewidth
        enlarge x limits={abs=1cm}, % Dynamic spacing
        nodes near coords, % Enable nodes near coordinates for dataset names
        every node near coord/.append style={rotate=90, anchor=west, font=\normalsize}, % Rotate dataset names inside bars
        grid=major,
        major grid style={dotted},
        axis line style={-},
    ]
    
    % Bar plot with different colors and labels inside the bars
    \addplot[
        ybar,
        fill=yellow,
        nodes near coords={\text{Mandelbulb}}, % Dataset name inside the bar
        every node near coord/.append style={anchor=east, font=\normalsize},
        bar shift=0pt
    ] coordinates {(Dataset1,0.616)};
    
    \addplot[
        ybar,
        fill=yellow,
        nodes near coords={\text{VisualAtom}}, % Dataset name inside the bar
        every node near coord/.append style={anchor=east, font=\normalsize},
        bar shift=0pt,
    ] coordinates {(Dataset2,0.32)};

    \addplot[
        ybar,
        fill=yellow,
        nodes near coords={\text{FractalDB-composite}}, % Dataset name inside the bar
        bar shift=0pt
    ] coordinates {(Dataset3,0.224)};

    \addplot[
        ybar,
        fill=black,
        nodes near coords={\text{\textit{Unsplash}}}, % Dataset name inside the bar
        every node near coord/.append style={text=black}, % Change text color to white
        bar shift=0pt
    ] coordinates {(Dataset4,0.153)};

    \addplot[
        ybar,
        fill=yellow,
        nodes near coords={\text{StyleGAN-oriented}}, % Dataset name inside the bar
        bar shift=0pt
    ] coordinates {(Dataset5,0.138)};
    
    \addplot[
        ybar,
        fill=yellow,
        nodes near coords={\text{Neural Fractal}}, % Dataset name inside the bar
        bar shift=0pt
    ] coordinates {(Dataset6,0.137)};

    \addplot[
        ybar,
        fill=orange,
        nodes near coords={\text{Mandelbulb-\textit{NNST}}}, % Dataset name inside the bar
        bar shift=0pt
    ] coordinates {(Dataset7,0.13)};
    
    \addplot[
        ybar,
        fill=orange,
        nodes near coords={\text{VisualAtom-\textit{NNST}}}, % Dataset name inside the bar
        bar shift=0pt
    ] coordinates {(Dataset8,0.126)};

    \addplot[
        ybar,
        fill=cyan,
        nodes near coords={\text{Mandelbulb-\textit{Gatys}}}, % Dataset name inside the bar
        bar shift=0pt
    ] coordinates {(Dataset9,0.115)};

    \addplot[
        ybar,
        fill=orange,
        nodes near coords={\text{StyleGAN-oriented-\textit{NNST}}}, % Dataset name inside the bar
        bar shift=0pt
    ] coordinates {(Dataset10,0.112)};

    \addplot[
        ybar,
        fill=orange,
        nodes near coords={\text{FractalDB-composite-\textit{NNST}}}, % Dataset name inside the bar
        bar shift=0pt
    ] coordinates {(Dataset11,0.112)};

    \addplot[
        ybar,
        fill=cyan,
        nodes near coords={\text{VisualAtom-\textit{Gatys}}}, % Dataset name inside the bar
        bar shift=0pt
    ] coordinates {(Dataset12,0.111)};

    \addplot[
        ybar,
        fill=orange,
        nodes near coords={\text{Neural Fractal-\textit{NNST}}}, % Dataset name inside the bar
        bar shift=0pt
    ] coordinates {(Dataset13,0.111)};

    \addplot[
        ybar,
        fill=cyan,
        nodes near coords={\text{StyleGAN-oriented-\textit{Gatys}}}, % Dataset name inside the bar
        every node near coord/.append style={font=\small},
        bar shift=0pt
    ] coordinates {(Dataset14,0.107)};

    \addplot[
        ybar,
        fill=cyan,
        nodes near coords={\text{FractalDB-composite-\textit{Gatys}}}, % Dataset name inside the bar
        every node near coord/.append style={font=\small},
        bar shift=0pt
    ] coordinates {(Dataset15,0.106)};
    
    \addplot[
        ybar,
        fill=cyan,
        nodes near coords={\text{Neural Fractal-\textit{Gatys}}}, % Dataset name inside the bar
        bar shift=0pt
    ] coordinates {(Dataset16,0.105)};

    \addplot[
        ybar,
        fill=black,
        nodes near coords={\text{ImageNet-100}}, % Dataset name inside the bar
        every node near coord/.append style={text=black}, % Change text color to white
        bar shift=0pt
    ] coordinates {(Dataset17,0.09)};

    \draw[dashed, red] (rel axis cs:0,0.21857) -- (rel axis cs:1,0.21857);
    \draw[dashed, red] (rel axis cs:0,0.12857) -- (rel axis cs:1,0.12857);
    
    \end{axis}

    % Legend
    \begin{scope}[shift={(rel axis cs:0.9035,0.855)}] % Move legend to the right
        \node[draw, rectangle, fill=white] (legend) at (0,0) {%
            \begin{tabular}{rl}
                \textcolor{black}{\rule{2em}{0.5em}} & Baseline real datasets \\
                \textcolor{yellow}{\rule{2em}{0.5em}} & Base synthetic datasets \\
                \textcolor{cyan}{\rule{2em}{0.5em}} & \textit{Gatys}-stylized datasets \\
                \textcolor{orange}{\rule{2em}{0.5em}} & \textit{NNST}-stylized datasets \\
            \end{tabular}
        };
    \end{scope}
    
    % Adding images below bars
    \node[anchor=north] at (rel axis cs:0.125,0) {\includegraphics[width=0.7cm]{figures/sample_Mandelbulb.png}};
    \node[anchor=north] at (rel axis cs:0.1796875,0) {\includegraphics[width=0.7cm]{figures/sample_VisualAtom.png}};
    \node[anchor=north] at (rel axis cs:0.234375,0) {\includegraphics[width=0.7cm]{figures/sample_FractalDB++.png}};
    \node[anchor=north] at (rel axis cs:0.2890625,0) {\includegraphics[width=0.7cm]{figures/sample_unsplash.png}};
    \node[anchor=north] at (rel axis cs:0.34375,0) {\includegraphics[width=0.7cm]{figures/sample_StyleGAN_oriented.png}};
    \node[anchor=north] at (rel axis cs:0.3984375,0) {\includegraphics[width=0.7cm]{figures/sample_neural_fractal_colored.png}};
    \node[anchor=north] at (rel axis cs:0.453125,0) {\includegraphics[width=0.7cm]{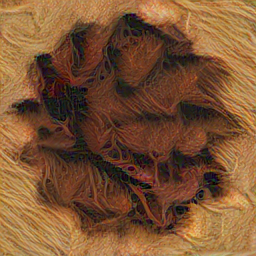}};
    \node[anchor=north] at (rel axis cs:0.5078125,0) {\includegraphics[width=0.7cm]{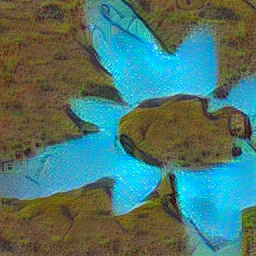}};
    \node[anchor=north] at (rel axis cs:0.5625,0) {\includegraphics[width=0.7cm]{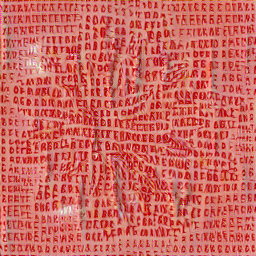}};
    \node[anchor=north] at (rel axis cs:0.6171875,0) {\includegraphics[width=0.7cm]{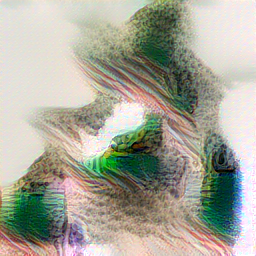}};
    \node[anchor=north] at (rel axis cs:0.671875,0) {\includegraphics[width=0.7cm]{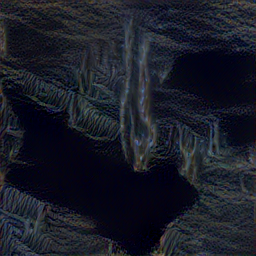}};
    \node[anchor=north] at (rel axis cs:0.7265625,0) {\includegraphics[width=0.7cm]{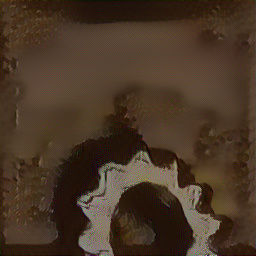}};
    \node[anchor=north] at (rel axis cs:0.78125,0) {\includegraphics[width=0.7cm]{figures/sample_unsplash_to_neural_fractal_colored_nnst.png}};
    \node[anchor=north] at (rel axis cs:0.8359375,0) {\includegraphics[width=0.7cm]{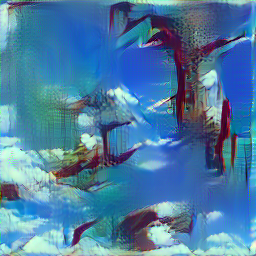}};
    \node[anchor=north] at (rel axis cs:0.890625,0) {\includegraphics[width=0.7cm]{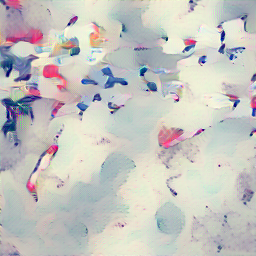}};
    \node[anchor=north] at (rel axis cs:0.9453125,0) {\includegraphics[width=0.7cm]{figures/sample_unsplash_to_neural_fractal_colored_gatys.png}};
    \node[anchor=north] at (rel axis cs:1,0) {\includegraphics[width=0.7cm]{figures/sample_imagenet.png}};
\end{tikzpicture}
\caption{Autoencoder reconstruction loss for different datasets. Datasets are sorted in increasing order by performance along the x-axis. The dashed lines represent the lower and upper bounds in performance that can be achieved using small and large real datasets respectively.}
\label{fig:compare_autoencoder}
\end{figure*}

\textbf{Performance of Synthetic Datasets in the Autoencoder Experiment:}
Below, we provide some additional figures that summarize and compare the performance of the different datasets that we experiment with.
\autoref{fig:compare_autoencoder} compares the performance of different synthetic datasets in the autoencoder experiment. 
We observe that non-stylized neural fractal outperforms other non-stylized synthetic datasets. 
Additionally, we find that stylization improves the accuracy of all base synthetic datasets. In this experiment, the \textit{Gatys} stylization method performs better than \textit{NNST}. Among the stylized datasets, neural fractal once again outperforms the others, achieving the lowest reconstruction error.

\begin{figure*}[htb]
\centering
\begin{tikzpicture}
    \begin{axis}[
        ybar,
        width=\textwidth, % Scale plot with the image width
        height=8.2cm,
        ymin=0,
        ymax=90,
        ylabel={Accuracy (\%)},
        xtick=\empty, % Remove x ticks
        symbolic x coords={Dataset1, Dataset2, Dataset3, Dataset4, Dataset5, Dataset6, Dataset7, Dataset8, Dataset9, Dataset10, Dataset11, Dataset12, Dataset13, Dataset14, Dataset15, Dataset16, Dataset17}, % Dynamically handle x coords
        xticklabels={}, % Remove x labels
        bar width=0.7cm, % Dynamic width scaling with linewidth
        enlarge x limits={abs=1cm}, % Dynamic spacing
        nodes near coords, % Enable nodes near coordinates for dataset names
        every node near coord/.append style={rotate=90, anchor=east, font=\normalsize}, % Rotate dataset names inside bars
        grid=major,
        major grid style={dotted},
        axis line style={-},
    ]
    
    % Bar plot with different colors and labels inside the bars
    \addplot[
        ybar,
        fill=yellow,
        nodes near coords={\text{VisualAtom}}, % Dataset name inside the bar
        bar shift=0pt,
    ] coordinates {(Dataset1,38.92)};

    \addplot[
        ybar,
        fill=yellow,
        nodes near coords={\text{Mandelbulb}}, % Dataset name inside the bar
        bar shift=0pt
    ] coordinates {(Dataset2,44.02)};

    \addplot[
        ybar,
        fill=black,
        nodes near coords={\text{\textit{Unsplash}}}, % Dataset name inside the bar
        every node near coord/.append style={text=white}, % Change text color to white
        bar shift=0pt
    ] coordinates {(Dataset3,44.8)};

    \addplot[
        ybar,
        fill=yellow,
        nodes near coords={\text{FractalDB-composite}}, % Dataset name inside the bar
        bar shift=0pt
    ] coordinates {(Dataset4,45.32)};

    \addplot[
        ybar,
        fill=yellow,
        nodes near coords={\text{StyleGAN-oriented}}, % Dataset name inside the bar
        bar shift=0pt
    ] coordinates {(Dataset5,48.06)};
    
    \addplot[
        ybar,
        fill=yellow,
        nodes near coords={\text{Neural Fractal}}, % Dataset name inside the bar
        bar shift=0pt
    ] coordinates {(Dataset6,48.48)};
    
    \addplot[
        ybar,
        fill=cyan,
        nodes near coords={\text{VisualAtom-\textit{Gatys}}}, % Dataset name inside the bar
        bar shift=0pt
    ] coordinates {(Dataset7,55.6)};
    
    \addplot[
        ybar,
        fill=orange,
        nodes near coords={\text{VisualAtom-\textit{NNST}}}, % Dataset name inside the bar
        bar shift=0pt
    ] coordinates {(Dataset8,55.8)};

    \addplot[
        ybar,
        fill=cyan,
        nodes near coords={\text{StyleGAN-oriented-\textit{Gatys}}}, % Dataset name inside the bar
        bar shift=0pt
    ] coordinates {(Dataset9,55.92)};
    
    \addplot[
        ybar,
        fill=cyan,
        nodes near coords={\text{FractalDB-composite-\textit{Gatys}}}, % Dataset name inside the bar
        bar shift=0pt
    ] coordinates {(Dataset10,56.14)};
    
    \addplot[
        ybar,
        fill=cyan,
        nodes near coords={\text{Mandelbulb-\textit{Gatys}}}, % Dataset name inside the bar
        bar shift=0pt
    ] coordinates {(Dataset11,56.76)};
    
    \addplot[
        ybar,
        fill=orange,
        nodes near coords={\text{Mandelbulb-\textit{NNST}}}, % Dataset name inside the bar
        bar shift=0pt
    ] coordinates {(Dataset12,57.1)};

    \addplot[
        ybar,
        fill=cyan,
        nodes near coords={\text{Neural Fractal-\textit{Gatys}}}, % Dataset name inside the bar
        bar shift=0pt
    ] coordinates {(Dataset13,57.38)};
    
    \addplot[
        ybar,
        fill=orange,
        nodes near coords={\text{FractalDB-composite-\textit{NNST}}}, % Dataset name inside the bar
        bar shift=0pt
    ] coordinates {(Dataset14,57.76)};
    
    \addplot[
        ybar,
        fill=orange,
        nodes near coords={\text{StyleGAN-oriented-\textit{NNST}}}, % Dataset name inside the bar
        bar shift=0pt
    ] coordinates {(Dataset15,58.92)};
    
    \addplot[
        ybar,
        fill=orange,
        nodes near coords={\text{Neural Fractal-\textit{NNST}}}, % Dataset name inside the bar
        bar shift=0pt
    ] coordinates {(Dataset16,59.18)};

    \addplot[
        ybar,
        fill=black,
        nodes near coords={\text{ImageNet-100}}, % Dataset name inside the bar
        every node near coord/.append style={text=white}, % Change text color to white
        bar shift=0pt
    ] coordinates {(Dataset17,78.2)};

    \draw[dashed, red] (rel axis cs:0,0.497777) -- (rel axis cs:1,0.497777);
    \draw[dashed, red] (rel axis cs:0,0.868888) -- (rel axis cs:1,0.868888);
    
    \end{axis}

    % Legend
    \begin{scope}[shift={(rel axis cs:0.2225,0.855)}] % Move legend to the right
        \node[draw, rectangle, fill=white] (legend) at (0,0) {%
            \begin{tabular}{rl}
                \textcolor{black}{\rule{2em}{0.5em}} & Baseline real datasets \\
                \textcolor{yellow}{\rule{2em}{0.5em}} & Base synthetic datasets \\
                \textcolor{cyan}{\rule{2em}{0.5em}} & \textit{Gatys}-stylized datasets \\
                \textcolor{orange}{\rule{2em}{0.5em}} & \textit{NNST}-stylized datasets \\
            \end{tabular}
        };
    \end{scope}
    
    % Adding images below bars
    \node[anchor=north] at (rel axis cs:0.125,0) {\includegraphics[width=0.7cm]{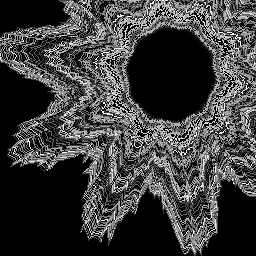}};
    \node[anchor=north] at (rel axis cs:0.1796875,0) {\includegraphics[width=0.7cm]{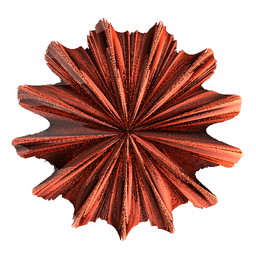}};
    \node[anchor=north] at (rel axis cs:0.234375,0) {\includegraphics[width=0.7cm]{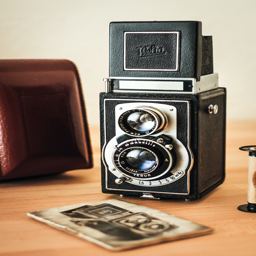}};
    \node[anchor=north] at (rel axis cs:0.2890625,0) {\includegraphics[width=0.7cm]{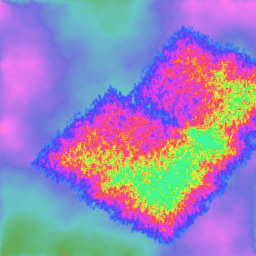}};
    \node[anchor=north] at (rel axis cs:0.34375,0) {\includegraphics[width=0.7cm]{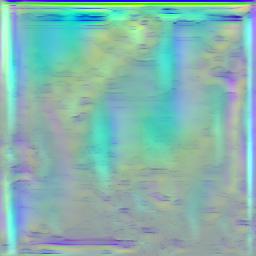}};
    \node[anchor=north] at (rel axis cs:0.3984375,0) {\includegraphics[width=0.7cm]{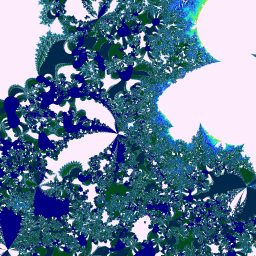}};
    \node[anchor=north] at (rel axis cs:0.453125,0) {\includegraphics[width=0.7cm]{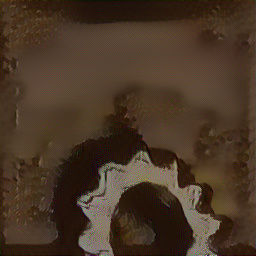}};
    \node[anchor=north] at (rel axis cs:0.5078125,0) {\includegraphics[width=0.7cm]{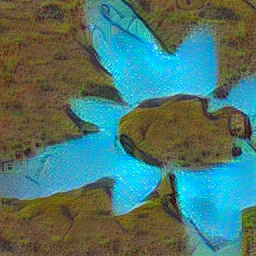}};
    \node[anchor=north] at (rel axis cs:0.5625,0) {\includegraphics[width=0.7cm]{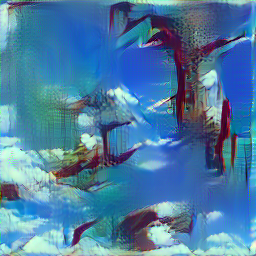}};
    \node[anchor=north] at (rel axis cs:0.6171875,0) {\includegraphics[width=0.7cm]{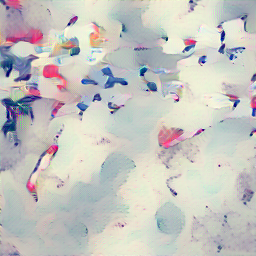}};
    \node[anchor=north] at (rel axis cs:0.671875,0) {\includegraphics[width=0.7cm]{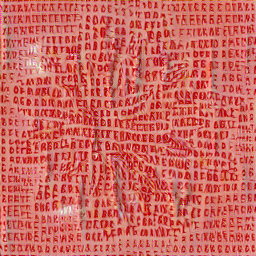}};
    \node[anchor=north] at (rel axis cs:0.7265625,0) {\includegraphics[width=0.7cm]{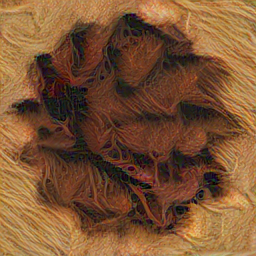}};
    \node[anchor=north] at (rel axis cs:0.78125,0) {\includegraphics[width=0.7cm]{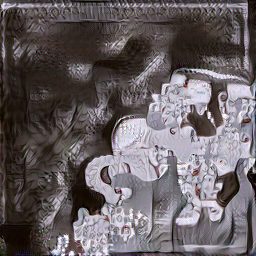}};
    \node[anchor=north] at (rel axis cs:0.8359375,0) {\includegraphics[width=0.7cm]{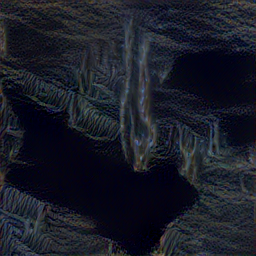}};
    \node[anchor=north] at (rel axis cs:0.890625,0) {\includegraphics[width=0.7cm]{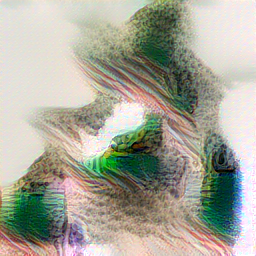}};
    \node[anchor=north] at (rel axis cs:0.9453125,0) {\includegraphics[width=0.7cm]{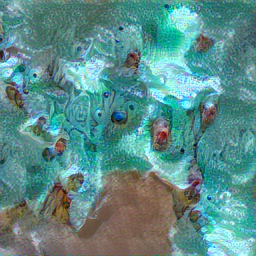}};
    \node[anchor=north] at (rel axis cs:1,0) {\includegraphics[width=0.7cm]{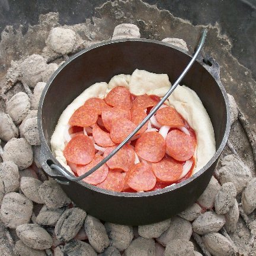}};
\end{tikzpicture}
\caption{Top-1 accuracy on ImageNet-100 for different datasets evaluated using the DINO-based evaluation pipeline~\cite{takashima2023visual}. Datasets are sorted in increasing order by performance along the x-axis. The dashed lines represent the lower and upper bounds in performance that can be achieved using small and large real datasets respectively.}
\label{fig:compare_dino}
\end{figure*}

% \autoref{fig:compare_alexnet} shows the results for the contrastive learning experiment with AlexNet. As described in Section~\ref{sec:exp_nfractal}, neural Fractal performs poorly when used to train the CNN-based AlexNet. We hypothesize that this is because the Neural Fractal dataset contains a high level of fine-grained, high-frequency details, which challenges AlexNet—a relatively small CNN model—especially in a contrastive training setting. Similar findings were reported by Nakashima et al.~\cite{nakashima2022can}, who observed that FractalDB yielded better performance with ViT compared to CNN-based models. \autoref{fig:compare_alexnet} shows also the effectiveness of stylization to improve the base synthetic dataset. 
% 
\textbf{Performance of Synthetic Datasets in the Contrastive Learning Experiment:}
\autoref{fig:compare_dino} presents the results of the contrastive learning experiment using DINO \cite{caron2021emerging}. DINO is a modern and expressive architecture designed for unsupervised settings, offering greater flexibility and performance. In this experiment, neural fractal achieves higher accuracy with and without stylization. The \textit{NNST} stylization method outperforms \textit{Gatys} in this setup.

\begin{figure*}[htb]
\centering
\begin{tikzpicture}
    \begin{axis}[
        ybar,
        width=\textwidth, % Scale plot with the image width
        height=8.2cm,
        ymin=0,
        ymax=10,
        ylabel={FID},
        xtick=\empty, % Remove x ticks
        symbolic x coords={Dataset1, Dataset2, Dataset3, Dataset4, Dataset5, Dataset6, Dataset7, Dataset8, Dataset9, Dataset10, Dataset11, Dataset12}, % Dynamically handle x coords
        xticklabels={}, % Remove x labels
        bar width=0.99cm, % Dynamic width scaling with linewidth
        enlarge x limits={abs=1cm}, % Dynamic spacing
        nodes near coords, % Enable nodes near coordinates for dataset names
        every node near coord/.append style={rotate=90, anchor=west, font=\normalsize}, % Rotate dataset names inside bars
        grid=major,
        major grid style={dotted},
        axis line style={-},
        clip=false, % Prevent clipping of images
    ]
    
    % Bar plot with different colors and labels inside the bars
    %% Unsplash 9.05
    \addplot[
        ybar,
        fill=cyan,
        nodes near coords={\text{\textit{Unsplash}}},
        every node near coord/.append style={anchor=east, text=black, font=\small},
        bar shift=0pt
    ] coordinates {(Dataset1,9.05)};

    %% Without Finetuning 8.99
    \addplot[
        ybar,
        fill=cyan,
        nodes near coords={\text{\textit{Without Finetuning}}},
        every node near coord/.append style={anchor=east, text=black, font=\small},
        bar shift=0pt
    ] coordinates {(Dataset2,8.99)};

    %% Mandelbulb 8.48
    \addplot[
        ybar,
        fill=yellow,
        nodes near coords={\text{Mandelbulb}},
        every node near coord/.append style={anchor=east, font=\small},
        bar shift=0pt
    ] coordinates {(Dataset3,8.48)};

    %% Fractal DB composite 8.43
    \addplot[
        ybar,
        fill=yellow,
        nodes near coords={\text{FractalDB-composite}},
        every node near coord/.append style={anchor=east, font=\small},
        bar shift=0pt
    ] coordinates {(Dataset4,8.43)};

    %% Visual Atom 8.28
    \addplot[
        ybar,
        fill=yellow,
        nodes near coords={\text{VisualAtom}},
        every node near coord/.append style={anchor=east, font=\small},
        bar shift=0pt,
    ] coordinates {(Dataset5,8.28)};

    %% StyleGAN-oriented 8.08
    \addplot[
        ybar,
        fill=yellow,
        nodes near coords={\text{StyleGAN-oriented}},
        every node near coord/.append style={anchor=east, font=\small},
        bar shift=0pt
    ] coordinates {(Dataset6,8.08)};

    %% Neural Fractal-100k 7.39
    \addplot[
        ybar,
        fill=yellow,
        nodes near coords={\text{Neural Fractal-100k}},
        every node near coord/.append style={anchor=east, font=\small},
        bar shift=0pt
    ] coordinates {(Dataset7,7.39)};

    %% Neural Fractal-100k NNST 7.34
    \addplot[
        ybar,
        fill=orange,
        nodes near coords={\text{Neural Fractal-100k-\textit{NNST}}},
        every node near coord/.append style={anchor=east, font=\small},
        bar shift=0pt
    ] coordinates {(Dataset8,7.34)};

    %% ImageNet-100k 7.21
    \addplot[
        ybar,
        fill=cyan,
        nodes near coords={\text{ImageNet-100k}},
        every node near coord/.append style={anchor=east, font=\small},
        bar shift=0pt
    ] coordinates {(Dataset9,7.21)};

    %% Neural Fractal-1m 6.52
    \addplot[
        ybar,
        fill=yellow,
        nodes near coords={\text{Neural Fractal-1M}},
        every node near coord/.append style={anchor=east, font=\small},
        bar shift=0pt
    ] coordinates {(Dataset10,6.52)};

    %% Neural Fractal NNST-1m 5.47
    \addplot[
        ybar,
        fill=orange,
        nodes near coords={\text{Neural Fractal-1M-\textit{NNST}}},
        every node near coord/.append style={anchor=east, font=\small},
        bar shift=0pt
    ] coordinates {(Dataset11,5.47)};

    %% ImageNet-1m 4.94
    \addplot[
        ybar,
        fill=cyan,
        nodes near coords={\text{ImageNet-1M}},
        every node near coord/.append style={anchor=east, font=\small},
        bar shift=0pt
    ] coordinates {(Dataset12,4.94)};

    \draw[dashed, red] (rel axis cs:0,0.494) -- (rel axis cs:1,0.494);
    \draw[dashed, red] (rel axis cs:0,0.905) -- (rel axis cs:1,0.905);

    % Place images below bars using (axis cs:...) coordinates
    \node[anchor=north] at (axis cs:Dataset1,0) {\includegraphics[width=0.99cm]{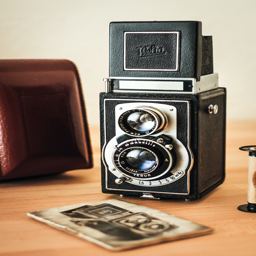}};
    % \node[anchor=north] at (axis cs:Dataset2,0) {\includegraphics[width=0.99cm]{figures/sample_VisualAtom.png}};
    \node[anchor=north] at (axis cs:Dataset3,0) {\includegraphics[width=0.99cm]{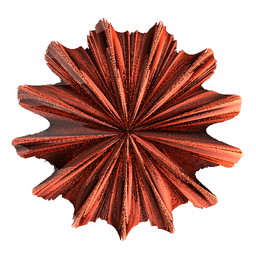}};
    \node[anchor=north] at (axis cs:Dataset4,0) {\includegraphics[width=0.99cm]{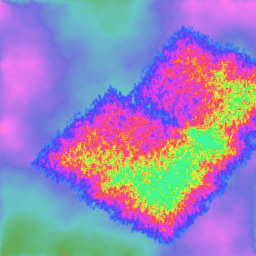}};
    \node[anchor=north] at (axis cs:Dataset5,0) {\includegraphics[width=0.99cm]{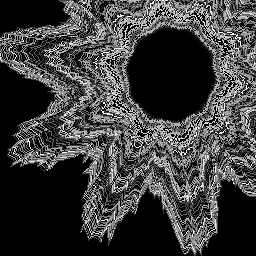}};
    \node[anchor=north] at (axis cs:Dataset6,0) {\includegraphics[width=0.99cm]{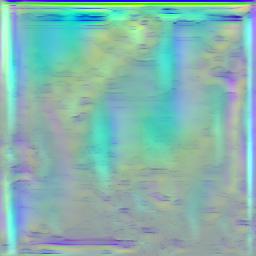}};
    \node[anchor=north] at (axis cs:Dataset7,0) {\includegraphics[width=0.99cm]{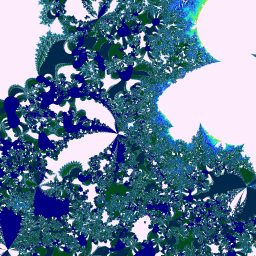}};
    \node[anchor=north] at (axis cs:Dataset8,0) {\includegraphics[width=0.99cm]{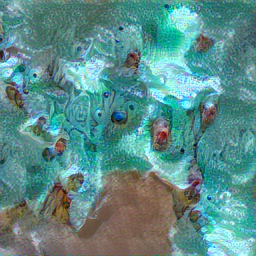}};
    \node[anchor=north] at (axis cs:Dataset9,0) {\includegraphics[width=0.99cm]{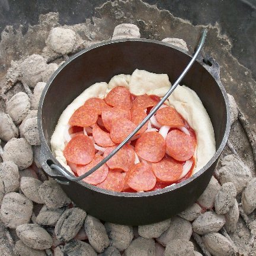}};
    \node[anchor=north] at (axis cs:Dataset10,0) {\includegraphics[width=0.99cm]{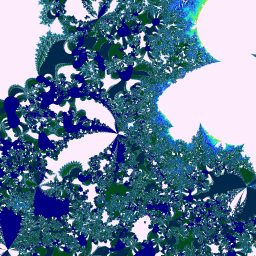}};
    \node[anchor=north] at (axis cs:Dataset11,0) {\includegraphics[width=0.99cm]{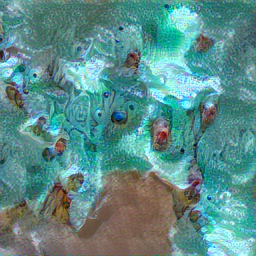}};
    \node[anchor=north] at (axis cs:Dataset12,0) {\includegraphics[width=0.99cm]{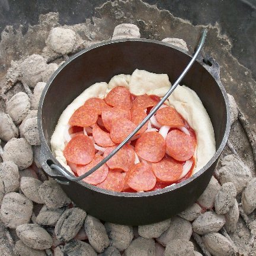}};

    \end{axis}

    % Legend
    \begin{scope}[shift={(rel axis cs:0.9035,0.855)}] % Move legend to the right
        \node[draw, rectangle, fill=white] (legend) at (0,0) {%
            \begin{tabular}{rl}
                \textcolor{cyan}{\rule{2em}{0.5em}} & Baseline real datasets \\
                \textcolor{yellow}{\rule{2em}{0.5em}} & Base synthetic datasets \\
                \textcolor{orange}{\rule{2em}{0.5em}} & \textit{NNST}-stylized datasets \\
            \end{tabular}
        };
    \end{scope}
    
\end{tikzpicture}
\caption{FID scores for models trained on the FFHQ dataset, illustrating the impact of different pretraining datasets. Models are grouped by dataset type: baseline real datasets (blue), base synthetic datasets (yellow), and NNST-stylized datasets (orange). Lower FID scores indicate better performance, with models sorted from highest to lowest FID. Sample images from each dataset are shown below the bars.}
\label{fig:compare_fid_ffhq}
\end{figure*}

\begin{figure*}[htb]
\centering
\begin{tikzpicture}
    \begin{axis}[
        ybar,
        width=\textwidth, % Scale plot with the image width
        height=8.2cm,
        ymin=0,
        ymax=160,
        ylabel={FID},
        xtick=\empty, % Remove x ticks
        symbolic x coords={Dataset1, Dataset2, Dataset3, Dataset4, Dataset5, Dataset6, Dataset7, Dataset8, Dataset9, Dataset10, Dataset11, Dataset12}, % Dynamically handle x coords
        xticklabels={}, % Remove x labels
        bar width=0.99cm, % Dynamic width scaling with linewidth
        enlarge x limits={abs=1cm}, % Dynamic spacing
        nodes near coords, % Enable nodes near coordinates for dataset names
        every node near coord/.append style={rotate=90, anchor=west, font=\normalsize}, % Rotate dataset names inside bars
        grid=major,
        major grid style={dotted},
        axis line style={-},
        clip=false, % Prevent clipping of images
    ]
    
    % Bar plot with different colors and labels inside the bars

    %% Without Finetuning 158.59
    \addplot[
        ybar,
        fill=cyan,
        nodes near coords={\text{\textit{Without Finetuning}}},
        every node near coord/.append style={anchor=east, text=black, font=\small},
        bar shift=0pt
    ] coordinates {(Dataset1,158.59)};

    %% Visual Atom 43.09
    \addplot[
        ybar,
        fill=yellow,
        nodes near coords={\text{VisualAtom}},
        every node near coord/.append style={font=\small},
        bar shift=0pt,
    ] coordinates {(Dataset2,43.09)};

    %% Unsplash 29.00
    \addplot[
        ybar,
        fill=cyan,
        nodes near coords={\text{\textit{Unsplash}}},
        every node near coord/.append style={text=black, font=\small},
        bar shift=0pt
    ] coordinates {(Dataset3,29.00)};

    %% Mandelbulb 28.83
    \addplot[
        ybar,
        fill=yellow,
        nodes near coords={\text{Mandelbulb}},
        every node near coord/.append style={font=\small},
        bar shift=0pt
    ] coordinates {(Dataset4,28.83)};

    %% StyleGAN-oriented 20.05
    \addplot[
        ybar,
        fill=yellow,
        nodes near coords={\text{StyleGAN-oriented}},
        every node near coord/.append style={font=\small},
        bar shift=0pt
    ] coordinates {(Dataset5,20.05)};
    
    %% Fractal DB composite 18.85
    \addplot[
        ybar,
        fill=yellow,
        nodes near coords={\text{FractalDB-composite}},
        every node near coord/.append style={font=\small},
        bar shift=0pt
    ] coordinates {(Dataset6,18.85)};

    %% Neural Fractal-100k 18.37
    \addplot[
        ybar,
        fill=yellow,
        nodes near coords={\text{Neural Fractal-100k}},
        every node near coord/.append style={font=\small},
        bar shift=0pt
    ] coordinates {(Dataset7,18.37)};
    
    %% Neural Fractal-100k NNST 16.34
    \addplot[
        ybar,
        fill=orange,
        nodes near coords={\text{Neural Fractal-100k-\textit{NNST}}},
        every node near coord/.append style={font=\small},
        bar shift=0pt
    ] coordinates {(Dataset8,16.34)};
    
    %% Neural Fractal-1m 15.64
    \addplot[
        ybar,
        fill=yellow,
        nodes near coords={\text{Neural Fractal-1M}},
        every node near coord/.append style={font=\small},
        bar shift=0pt
    ] coordinates {(Dataset9,15.64)};

    %% Neural Fractal NNST-1m 10.19
    \addplot[
        ybar,
        fill=orange,
        nodes near coords={\text{Neural Fractal-1M-\textit{NNST}}},
        every node near coord/.append style={font=\small},
        bar shift=0pt
    ] coordinates {(Dataset10,10.19)};
    
    %% ImageNet-100k 8.85
    \addplot[
        ybar,
        fill=cyan,
        nodes near coords={\text{ImageNet-100k}},
        every node near coord/.append style={font=\small},
        bar shift=0pt
    ] coordinates {(Dataset11,8.85)};
    
    %% ImageNet-1m 6.89
    \addplot[
        ybar,
        fill=cyan,
        nodes near coords={\text{ImageNet-1M}},
        every node near coord/.append style={font=\small},
        bar shift=0pt
    ] coordinates {(Dataset12,6.89)};

    \draw[dashed, red] (rel axis cs:0,0.0430625) -- (rel axis cs:1,0.0430625);
    \draw[dashed, red] (rel axis cs:0,0.18125) -- (rel axis cs:1,0.18125);

    % Place images below bars using (axis cs:...) coordinates
    % \node[anchor=north] at (axis cs:Dataset1,0) {\includegraphics[width=0.99cm]{figures/sample_VisualAtom.png}};
    \node[anchor=north] at (axis cs:Dataset2,0) {\includegraphics[width=0.99cm]{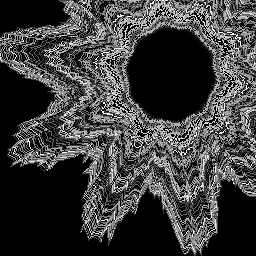}};
    \node[anchor=north] at (axis cs:Dataset3,0) {\includegraphics[width=0.99cm]{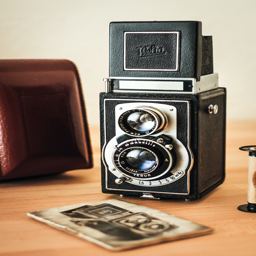}};
    \node[anchor=north] at (axis cs:Dataset4,0) {\includegraphics[width=0.99cm]{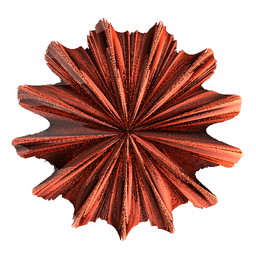}};
    \node[anchor=north] at (axis cs:Dataset5,0) {\includegraphics[width=0.99cm]{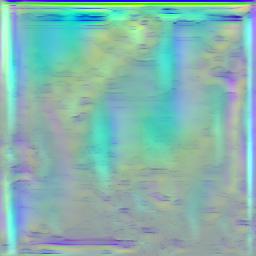}};
    \node[anchor=north] at (axis cs:Dataset6,0) {\includegraphics[width=0.99cm]{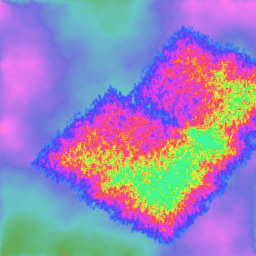}};
    \node[anchor=north] at (axis cs:Dataset7,0) {\includegraphics[width=0.99cm]{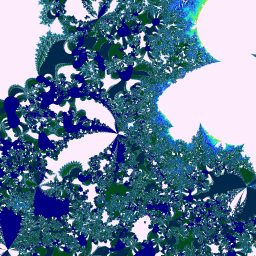}};
    \node[anchor=north] at (axis cs:Dataset8,0) {\includegraphics[width=0.99cm]{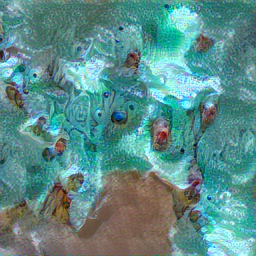}};
    \node[anchor=north] at (axis cs:Dataset9,0) {\includegraphics[width=0.99cm]{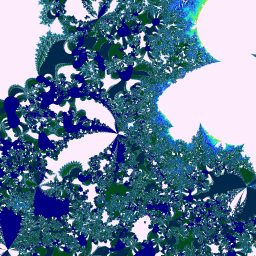}};
    \node[anchor=north] at (axis cs:Dataset10,0) {\includegraphics[width=0.99cm]{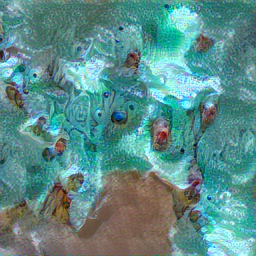}};
    \node[anchor=north] at (axis cs:Dataset11,0) {\includegraphics[width=0.99cm]{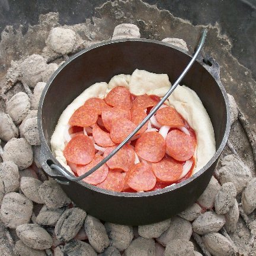}};
    \node[anchor=north] at (axis cs:Dataset12,0) {\includegraphics[width=0.99cm]{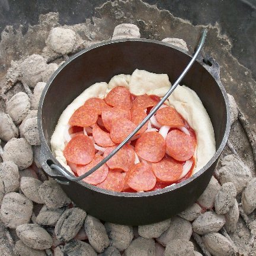}};

    \end{axis}

    % Legend
    \begin{scope}[shift={(rel axis cs:0.9035,0.855)}] % Move legend to the right
        \node[draw, rectangle, fill=white] (legend) at (0,0) {%
            \begin{tabular}{rl}
                \textcolor{cyan}{\rule{2em}{0.5em}} & Baseline real datasets \\
                \textcolor{yellow}{\rule{2em}{0.5em}} & Base synthetic datasets \\
                \textcolor{orange}{\rule{2em}{0.5em}} & \textit{NNST}-stylized datasets \\
            \end{tabular}
        };
    \end{scope}
    
\end{tikzpicture}
\caption{FID scores for models trained on the Flowers dataset, highlighting the influence of different pretraining datasets. Models are categorized into baseline real datasets (blue), base synthetic datasets (yellow), and NNST-stylized datasets (orange). Lower FID scores represent improved performance, with models sorted from highest to lowest FID. Sample images from each dataset are displayed beneath the bars.}
\label{fig:compare_fid_flowers}
\end{figure*}

\textbf{Performance of Synthetic Datasets in the Diffusion Experiment:}
\autoref{fig:compare_fid_ffhq} presents the results of the diffusion experiment on the FFHQ dataset. 
In this experiment, pretraining with neural fractal achieves the lowest FID after fine-tuning. 
Stylization further enhances the performance of the neural fractal dataset. 
Although we did not conduct experiments with other stylized synthetic datasets, we expect similar improvements with stylization. 
We also evaluate the 1M version of the neural fractal dataset, which consists of 1 million images, and observe that stylization again leads to better performance.
\autoref{fig:compare_fid_flowers} shows the results of the diffusion experiment on the Flowers dataset, where a similar trend is observed.
\autoref{fig:model_comparison_generated_ffhq} and \autoref{fig:model_comparison_generated_flowers} present an uncurated set of generated samples. All models were initialized with the same random number generation seed.

\textbf{Latent Space Visualizations of Autoencoders Trained on Different Synthetic Datasets:}
To further understand the impact of different training datasets on the learned representations of autoencoders, we visualize the latent spaces in \autoref{fig:feature_visualization_latents}. This figure presents the latent representations of samples from the FFHQ and Flowers dataset. For each sample, we display the latent visualizations produced by autoencoders trained on eight different datasets: the input image, ImageNet-100k, VisualAtom, Mandelbulb, FractalDB, StyleGAN, neural fractal (non-stylized), and Stylized Neural Fractal. Upon examining the visualizations, it is evident that the latent representations of the Stylized Neural Fractal are most similar to those of the autoencoder trained on ImageNet-100k. Both exhibit a balanced level of structural detail and noise, unlike some other datasets where the latents appear either excessively noisy (e.g., StyleGAN) or overly structured (e.g., Mandelbulb). This similarity suggests that the stylization process effectively reduces the domain gap between the synthetic neural fractal dataset and real-world data, as represented by ImageNet-100k. While the presence of noise or clear structure in the latents is observable, we refrain from drawing conclusions about their quality or utility based solely on this visualization, as such properties are more directly tied to reconstruction performance (discussed in the main paper, e.g., reconstruction loss metrics). Instead, the key takeaway is that the reduced domain gap between Stylized Neural Fractal and ImageNet-100k may contribute to the improved performance observed in our experiments, highlighting the effectiveness of stylization in aligning synthetic and real data distributions.

\textbf{Measuring Distributional Distance between Synthetic and Real Datasets:}
We analyze the domain gap between our reverse-stylized Neural Fractals and real data by measuring both distribution-level and network-level similarities. The rightmost column of \autoref{tab:cosine_similarity3} reports the Kernel Inception Distance (KID) and Fréchet Inception Distance (FID) between each synthetic dataset and ImageNet, quantifying how closely their distributions align. KID is particularly robust in high-divergence regimes due to its unbiased estimation and ability to report confidence intervals. To compute KID, we use 100K images and 100 subsets; for FID, we use 50K images.
Neural fractal achieves the lowest KID and FID scores, indicating the smallest domain gap and best alignment with real data distributions.

Beyond distributional similarity, we compare how networks trained on different datasets attend to visual features by calculating the cosine similarity of attention maps. The first two columns of \autoref{tab:cosine_similarity3} show this similarity for the autoencoder (left) and the  diffusion model (middle), both using ImageNet-100k as the reference. Neural fractal consistently achieves the highest attention map similarity, suggesting that networks trained on it behave most similarly to those trained on ImageNet.

\begin{table*}[t]
    \renewcommand{\arraystretch}{0.8} % Reduce space between rows
    % \setlength{\tabcolsep}{6pt} % Reduce horizontal padding
    % % \small % Set a smaller font size
    \centering
    \begin{tabular}{lcc| c | cc}
        \toprule
        & \multicolumn{2}{c}{\textbf{AutoEncoding}$\uparrow$}  & \multicolumn{1}{c}{\textbf{ImageGen}$\uparrow$} & \multicolumn{2}{c}{\textbf{Domain Gap}$\downarrow$}\\ 
        \cmidrule(lr){2-3} \cmidrule(lr){4-4} \cmidrule(lr){5-6}
        \textbf{Train}/\textbf{Eval Dataset} & \textbf{ImgNet} & \textbf{COCO} & \textbf{ImgNet} & \textbf{KID}  & \textbf{FID} \\ 
        \midrule
        VisualAtom & 0.28 & 0.26 & 0.42 & 0.203 & 239.6\\
        Mandelbulb & 0.17 & 0.15 & 0.35 & 0.165 & 211.4\\
        FractalDB-composite & 0.50 & 0.49 & 0.51 & 0.214 & 207.6\\
        StyleGAN-oriented & 0.47 & 0.45 & 0.54 & 0.170 & 193.3\\
        \midrule
        NeuralFractal (ours) & \textbf{0.55} & \textbf{0.54} & \textbf{0.57} & \textbf{0.162} & \textbf{178.6}
        \\
        % \rowcolor{lightgray}
        \midrule
        +RevStylize (ours) & \textbf{0.68} & \textbf{0.67} & \textbf{0.61} & \textbf{0.120} & \textbf{148.5} \\
        \bottomrule
    \end{tabular}
    \caption{
    The first two columns show cosine similarity between attention maps of a network trained on ImageNet-100 and those trained on synthetic datasets. The left column corresponds to an autoencoder \cite{rombach2022high} trained for image reconstruction, and the middle column to EDM2 \cite{karras2024analyzing} trained with a diffusion loss. Neural fractal achieves the highest cosine similarity, suggesting that neural fractal trained network behaves more similar to the ImageNet trained network.
    The rightmost column reports two distribution divergence metrics:  Kernel Inception Distance (KID) and  Fréchet Inception Distance (FID).
    The metrics are computed between each synthetic dataset and ImageNet, measuring distribution-level similarity. 
    Neural fractal achieves the lowest KID and FID score, indicating the smallest domain gap among all evaluated datasets.}
\label{tab:cosine_similarity3}
\end{table*}

% \include{sec/supple_fid_figure_ffhq}

% \begin{figure}[t]
%    \centering
%    \includegraphics[width= 0.8\linewidth]{figures/new_samples/nnst_fractals_1m.png}
%    \caption{Samples from stylized fractals by nnst method}
%    \label{fig:suppl_nnst_fractal_1m}
% \end{figure}

% \begin{figure}[t]
%    \centering
%    \includegraphics[width= 0.8\linewidth]{figures/new_samples/gatys.png}
%    \caption{Samples from stylized fractals by gatys method}
%    \label{fig:suppl_gatys}
% \end{figure}

\input{sec/supple_flowers_results}

\input{sec/supple_ffhq_results}

\input{sec/supple_autoencoder_visualizations}

%% file: sec/tables/stylized_autoencoder_full.tex
\begin{table*}[t]
\renewcommand{\arraystretch}{0.8} % Reduce space between rows
\setlength{\tabcolsep}{6pt} % Reduce horizontal padding
\small % Set a smaller font size
% \resizebox{\linewidth}{!}{
\centering
    \begin{tabular}{lcccc}
        \toprule
        \textbf{Synthetic dataset} & \multicolumn{2}{c}{\textbf{ImageNet}} & \multicolumn{2}{c}{\textbf{COCO}}\\
        \cmidrule(lr){2-3} \cmidrule(lr){4-5}
        & \textit{Gatys} & \textit{NNST} & \textit{Gatys} & \textit{NNST}\\
        \midrule
        VisualAtom & 0.111 \textcolor{forestgreen}{(+65\%)} & 0.126 \textcolor{forestgreen}{(+61\%)} & 0.087 \textcolor{forestgreen}{(+71\%)} & 0.102 \textcolor{forestgreen}{(+66\%)}\\
        Mandelbulb & 0.115 \textcolor{forestgreen}{(+81\%)} & 0.130 \textcolor{forestgreen}{(+79\%)} & 0.100 \textcolor{forestgreen}{(+80\%)} & 0.117 \textcolor{forestgreen}{(+77\%)}\\
        FractalDB-composite & \underline{0.106} \textcolor{forestgreen}{(+53\%)} & \underline{0.112} \textcolor{forestgreen}{(+50\%)} & \textbf{0.084} \textcolor{forestgreen}{(+57\%)} & \underline{0.089} \textcolor{forestgreen}{(+54\%)}\\
        StyleGAN-oriented & 0.107 \textcolor{forestgreen}{(+22\%)} & \underline{0.112} \textcolor{forestgreen}{(+19\%)} & \textbf{0.084} \textcolor{forestgreen}{(+26\%)} & \underline{0.089} \textcolor{forestgreen}{(+22\%)}\\
        Neural Fractals & \textbf{0.105} \textcolor{forestgreen}{(+23\%)} & \textbf{0.111} \textcolor{forestgreen}{(+19\%)} & \textbf{0.084} \textcolor{forestgreen}{(+28\%)} & \textbf{0.088} \textcolor{forestgreen}{(+23\%)}\\
        \bottomrule
    \end{tabular}
    % }
\caption{Performance comparison of our stylized synthetic datasets for Autoencoder reconstruction loss, where the Autoencoder is trained on synthetic datasets and evaluated on ImageNet and COCO. Improvements are calculated over the corresponding non-stylized synthetic baseline * from \autoref{tab:base_autoencoder}, $\text{improvement}=100-100\times\frac{\text{stylized}}{\text{nonStylized}}$. Reverse stylization improves all synthetic datasets. We also see that \textit{Gatys} datasets outperform \textit{NNST} datasets, with neural fractals datasets remaining the best even after stylization, closely followed by our reverse stylized versions of FractalDB-composite and StyleGAN-oriented.}
\vspace{-0.25em}
\label{tab:stylized_autoencoder_full}
\end{table*}
% & \multicolumn{2}{c|}{{\textbf{AutoEncoding}}\normalsize{$\downarrow$}} & \multicolumn{2}{c}{{\textbf{ImageGen}}\normalsize{$\downarrow$}} & \multicolumn{2}{c}{\footnotesize{\textbf{Representation Learning}}\normalsize{$\uparrow$}} \\

%% file: sec/tables/stylized_dino.tex
\begin{table*}[t]
\renewcommand{\arraystretch}{0.8} % Reduce space between rows
\setlength{\tabcolsep}{6pt} % Reduce horizontal padding
\small % Set a smaller font size
% \resizebox{\linewidth}{!}{
\centering
    \begin{tabular}{lcccccc}
        \toprule
        \textbf{Synthetic dataset} & \multicolumn{2}{c}{\textbf{ImageNet-100}} & \multicolumn{2}{c}{\textbf{Flowers}} & \multicolumn{2}{c}{\textbf{Food101}}\\
        \cmidrule(lr){2-3} \cmidrule(lr){4-5} \cmidrule(lr){6-7}
        & \textit{Gatys} & \textit{NNST} & \textit{Gatys} & \textit{NNST} & \textit{Gatys} & \textit{NNST}\\
        \midrule
        VisualAtom & 55.6 \textcolor{forestgreen}{(+17\%)} & 55.8 \textcolor{forestgreen}{(+17\%)} & 65.4 \textcolor{forestgreen}{(+29\%)} & 68.1 \textcolor{forestgreen}{(+32\%)} & 51.7 \textcolor{forestgreen}{(+27\%)} & 50.7 \textcolor{forestgreen}{(+26\%)}\\
        Mandelbulb & 56.8 \textcolor{forestgreen}{(+13\%)} & 57.1 \textcolor{forestgreen}{(+13\%)} & 63.6 \textcolor{forestgreen}{(+13\%)} & 66.7 \textcolor{forestgreen}{(+17\%)} & 52.6 \textcolor{forestgreen}{(+12\%)} & 54.0 \textcolor{forestgreen}{(+13\%)}\\
        % FDB
        FractalDB-composite & 56.1 \textcolor{forestgreen}{(+11\%)} & 57.8 \textcolor{forestgreen}{(+13\%)} & 67.5 \textcolor{forestgreen}{(+12\%)} & 70.2 \textcolor{forestgreen}{(+15\%)} & 52.9 \textcolor{forestgreen}{(+16\%)} & 54.3 \textcolor{forestgreen}{(+17\%)}\\
        StyleGAN-oriented & 55.9 \textcolor{forestgreen}{(+8\%)} & 58.9 \textcolor{forestgreen}{(+11\%)} & 68.1 \textcolor{forestgreen}{(+20\%)} & 70.7 \textcolor{forestgreen}{(+23\%)} & 52.7 \textcolor{forestgreen}{(+7\%)} & 54.5 \textcolor{forestgreen}{(+9\%)}\\
        % \midrule
        % \rowcolor{lightgray}
        Neural Fractals & \textbf{57.4} \textcolor{forestgreen}{(+9\%)} & \textbf{59.2} \textcolor{forestgreen}{(+11\%)} & \textbf{68.7} \textcolor{forestgreen}{(+8\%)} & \textbf{71.3} \textcolor{forestgreen}{(+11\%)} & \textbf{53.8} \textcolor{forestgreen}{(+4\%)} & \textbf{55.0} \textcolor{forestgreen}{(+5\%)}\\
        \bottomrule
    \end{tabular}
% }
\caption{Performance comparison of our stylized synthetic datasets for DINO with a ViT-S encoder, where the encoder is trained on synthetic datasets and linear evaluation is performed on ImageNet-100, Flowers and Food, measured in accuracy. Improvements are calculated over the corresponding non-stylized synthetic baseline * from \autoref{tab:base_contrastive2}. Unlike \autoref{tab:stylized_all} and \autoref{tab:stylized_autoencoder_full}, $\text{improvement}=\text{stylized}-\text{nonStylized}$ is used to reflect improvements. Reverse stylization improves all synthetic datasets. We also see that \textit{NNST} datasets outperform \textit{Gatys} datasets (except for VisualAtom), with neural fractals datasets remaining the best after stylization.}
\label{tab:stylized_dino}
\end{table*}

%% file: sec/figs/base_grid.tex
\begin{figure*}[htbp]
    \centering
    \begin{minipage}{0.46\textwidth}
        \centering
        \includegraphics[width=\textwidth]{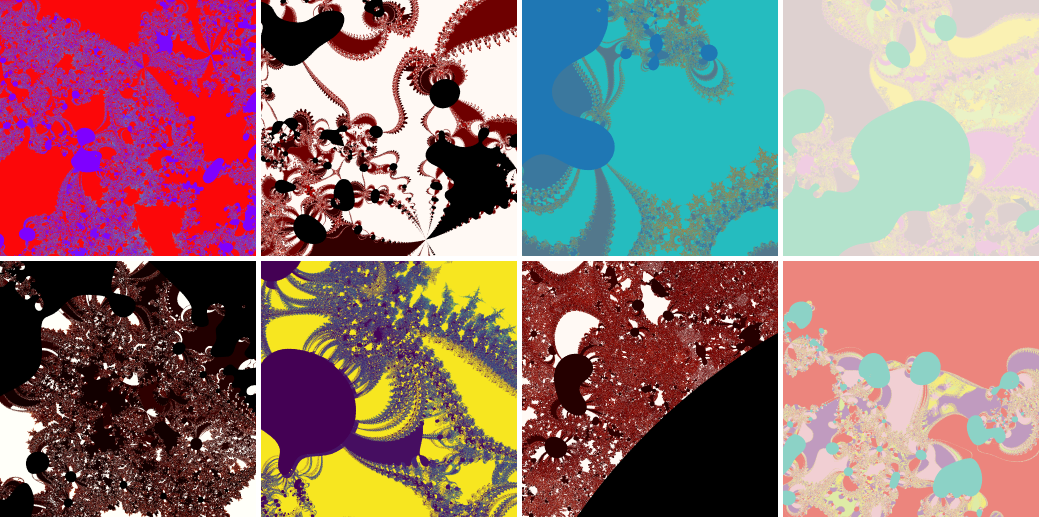}
        \subcaption{Neural Fractal}
        \label{fig:image6}
    \end{minipage}
    \hspace{0.2cm}
    \begin{minipage}{0.46\textwidth}
        \centering
        \includegraphics[width=\textwidth]{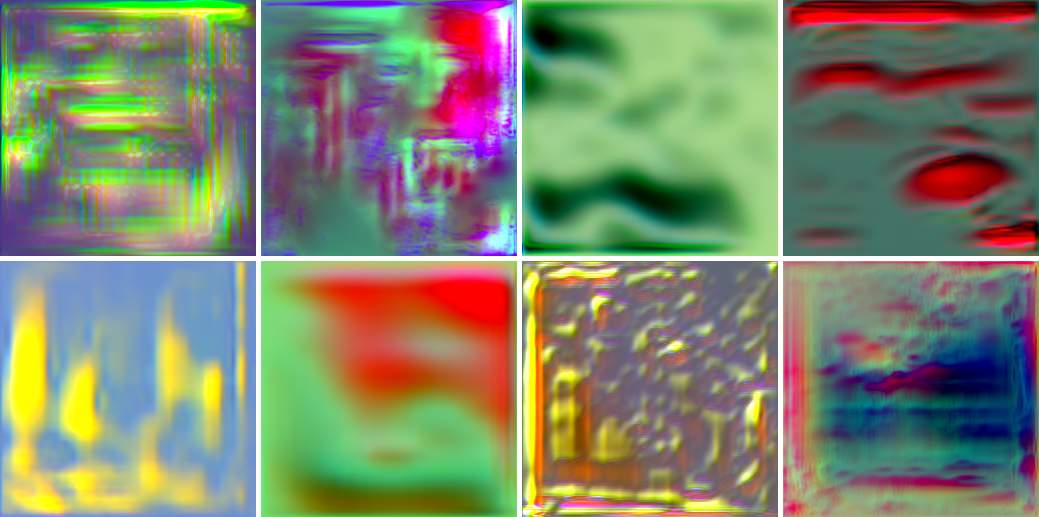}
        \subcaption{StyleGAN-oriented}
        \label{fig:image5}
    \end{minipage}
    \vspace{0.2cm}
    \begin{minipage}{0.3\textwidth}
    \centering
    \includegraphics[width=\textwidth]{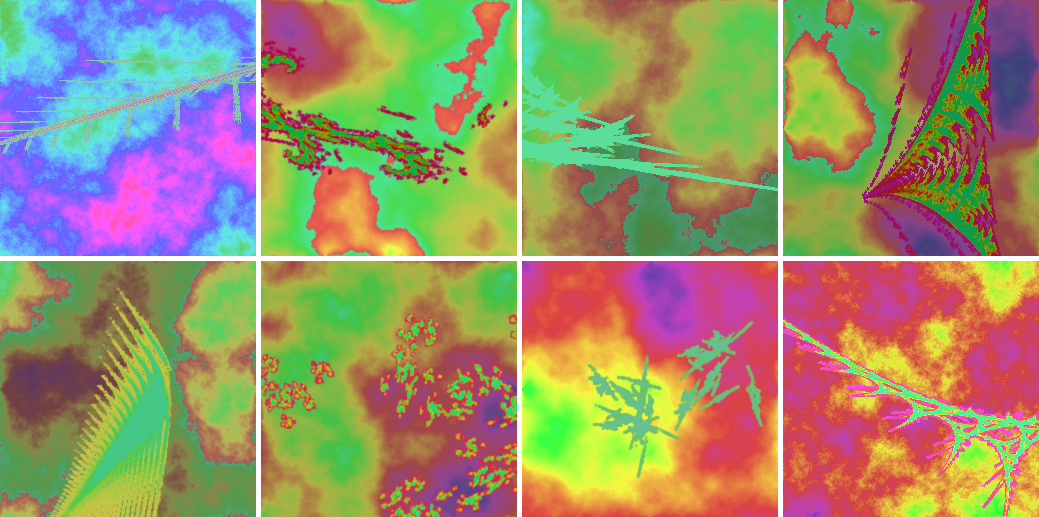}
    \subcaption{FractalDB-composite}
    \label{fig:image2}
    \end{minipage}
    \hspace{0.2cm}
    \begin{minipage}{0.3\textwidth}
        \centering
        \includegraphics[width=\textwidth]{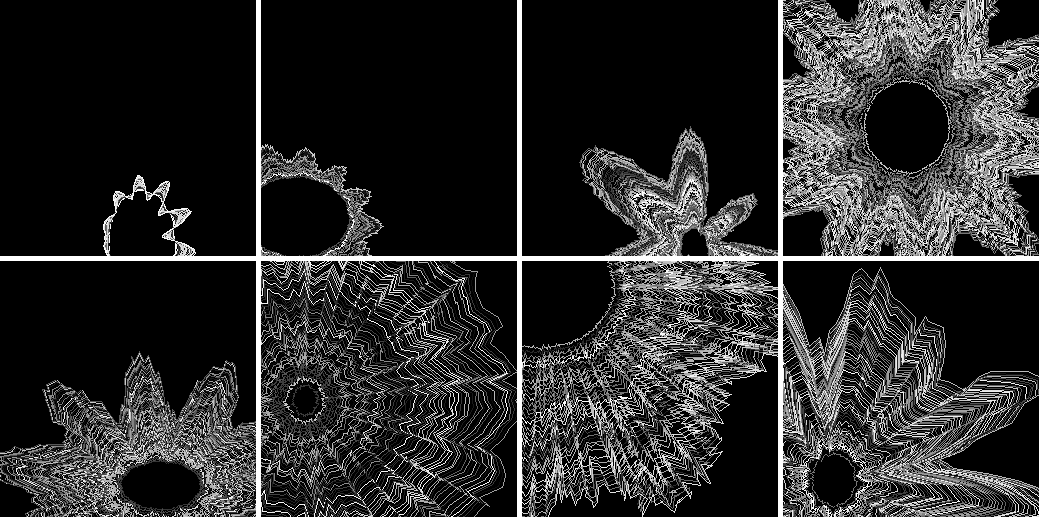}
        \subcaption{VisualAtom}
        \label{fig:image1}
    \end{minipage}
    \hspace{0.2cm}
    \begin{minipage}{0.3\textwidth}
        \centering
        \includegraphics[width=\textwidth]{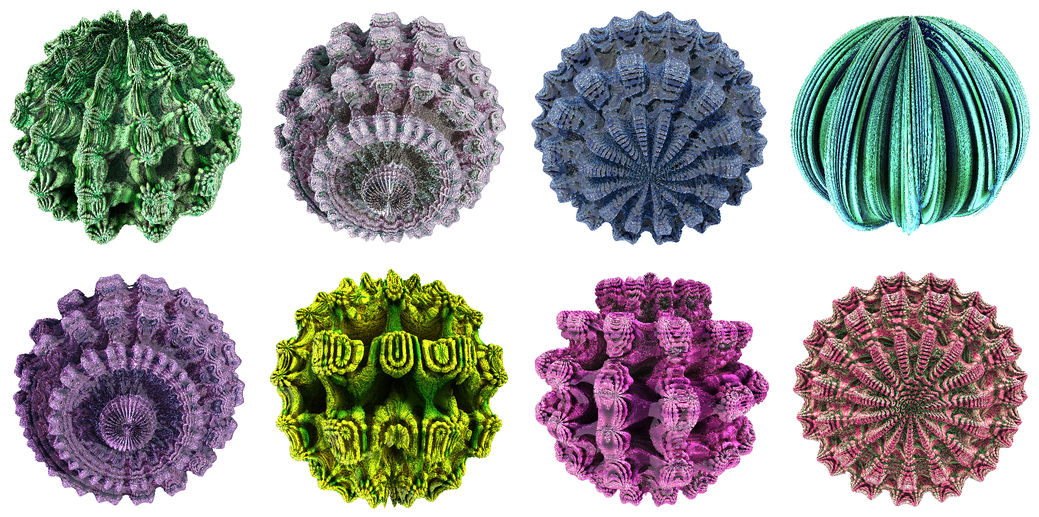}
        \subcaption{Mandelbulb}
        \label{fig:image3}
    \end{minipage}
    \caption{Samples from different base synthetic datasets.}
    \label{fig:base_grid}
\end{figure*}

%% file: sec/supple_flowers_results.tex
\begin{figure*}[htbp]
\centering
\begin{tabular}{@{} >{\raggedleft\arraybackslash}m{4cm} l @{}}

% First row
Without Finetuning & \includegraphics[width=0.6\textwidth]{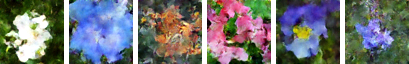} \\

% Second row
Visual Atom & \includegraphics[width=0.6\textwidth]{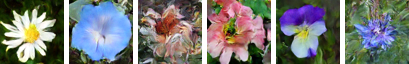} \\

% Third row
Unsplash 64 & \includegraphics[width=0.6\textwidth]{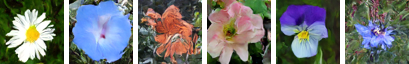} \\

% Fourth row
Mandelbulb & \includegraphics[width=0.6\textwidth]{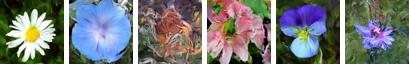} \\

% Fifth row
Style GAN Oriented & \includegraphics[width=0.6\textwidth]{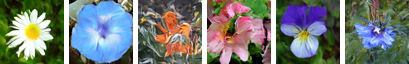} \\

% Sixth row
Fractal DB Composite & \includegraphics[width=0.6\textwidth]{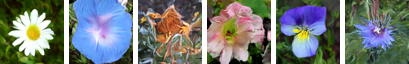} \\

% Seventh row
Neural Fractal-100K & \includegraphics[width=0.6\textwidth]{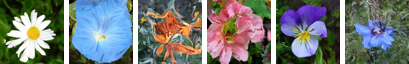} \\

% Eighth row
Neural Fractal 100k NNST & \includegraphics[width=0.6\textwidth]{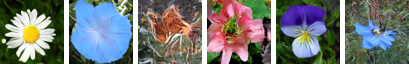} \\

% Ninth row
Neural Fractal-1M & \includegraphics[width=0.6\textwidth]{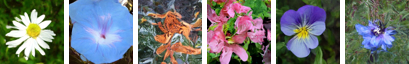} \\

% Tenth row
Neural Fractal-1M-NNST & \includegraphics[width=0.6\textwidth]{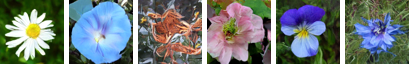} \\

% Eleventh row
ImageNet-100K & \includegraphics[width=0.6\textwidth]{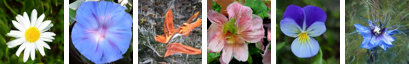} \\

% Twelfth row
ImageNet-1M & \includegraphics[width=0.6\textwidth]{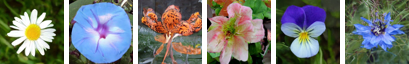} \\

\end{tabular}
\centering % Center the caption
\caption{Comparison of sample images generated by different models trained on the Flowers dataset. The models differ in their pretraining datasets. The images are ordered by their FID scores, with the highest FID at the top and the lowest FID at the bottom.}
\label{fig:model_comparison_generated_flowers}
\end{figure*}

%% file: sec/supple_ffhq_results.tex
\begin{figure*}[htbp]
\centering
\begin{tabular}{@{} >{\raggedleft\arraybackslash}m{4cm} l @{}}

% First row
Without Finetuning & \includegraphics[width=0.6\textwidth]{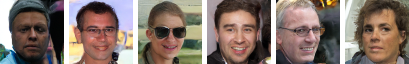} \\

% Second row
Unsplash 64 & \includegraphics[width=0.6\textwidth]{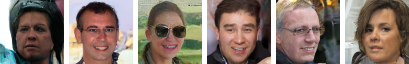} \\

% Third row
Mandelbulb & \includegraphics[width=0.6\textwidth]{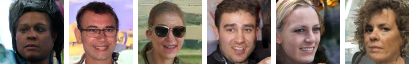} \\

% Fourth row
Fractal DB Composite & \includegraphics[width=0.6\textwidth]{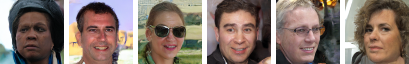} \\

% Fifth row
Visual Atom & \includegraphics[width=0.6\textwidth]{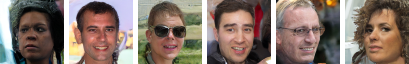} \\

% Sixth row
Style GAN Oriented & \includegraphics[width=0.6\textwidth]{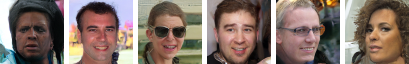} \\

% Seventh row
Neural Fractal-100K & \includegraphics[width=0.6\textwidth]{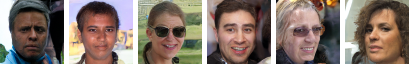} \\

% Eighth row
Neural Fractal 100k NNST & \includegraphics[width=0.6\textwidth]{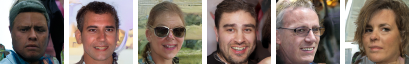} \\

% Ninth row
ImageNet-100K & \includegraphics[width=0.6\textwidth]{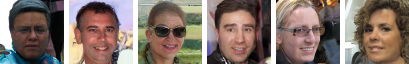} \\

% Tenth row
Neural Fractal-1M & \includegraphics[width=0.6\textwidth]{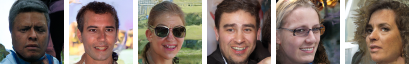} \\

% Eleventh row
Neural Fractal-1M-NNST & \includegraphics[width=0.6\textwidth]{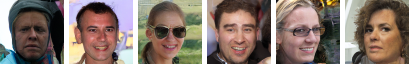} \\

% Twelfth row
ImageNet-1M & \includegraphics[width=0.6\textwidth]{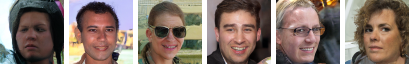} \\

\end{tabular}
\centering % Center the caption
\caption{Comparison of sample images generated by different models trained on the FFHQ dataset. The models differ in their pretraining datasets. The images are ordered by their FID scores, with the highest FID at the top and the lowest FID at the bottom.}
\label{fig:model_comparison_generated_ffhq}
\end{figure*}

%% file: sec/supple_autoencoder_visualizations.tex
\begin{figure*}[htbp]
    \centering
    % We'll create 4 rows (for samples 0,5 from ffhq and 3,4 from flowers)
    % and 8 columns (the 8 image types).
    % The "c|c|cccc|c|c" ensures we have the vertical lines in the correct places:
    %  1) between column 1 (Original) and column 2 (Base ImageNet)
    %  2) between column 2 (Base ImageNet) and columns 3-6
    %  3) between column 6 (Base StyleGAN) and column 7 (Base NeuralFractal)

    \begin{tabular}{c|c|c c c c|c|c}       
        % ========== ROW 1: Sample 0 (ffhq) ==========
        \includegraphics[width=0.10\textwidth]{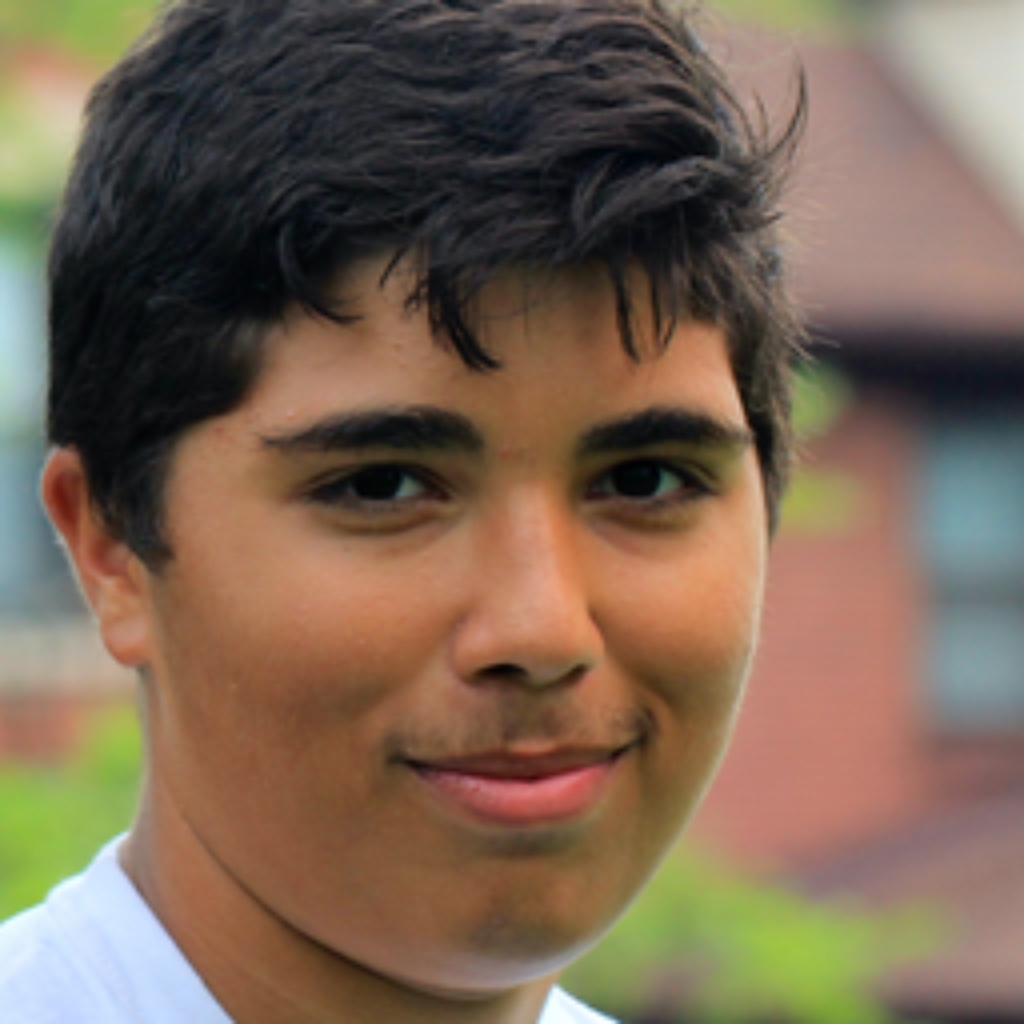} &
        \includegraphics[width=0.10\textwidth]{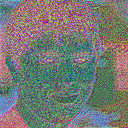} &
        \includegraphics[width=0.10\textwidth]{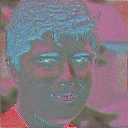} &
        \includegraphics[width=0.10\textwidth]{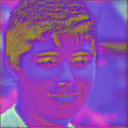} &
        \includegraphics[width=0.10\textwidth]{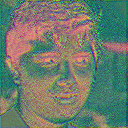} &
        \includegraphics[width=0.10\textwidth]{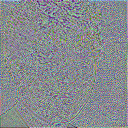} &
        \includegraphics[width=0.10\textwidth]{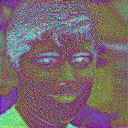} &
        \includegraphics[width=0.10\textwidth]{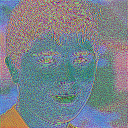} \\
        % ========== ROW 2: Sample 5 (ffhq) ==========
        \includegraphics[width=0.10\textwidth]{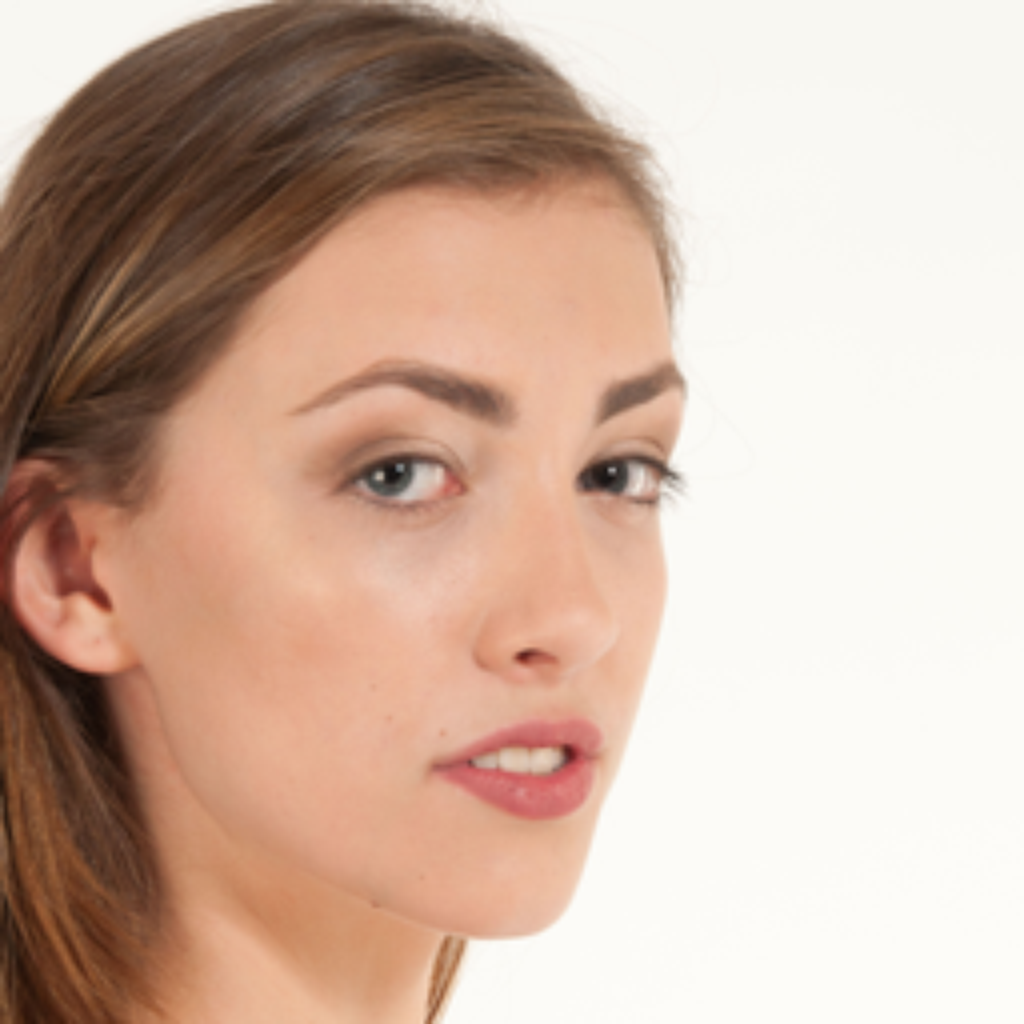} &
        \includegraphics[width=0.10\textwidth]{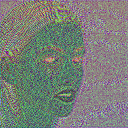} &
        \includegraphics[width=0.10\textwidth]{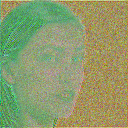} &
        \includegraphics[width=0.10\textwidth]{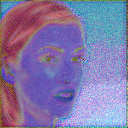} &
        \includegraphics[width=0.10\textwidth]{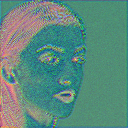} &
        \includegraphics[width=0.10\textwidth]{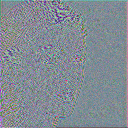} &
        \includegraphics[width=0.10\textwidth]{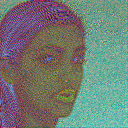} &
        \includegraphics[width=0.10\textwidth]{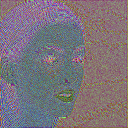} \\
        % ========== ROW 3: Sample 3 (flowers) ==========
        \includegraphics[width=0.10\textwidth]{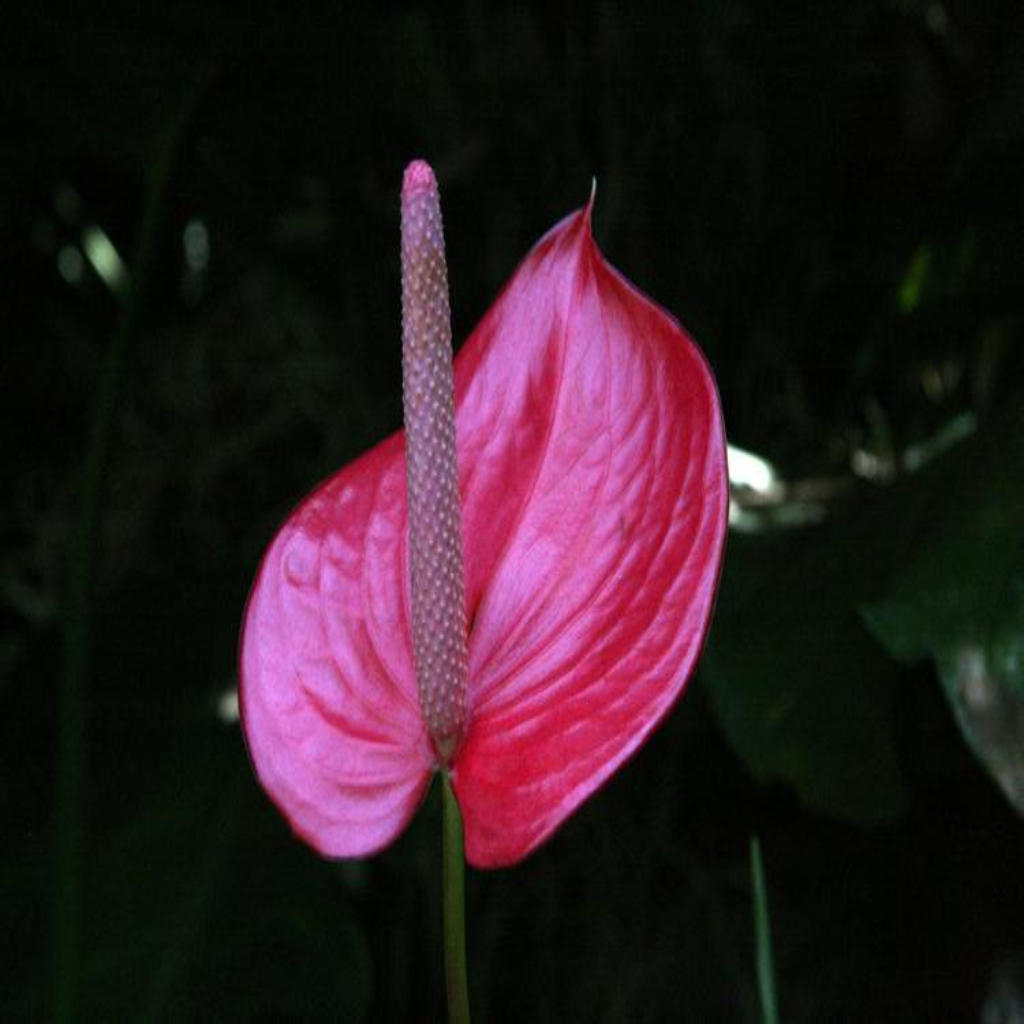} &
        \includegraphics[width=0.10\textwidth]{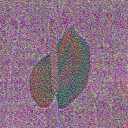} &
        \includegraphics[width=0.10\textwidth]{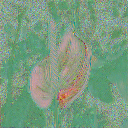} &
        \includegraphics[width=0.10\textwidth]{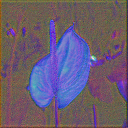} &
        \includegraphics[width=0.10\textwidth]{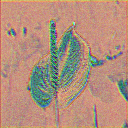} &
        \includegraphics[width=0.10\textwidth]{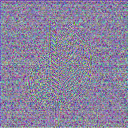} &
        \includegraphics[width=0.10\textwidth]{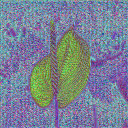} &
        \includegraphics[width=0.10\textwidth]{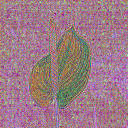} \\
        % ========== ROW 4: Sample 4 (flowers) ==========
        \includegraphics[width=0.10\textwidth]{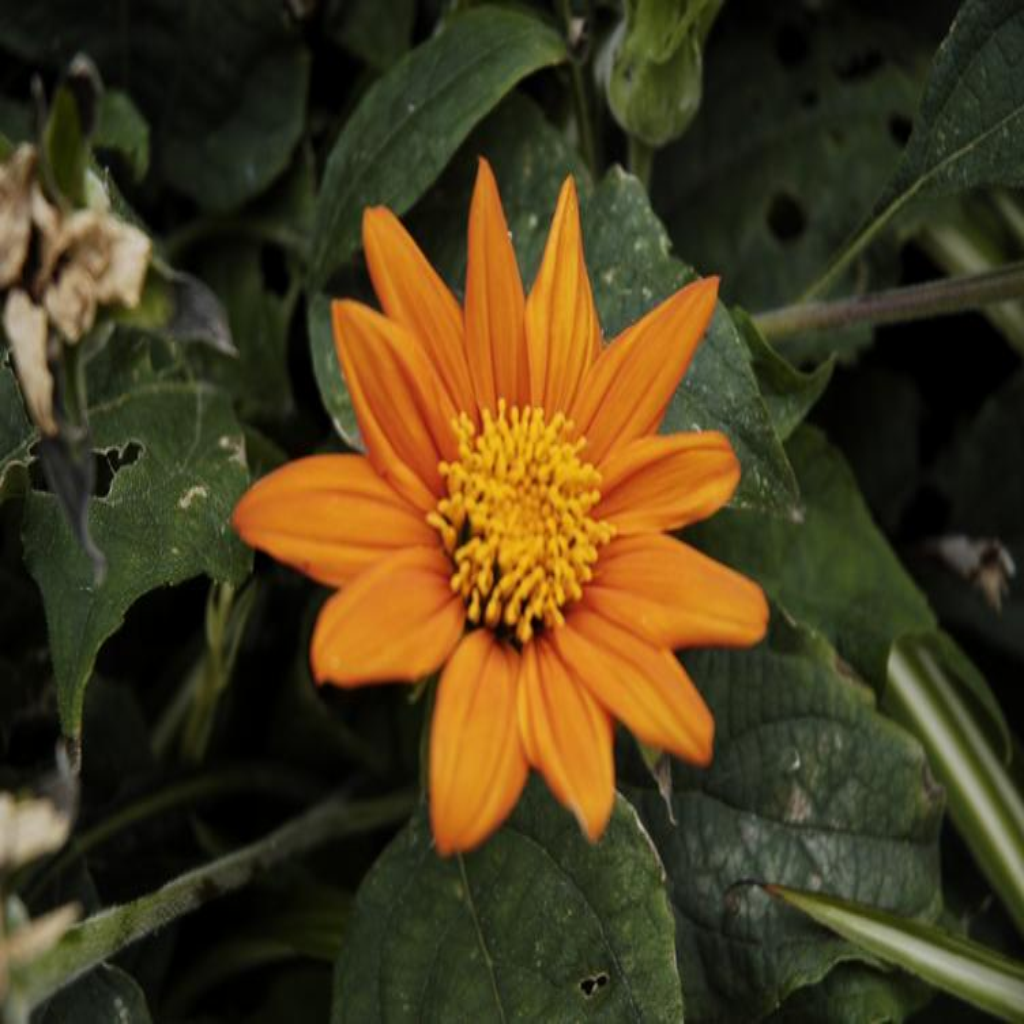} &
        \includegraphics[width=0.10\textwidth]{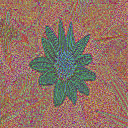} &
        \includegraphics[width=0.10\textwidth]{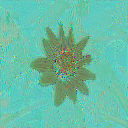} &
        \includegraphics[width=0.10\textwidth]{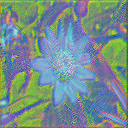} &
        \includegraphics[width=0.10\textwidth]{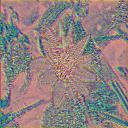} &
        \includegraphics[width=0.10\textwidth]{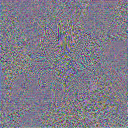} &
        \includegraphics[width=0.10\textwidth]{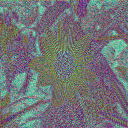} &
        \includegraphics[width=0.10\textwidth]{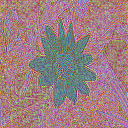} \\

        % ------------------- Bottom Row: Subcaptions -------------------
        \footnotesize{Input} & \footnotesize{ImgNet100} & \footnotesize{VisualAtom} & \footnotesize{Mandelbulb} & \footnotesize{FractalDB} & \footnotesize{StyleGAN} & \footnotesize{Neural Fractal} & \footnotesize{Stylized NF} \\
    \end{tabular}
    \caption{Visualization of latent representations from autoencoders trained on different datasets. The figure consists of samples from FFHQ and Flowers dataset. The similarity between the latents of the Stylized Neural Fractal and ImageNet-100k suggests a reduced domain gap, supporting the effectiveness of the stylization approach.}
    \label{fig:feature_visualization_latents}
\end{figure*}

\begin{figure*}[htbp]
    \centering
    % We'll create 4 rows (for samples 0,5 from ffhq and 3,4 from flowers)
    % and 8 columns (the 8 image types).
    % The "c|c|cccc|c|c" ensures we have the vertical lines in the correct places:
    %  1) between column 1 (Original) and column 2 (Base ImageNet)
    %  2) between column 2 (Base ImageNet) and columns 3-6
    %  3) between column 6 (Base StyleGAN) and column 7 (Base NeuralFractal)

    \begin{tabular}{c|c|c c c c|c|c}       
        % ========== ROW 1: Sample 0 (ffhq) ==========
        \includegraphics[width=0.10\textwidth]{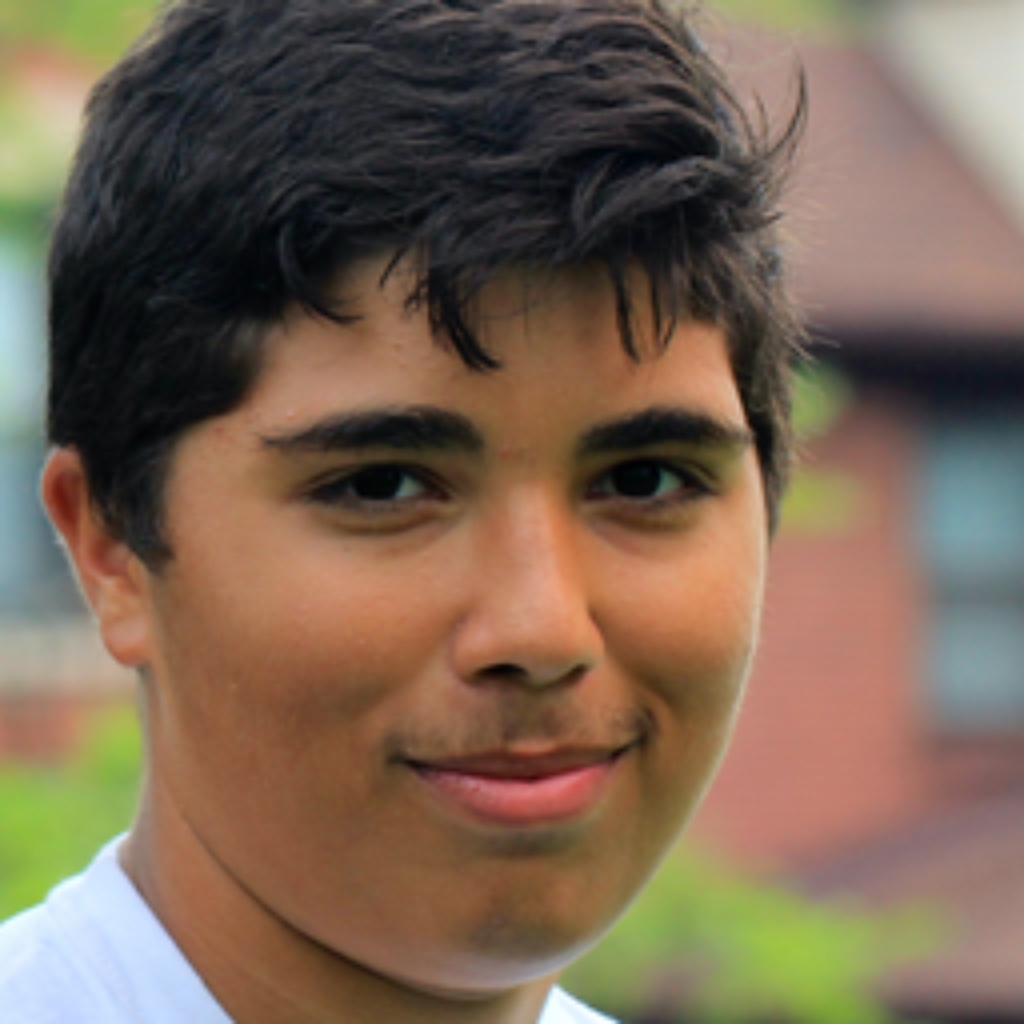} &
        \includegraphics[width=0.10\textwidth]{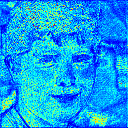} &
        \includegraphics[width=0.10\textwidth]{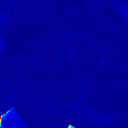} &
        \includegraphics[width=0.10\textwidth]{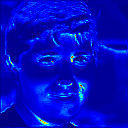} &
        \includegraphics[width=0.10\textwidth]{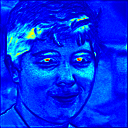} &
        \includegraphics[width=0.10\textwidth]{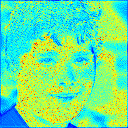} &
        \includegraphics[width=0.10\textwidth]{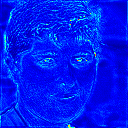} &
        \includegraphics[width=0.10\textwidth]{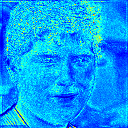} \\
        % ========== ROW 2: Sample 5 (ffhq) ==========
        \includegraphics[width=0.10\textwidth]{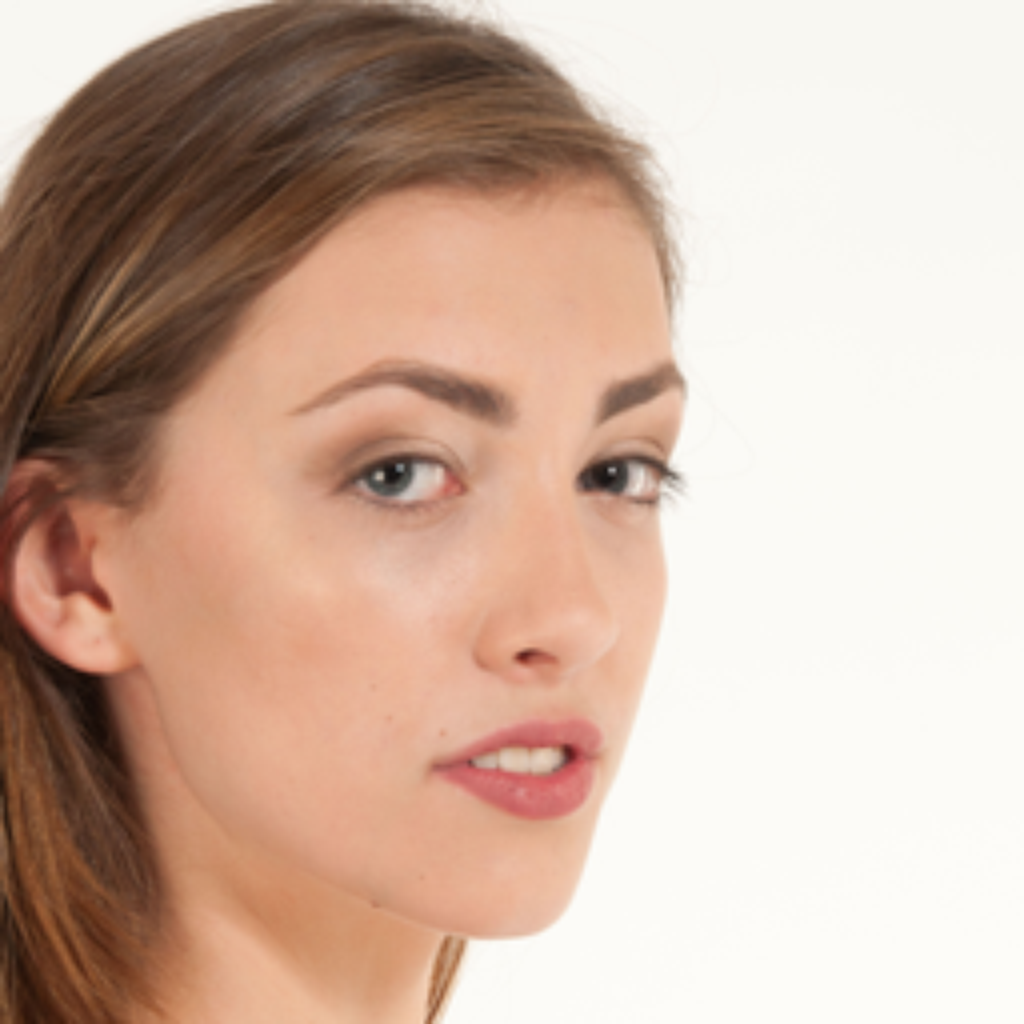} &
        \includegraphics[width=0.10\textwidth]{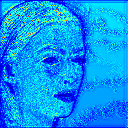} &
        \includegraphics[width=0.10\textwidth]{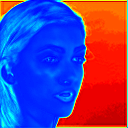} &
        \includegraphics[width=0.10\textwidth]{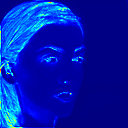} &
        \includegraphics[width=0.10\textwidth]{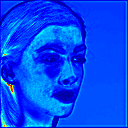} &
        \includegraphics[width=0.10\textwidth]{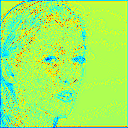} &
        \includegraphics[width=0.10\textwidth]{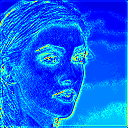} &
        \includegraphics[width=0.10\textwidth]{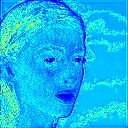} \\
        % ========== ROW 3: Sample 3 (flowers) ==========
        \includegraphics[width=0.10\textwidth]{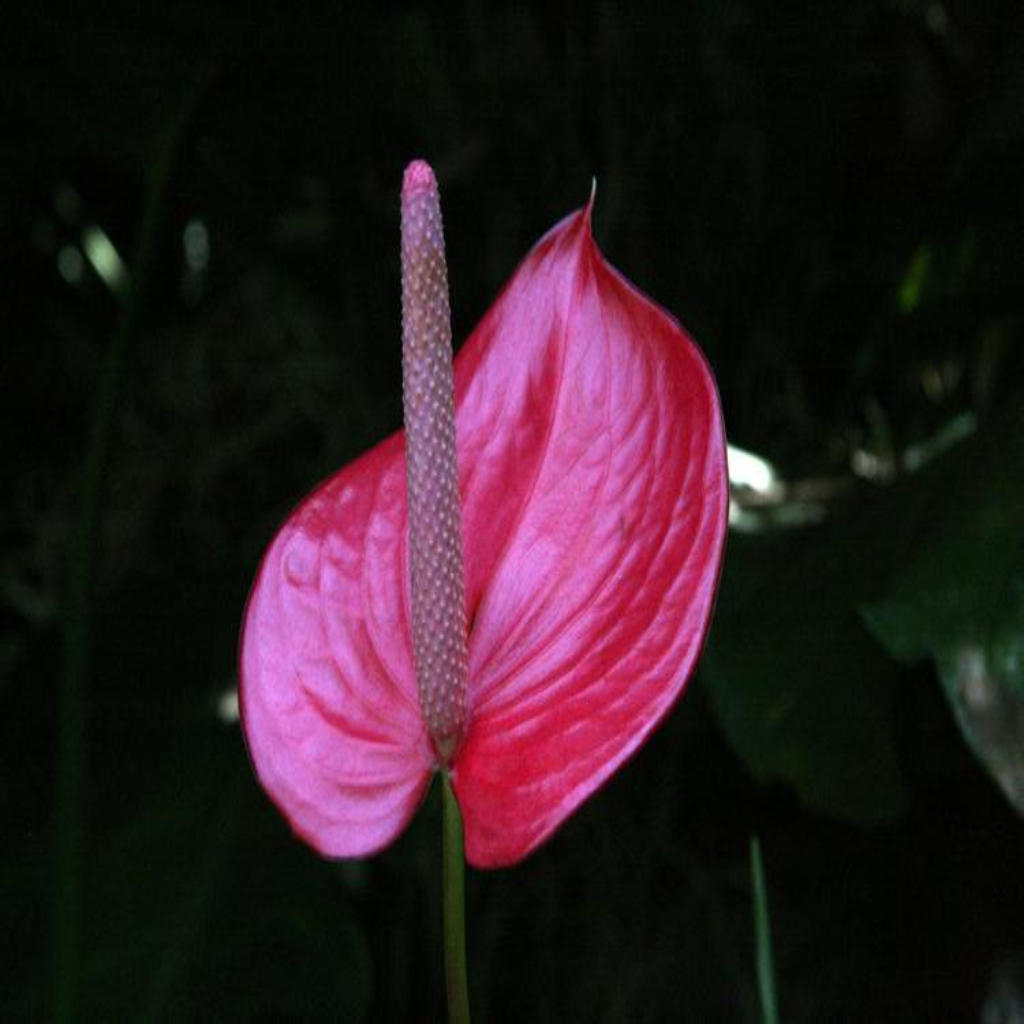} &
        \includegraphics[width=0.10\textwidth]{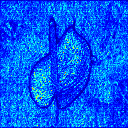} &
        \includegraphics[width=0.10\textwidth]{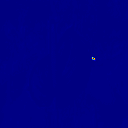} &
        \includegraphics[width=0.10\textwidth]{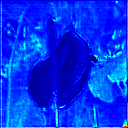} &
        \includegraphics[width=0.10\textwidth]{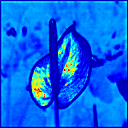} &
        \includegraphics[width=0.10\textwidth]{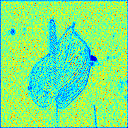} &
        \includegraphics[width=0.10\textwidth]{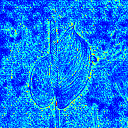} &
        \includegraphics[width=0.10\textwidth]{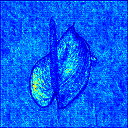} \\
        % ========== ROW 4: Sample 4 (flowers) ==========
        \includegraphics[width=0.10\textwidth]{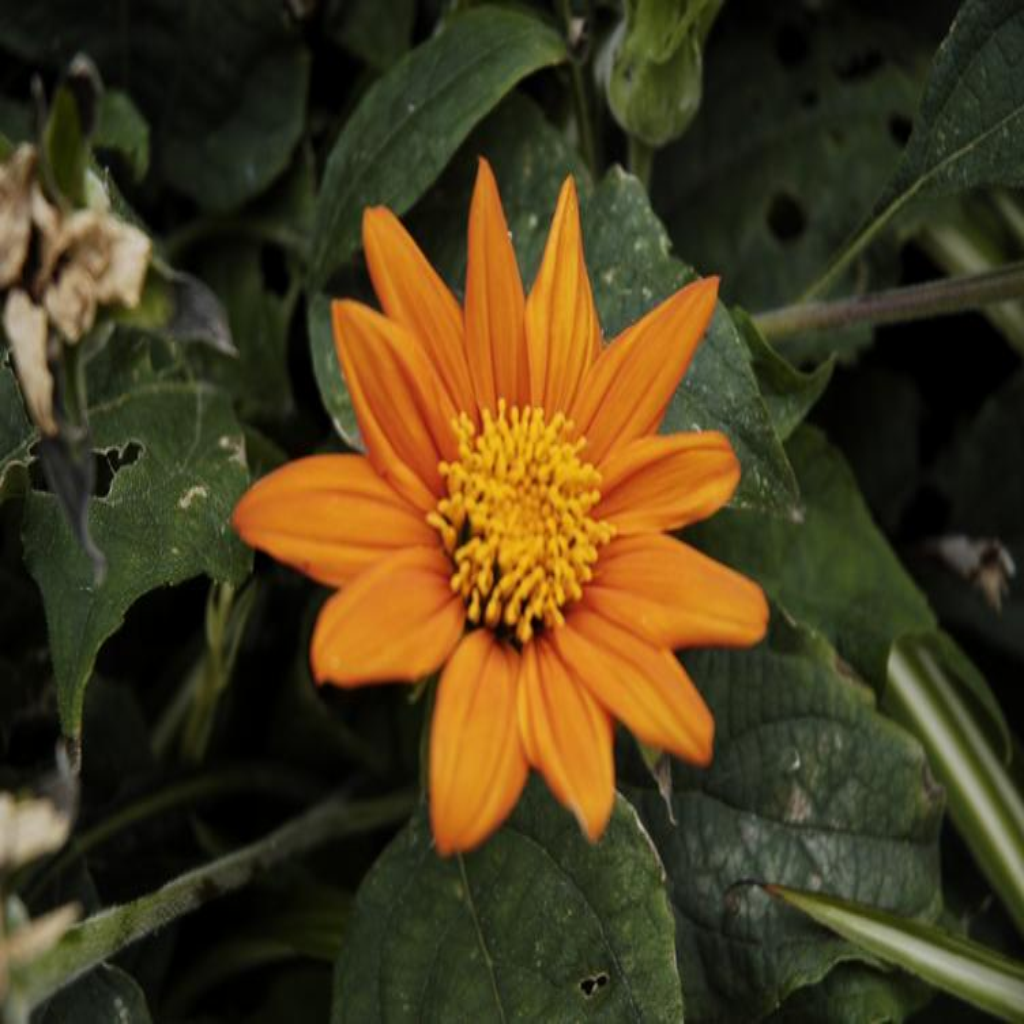} &
        \includegraphics[width=0.10\textwidth]{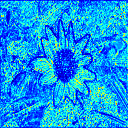} &
        \includegraphics[width=0.10\textwidth]{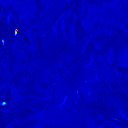} &
        \includegraphics[width=0.10\textwidth]{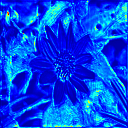} &
        \includegraphics[width=0.10\textwidth]{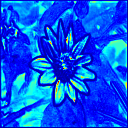} &
        \includegraphics[width=0.10\textwidth]{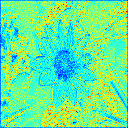} &
        \includegraphics[width=0.10\textwidth]{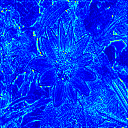} &
        \includegraphics[width=0.10\textwidth]{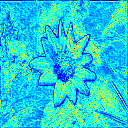} \\

        % ------------------- Bottom Row: Subcaptions -------------------
        \footnotesize{Input} & \footnotesize{ImgNet100} & \footnotesize{VisualAtom} & \footnotesize{Mandelbulb} & \footnotesize{FractalDB} & \footnotesize{StyleGAN} & \footnotesize{Neural Fractal} & \footnotesize{Stylized NF} \\
    \end{tabular}
    \caption{Visualization of attention maps representations from autoencoders trained on different datasets.}
    \label{fig:feature_visualization_attention}
\end{figure*}